\documentclass[10pt,twocolumn,letterpaper]{article}

\usepackage[pagenumbers]{iccv} % To force page numbers, e.g. for an arXiv version

\definecolor{iccvblue}{rgb}{0.21,0.49,0.74}
\usepackage{graphicx}
\usepackage{colortbl}
\usepackage{xcolor}
\usepackage{booktabs}
\usepackage{caption}
\usepackage{float}
\usepackage{subcaption}
\usepackage{booktabs}
\usepackage{placeins}
\usepackage{microtype}
\usepackage{array,multirow}
\usepackage[pagebackref,breaklinks,colorlinks,allcolors=iccvblue]{hyperref}
\usepackage[capitalize]{cleveref}  % Should be loaded after 'hyperref', and works perfectly with 'subfigure'.

\newcommand{\emoji}[1]{\includegraphics[height=0.8em]{#1}\xspace}

\def\paperID{2057} % *** Enter the Paper ID here
\def\confName{ICCV}
\def\confYear{2025}

\def\authorBlock{
	Zitong Zhang \qquad
	Suranjan Gautam \qquad
	Rui Yu \\
	University of Louisville \\
	{\tt\small \{zitong.zhang, suranjan.gautam, rui.yu\}@louisville.edu} \\
    \small{\url{https://top2pano.github.io/}}
}

\begin{document}

% \title{Top2Pano: Generating Indoor Panoramic Images based on Top-down View}
\title{Top2Pano: Learning to Generate Indoor Panoramas from Top-Down View}

%%%%%%%%% AUTHORS - PLEASE UPDATE
% \and

\author{\authorBlock}

\twocolumn[{%
\renewcommand\twocolumn[1][]{#1}%
\maketitle

\begin{center}
    \centering
    \vspace{-7mm}
    \includegraphics[trim={1cm 0 0 0}, width=\linewidth]{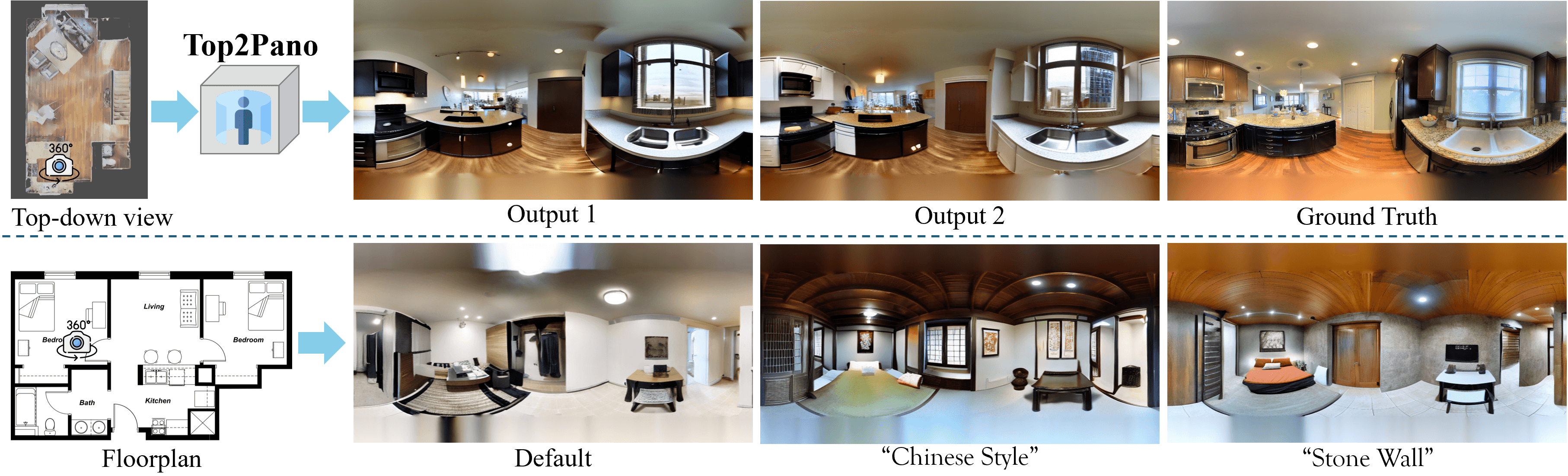}
    \vspace{-6mm}
    \captionof{figure}{{\color{black}\textbf{Top:} We present \textit{Top2Pano}, a method for synthesizing high-quality indoor panoramas from a top-down view. Given a camera position, Top2Pano generates panoramas that are both visually compelling and geometrically accurate. \textbf{Bottom:} Our model demonstrates strong generalization capabilities. When provided with schematic floor plans as input, Top2Pano produces photorealistic and structurally coherent panoramas. Additionally, our approach can be easily adapted for stylized synthesis, enabling diverse design variations. Note: The original dataset (Matterport3D~\cite{Matterport3D}) contains blurry regions near the upper and lower edges of the panoramic image.}
    }
    \label{fig:teaser}
\end{center}%
}]

\begin{abstract}

Generating immersive 360° indoor panoramas from 2D top-down views has applications in virtual reality, interior design, real estate, and robotics. This task is challenging due to the lack of explicit 3D structure and the need for geometric consistency and photorealism. We propose \textbf{Top2Pano}, an end-to-end model for synthesizing realistic indoor panoramas from top-down views. Our method estimates volumetric occupancy to infer 3D structures, then uses volumetric rendering to generate coarse color and depth panoramas. These guide a diffusion-based refinement stage using ControlNet, enhancing realism and structural fidelity. Evaluations on two datasets show Top2Pano outperforms baselines, effectively reconstructing geometry, occlusions, and spatial arrangements. It also generalizes well, producing high-quality panoramas from schematic floorplans. Our results highlight Top2Pano's potential in bridging top-down views with immersive indoor synthesis.
\vspace{-1em}

\end{abstract}
    
\begin{figure*}[t]
    \centering
    \includegraphics[width=\linewidth]{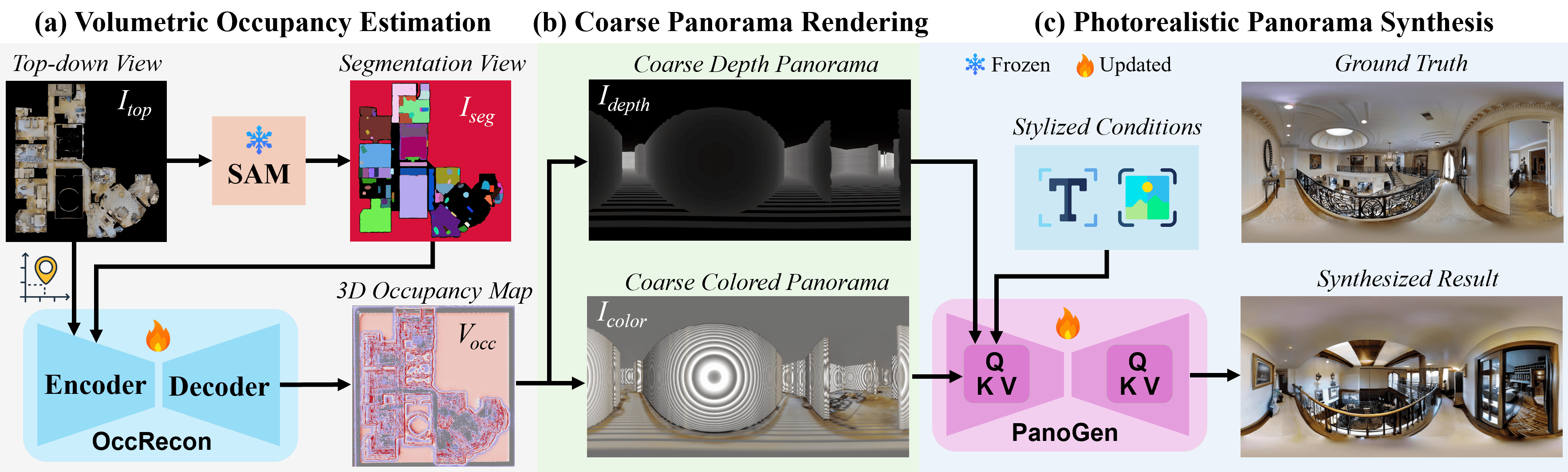}
    \caption{{\color{black}\textbf{Overview of the proposed Top2Pano pipeline.} The pipeline begins by segmenting the top-down view using SAM~\cite{kirillov2023segment}. Both the segmented top-down image and the original top-down view are then processed by the OccRecon module to estimate the scene's 3D volumetric occupancy. Next, given the camera position, the system employs volumetric rendering to generate coarse depth and color panoramas. These coarse images are subsequently refined by the PanoGen module to produce the final photorealistic panorama. The PanoGen module also supports stylized panorama generation based on textual or visual conditions.}}
    \label{fig:pipeline}
\end{figure*}

\section{Introduction}

Understanding and synthesizing immersive indoor scenes from minimal structural information is a fundamental challenge in computer vision and graphics~\cite{liu2025computer,zheng2025panorama,yu2024nerf,huang2025duospacenet,liu2025towards,tan2025planarsplatting}. The ability to generate realistic indoor panorama images from a 2D top-down view holds immense potential for a wide range of applications, including virtual reality (VR)~\cite{lyu2019refocusable}, interior design~\cite{pintore2023deep}, real estate visualization~\cite{huang2025scene4u}, and robotics~\cite{guerrero2020s}. For instance, real estate platforms can leverage this technology to offer potential buyers photorealistic virtual walkthroughs generated directly from architectural floorplans, enhancing the property viewing experience. Similarly, VR applications can benefit from automatically synthesized environments that create more engaging and immersive user experiences. Additionally, robots operating in indoor environments can utilize synthesized panoramas to improve their spatial understanding and navigation capabilities, enabling more efficient and accurate movement in complex spaces.
Despite its broad applicability, the task of generating high-quality indoor panoramas from top-down views remains surprisingly underexplored in the literature. Recent advancements in large multimodal models have enabled the synthesis of panoramas directly from text input~\cite{feng2023diffusion360,panfusion2024,ye2024diffpano}; however, these approaches often overlook critical geometric and textural constraints. Other studies have focused on generating 3D models from semantic layouts~\cite{bahmani2023cc3d,yang2024scenecraft,chen2025layout2scene,feng2024prim2room,schult24controlroom3d,fang2023ctrl}; however, these methods are often limited by the quality of the resulting 3D meshes, making them unsuitable for rendering high-quality panoramas. Furthermore, semantic information is often unavailable in top-down views or floorplans, posing an additional challenge for existing approaches.

Addressing this gap requires overcoming several significant technical challenges. First, a 2D top-down view provides only limited visual cues about the actual appearance and layout of the scene, making it difficult to infer occluded structures and fine texture details. Second, generating geometrically consistent indoor scenes demands accurate reasoning about 3D spatial occupancy from a 2D input, which is inherently ambiguous. Third, achieving photorealism while maintaining structural coherence necessitates a synthesis approach that effectively balances fidelity and realism, ensuring that the generated scenes are both visually appealing and functionally accurate.

To tackle these challenges, we introduce \textbf{Top2Pano}, a novel framework for generating photorealistic indoor panorama images from 2D top-down views. Our approach consists of three main stages. First, we learn the volumetric occupancy of the indoor scene, enabling the model to infer plausible spatial structures and layout configurations. Next, we employ volumetric rendering to generate coarse depth and colored panorama images, providing an initial estimate of the scene’s appearance and geometry. Finally, we refine the synthesized panoramas using a diffusion-based model~\cite{zhang2023adding} conditioned on the coarse representations, enhancing both realism and structural consistency. By incorporating learned occupancy priors and diffusion-based refinement, our model effectively bridges the gap between schematic top-down views and immersive indoor panoramas, producing results that are both visually compelling and geometrically accurate.

We evaluate Top2Pano on two indoor datasets and demonstrate its effectiveness compared to baseline methods. Our model not only generates higher-quality images with improved geometric consistency but also exhibits strong generalization capabilities. Even when provided with schematic floorplans as input, Top2Pano can produce photorealistic and structurally coherent panoramas. Moreover, we show that our method can be easily adapted for stylized synthesis, allowing for diverse design variations and enabling users to explore different interior aesthetics with ease.

Our key contributions are as follows:
\begin{itemize}
  \item We introduce Top2Pano, a novel framework for generating indoor panoramas from 2D top-down views, integrating volumetric occupancy learning, coarse synthesis, and diffusion-based refinement to achieve high-quality results.
  
  \item We conduct extensive experiments on two indoor datasets, demonstrating that our model surpasses baseline methods in both image quality and structural consistency, setting a new benchmark for this task.
  
  \item We show that Top2Pano generalizes well to schematic floorplans, producing high-quality, geometry-consistent panoramas. Furthermore, our approach supports stylish synthesis, enabling the generation of panoramas with diverse interior design aesthetics, making it a versatile tool for various applications.
\end{itemize}

\section{Related Work}

\subsection{Panorama Generation}
Traditionally, panoramas were generated using image stitching and feature matching methods. With the recent advancements in generative machine learning, text-driven panorama generation techniques have gained popularity. These methods have utilized GANs, VAEs or a combination of GANs and VAEs~\cite{chen2022text2light} and more recently, diffusion models~\cite{feng2023diffusion360,panfusion2024,ye2024diffpano} to synthesize panoramic images from textual descriptions. Another popular field of research is panorama synthesis from narrow-FoV images using image out-painting. Some methods rely solely on narrow-FOV images~\cite{akimoto2022diverse}, while other incorporate textual descriptions alongside the images~\cite{10044439,Wang_2023,kalischek2025cubediff}. Cross-view panorama generation is another well-explored area, particularly challenging due to large shifts in camera perspective. Most techniques in this domain have focused on generating ground-view panoramas from aerial images. Some approaches directly use top-down images as input~\cite{wu2022cross}, while others~\cite{shi2022geometry,9156602,10377305,xu2024geospecific} extract geometric and segmentation information from top-down images to enhance quality. To the best of our knowledge, there is no prior work that explores the generation of indoor panoramic images from floor plans or top-down views of indoor spaces.

{\color{black}
\subsection{Layout-Guided 3D Scene Generation}
Recent approaches to 3D scene generation leverage layouts for semantic and physical plausibility. Plan2Scene~\cite{Vidanapathirana2021Plan2Scene} reconstructs 3D meshes from floor plans, while ATISS~\cite{Paschalidou2021NEURIPS} employs Transformers conditioned on scene layout. CC3D~\cite{bahmani2023cc3d} follows a 3D GAN-based approach using 2D semantic layouts. Diffusion-based methods such as SceneCraft~\cite{yang2024scenecraft}, Layout2Scene~\cite{chen2025layout2scene}, and Prim2Room~\cite{feng2024prim2room} have further improved synthesis quality. ControlRoom3D~\cite{schult24controlroom3d} and Ctrl-Room~\cite{fang2023ctrl} are closely related to our work, as both generate panoramas during reconstruction process. Unlike prior methods that rely on \textit{explicit semantic layouts} detailing object classes, positions, orientations, and sizes, our approach only requires a top-down view -- an easily obtainable and lightweight input format. Instead of generating full 3D scenes, we produce panoramas, offering a more efficient and realistic solution for AR/VR, and autonomous robotic navigation by enabling immersive experiences like virtual tours and supporting real-time robotic navigation.
}

\section{Method}
The pipeline of our Top2Pano method is illustrated in Figure~\ref{fig:pipeline}. Given an input top-down view $I_{\mathrm{top}} \in \mathbb{R}^{H \times W \times 3}$, we first generate its segmentation map $I_{\mathrm{seg}} \in \mathbb{R}^{H \times W \times 3}$ using a pretrained model. Both the top-down view and its segmentation, along with the specified camera position, are then fed into an encoder-decoder occupancy estimation module, \mbox{\textit{OccRecon}}, which reconstructs a 3D volumetric occupancy map $V_{\mathrm{occ}} \in \mathbb{R}^{H \times W \times N}$, where $N$ represents the number of vertical voxels. From this occupancy map, we render a coarse depth panorama $I_{\mathrm{depth}}$ via volumetric rendering and project colors from the top-down view to obtain a coarse color panorama $I_{\mathrm{color}}$. To ensure geometric consistency, we enforce structural constraints on walls and floor, preserving occlusion relationships and realistic spatial structure. Finally, both coarse depth and color panoramas serve as conditions for a diffusion-based synthesis module, \textit{PanoGen}, which generates photorealistic panoramic images $I_{\mathrm{pano}} \in \mathbb{R}^{H \times W \times 3}$. These images faithfully capture the scene's spatial layout, furniture, and fine color details. Additionally, PanoGen module supports stylized synthesis with optional textual or imagery-based controls.

\subsection{Volumetric Occupancy Estimation}

The 2D top-down views lack 3D structural information about objects and furniture. To render panoramas that accurately reflect the geometric spatial relationships of objects, we propose training an \textit{OccRecon} module to estimate the scene's 3D occupancy or density.

\smallskip
\noindent{\bf Input Representations.}
Unlike the layout-guided 3D scene generation setting, our top-down input lacks semantic information. In our preliminary study, we found that current semantic segmentation models struggle to generalize to indoor top-down views, making it challenging to estimate the 3D structure without semantic guidance. To address this, we propose leveraging a pretrained segmentation model SAM~\cite{kirillov2023segment} to extract the 2D structure of the scene. Both the top-down image and the segmentation view are then fed into the encoder of the OccRecon module. The segmentation provides valuable details, such as room boundaries, furniture positions, and shapes, which significantly enhance the OccRecon module's ability to learn the overall 3D structure of the rooms. Moreover, this semantic-free input design enables our model to generalize effectively to more abstract inputs, such as semantic floorplans.

\smallskip
\noindent{\bf OccRecon Module.}
We propose a diffusion-based encoder-decoder framework that efficiently extracts and reconstructs the 3D spatial structure of a scene. Instead of using computationally intensive 3D diffusion models, our approach leverages a 2D diffusion model~\cite{zhang2023adding}, significantly reducing resource demands. The OccRecon module processes 2D images and segmentations to extract spatial information and reconstruct room structures, including wall and furniture heights. In the final step, a 3D convolutional layer integrates the learned spatial features to generate a comprehensive 3D occupancy map. By relying on 2D inputs for most of the process, our method drastically lowers computational costs while still capturing the full scene layout. This balance of efficiency and accuracy makes it a practical solution for 3D structural modeling without semantic inputs.
The OccRecon module outputs a 3D volumetric occupancy map $V_{\mathrm{occ}}$, which explicitly represents the scene's 3D geometry.
\begin{equation}
\label{DensityMapping}
V_{\mathrm{occ}} = \texttt{OccRecon}(I_{\mathrm{top}},I_{\mathrm{seg}})_{condition} 
\end{equation} 
We normalize the occupancy values to the range $[0,1]$. Given the scale information of the scene, we transform the real-world coordinates of a 3D point into the estimated volumetric space. The point's density is then obtained by querying the learned occupancy map $V_{\mathrm{occ}}$ using tri-linear interpolation.

\smallskip
\noindent{\bf Structural Reinforcement.}
Top-down views offer a comprehensive overview of the entire floor, but when observed from a first-person perspective, the scene typically reveals only the details of the current space, obscuring rooms beyond the walls. After the model learns the overall geometric structure of the rooms through the OccRecon module, we apply structural reinforcement to refine the wall geometry. This provides precise depth information, enhancing the reconstruction of the room's geometric layout. We solidify the wall voxels by applying the maximum value (1 after normalization). Additionally, top-down priors can constrain the floor's geometry and help infer the texture of the furniture, which is essential for generating accurate indoor panoramas with correctly placed and colored furniture. Since the OccNet modules are optimized end-to-end with the subsequent modules, the height of the furniture can be inferred from the 3D occupancy map. By encoding structural constraints from the walls and floor, we achieve a more accurate representation of the scene's geometry, including the positions, colors, and other attributes of the furniture.

%%%%%%%%

\subsection{Coarse Panorama Rendering}

Given the 3D occupancy map of the scene, we render coarse depth and color images based on the specified camera position. To ensure accurate mapping, we first compute the ratio between pixel resolution and physical dimensions. With the room height known, we apply this ratio along with the pixel coordinates in the top-down image to determine the corresponding position in the occupancy map $V_{\mathrm{occ}}$. To generate the panoramic image from the 3D occupancy map, we employ equirectangular projection along with a spherical coordinate system.

The coarse depth panorama $I_{\mathrm{depth}}$ is obtained through volumetric rendering~\cite{max2002optical} of the occupancy map. The depth of a projected ray at pixel $(u, v)$ is computed as follows:
\begin{equation}
\label{depth}
I_{\mathrm{depth}}^{(u,v)} = \sum_{i=1}^{S} T_{i} \alpha_{i} d_{i},
\quad 
T_{i} = \prod_{j=1}^{i-1} (1 - \alpha_{j}),
\end{equation}
where $\alpha_{i}$ represents the transparency level, and $d_{i}$ denotes the distance from the sampled position to the camera.

The coarse color panorama $I_{\mathrm{color}}$ is obtained by directly projecting the indoor top-down view along the camera rays~\cite{10377305}, assigning colors based on the corresponding intersections and bilinear interpolation. Specifically, the color at pixel $(u, v)$ is determined by sampling the top-down image without learning a radiance field. 
\begin{equation} 
I_{\mathrm{color}}^{(u,v)} = \sum_{i}^{S} T_i \alpha_i c_i
\end{equation}
where $c_i$ is the color copied from the top-down image. This approach directly maps color from the top-down view, offering a more straightforward and computationally efficient solution compared to NeRF~\cite{mildenhall2020nerf}, which reconstructs scenes by learning a radiance field for view synthesis.

We employed a uniform voxel sampling strategy, where voxels were sampled along a fixed-length ray for both coarse color and coarse depth. However, this approach introduced banding artifacts, particularly noticeable on the floor directly beneath the camera in the coarse color image. Since the top-down view served as the floor texture, maintaining high-quality details was essential for achieving realistic rendering. To mitigate these artifacts, we reduced the ray length by half for coarse color sampling. This adjustment increased the density of sample points within the same spatial region, enhancing sampling accuracy and producing smoother color and texture transitions. As a result, banding artifacts were significantly diminished, improving overall rendering quality. For coarse depth, preserving scene structure was paramount, so we retained the original sampling strategy.

\subsection{Photorealistic Synthesis}

Generating photorealistic panoramic images directly from top-down views is challenging. To address this, we propose a two-stage pipeline. In the second stage, the PanoGen module synthesizes photorealistic indoor panoramic images from coarse color and depth inputs. We implement PanoGen using a diffusion-based ControlNet~\cite{zhang2023adding}, enabling the restoration of fine details. This two-stage approach not only reconstructs the house’s structural layout, including precise wall positions, but also restores elements such as windows, lighting, and furniture. The final panoramic image $I_{\mathrm{pano}}$ is generated based on the two coarse inputs from the previous stage:
\begin{equation}
\label{Generation}
I_{\mathrm{pano}} = \texttt{PanoGen}(I_{\mathrm{color}},I_{\mathrm{depth}})_{condition}
\end{equation}

The PanoGen module modifies the conditioning mechanism of ControlNet by treating coarse color and depth images as separate inputs before combining them. This design enables PanoGen to capture both the scene's geometric structure and the furniture's color and position, resulting in more accurate and high-quality panoramic images. To further enhance consistency, we incorporate an alignment loss to prevent structural distortions when the viewpoint changes and a color loss to ensure accurate color reproduction in the synthesized output.

\subsection{Training and Optimization}
The Top2Pano model employs denoising MSE loss, alignment loss, and color loss functions for optimization. The denoising MSE loss function \cite{zhang2023adding} is defined as:
\begin{equation}
    \mathcal{L}_{\mathrm{diff}} = \mathbb{E}_{z_0, t, c_t, c_c, c_d, \epsilon \sim \mathcal{N}(0,1)} \left[ \|\epsilon - \epsilon_{\theta}(z_t, t, c_t, c_c, c_d)\|_2^2 \right],
\end{equation}
where \( t \) denotes the number of noise addition steps, \( c_{\text{c}} \) represents the corresponding coarse color image, and \( c_{\text{d}} \) represents the coarse depth image. The variables \( t \) and \( c_t \) correspond to the time step and simple text prompts, respectively. The diffusion algorithm trains a neural network \( \epsilon_{\theta} \) to predict the noise added to the noisy image \( z_t \).
The alignment loss is formulated as:
\begin{equation}
    \mathcal{L}_{\mathrm{alignment}} =  \left[ \| I_{D}-\hat{I_{D}} \|_2^2 \right],
\end{equation}
where $I_{D}$ represents the depth images generated by the model and $\hat{I_{D}}$ represents the ground truth depth images. This loss function is used to address distortions in the furniture.

Let \( I \) be the rendered image and \( G \) the ground truth image, both with \( C = 3 \) color channels. For each channel \( c \in \{1, \dots, C\} \), we define the normalized histogram as  
\begin{equation}
H^c(I) = \bigl( h^c_1(I), h^c_2(I), \dots, h^c_{\text{bins}}(I) \bigr),
\end{equation}
where \(\text{bins}\) is the number of histogram bins (\textit{e.g.}, 256), and each \( h^c_k(I) \) represents the normalized frequency (probability) of pixel intensities falling into bin \( k \). Likewise, for the ground truth image \( G \), we have  
\begin{equation}
H^c(G) = \bigl( h^c_1(G), h^c_2(G), \dots, h^c_{\text{bins}}(G) \bigr).
\end{equation}
The color histogram loss, measured using the \( L_1 \) norm, is given by  
\begin{equation}
\begin{split}
\mathcal{L}_{\mathrm{color}}(I, G) &= \sum_{c=1}^{C} \left\| H^c(I) - H^c(G) \right\|_{1} \\
&= \sum_{c=1}^{C} \sum_{k=1}^{\text{bins}} \bigl| h^c_k(I) - h^c_k(G) \bigr|.
\end{split}
\end{equation}
The final loss function combines three loss terms as follows:  
\begin{equation}
    \mathcal{L} = \mathcal{L}_{\mathrm{diff}} + \mathcal{L}_{\mathrm{alignment}} + \mathcal{L}_{\mathrm{color}}.
\end{equation}

\subsection{Generalization and Stylized Synthesis}
\noindent{\bf Generalization to Floorplan.}
We aim to generate first-person view panoramas that faithfully represent a scene. While top-down views lack vertical details like walls and windows, this omission grants flexibility in generating panoramas. This flexibility is especially useful in interior design, where inputs are often simple floorplans. To improve generalization, particularly to schematic floorplans, we train on \textit{orthographic} rather than perspective views, as they better match floorplans. Empirical results show that Top2Pano generalizes well to schematic and even hand-drawn floorplans while maintaining photorealism. Additionally, our model enables stylized synthesis guided by text or images, supporting diverse design needs.

\smallskip
\noindent{\bf Text-Guided Stylization.}
The PanoGen module, built upon the text-driven Stable Diffusion model, inherently supports text-conditioned image generation. As illustrated in Figure~\ref{fig:pipeline}, PanoGen synthesizes panoramas using three conditions: coarse depth, coarse colored image, and stylized textual guidance. When the input is a textureless floorplan, the stylized textual condition effectively guides the style of the synthesized result, as demonstrated in Figure~\ref{fig:teaser}. However, when a colored top-down view is provided, the influence of text-guided stylization becomes less pronounced. This occurs because the rendered coarse colored panorama constrains the final output to closely align with the input view, thereby diminishing the impact of the textual stylization.

To enhance text-guided stylization, the weight of the coarse colored panorama in the PanoGen conditions can be reduced. However, this introduces a tradeoff: prioritizing stylization may come at the expense of fidelity to the top-down view. Regardless of this tradeoff, the coarse depth condition consistently ensures that the synthesized panorama adheres to the underlying scene geometry. Notably, this tradeoff does not apply to the text-to-panorama generation task, where textual guidance plays a more dominant role.

\smallskip
\noindent{\bf Image-Guided Stylization.}
Given several scene images (not necessarily panoramas), we can fine-tune the PanoGen module using low-rank adaptation (LoRA)~\cite{hu2022lora} to generate panoramas that align with the visual styles present in the provided images. To guide this process, we introduce structured textural prompts augmented with style tags (\textit{e.g.}, \texttt{[Japanese]}) at the beginning of the input prompts. These tags act as conditional modifiers, steering the model toward synthesizing images that follow specific aesthetic themes, such as regional design styles.
The framework is applied solely to the PanoGen module, ensuring both computational and parameter efficiency. Notably, our method requires fewer than five in-the-wild images per target style, substantially reducing data demands. This approach achieves its efficiency by decomposing weight updates into low-rank matrices. For a pretrained weight matrix $\mathbf{W}_0 \in \mathbb{R}^{d \times d}$, the update $\Delta \mathbf{W}$ is constrained as follows:
\begin{equation}
\Delta \mathbf{W} = \mathbf{B} \mathbf{A}, \quad \text{where } \mathbf{A} \in \mathbb{R}^{d \times r}, \mathbf{B} \in \mathbb{R}^{r \times d}
\end{equation}
Here, $r \ll d$ represents the intrinsic rank (we use $r = 8$). During fine-tuning, only the matrices $\mathbf{A}$ and $\mathbf{B}$ are updated, while the original weights $\mathbf{W}_0$ remain fixed. The forward pass is modified as follows:
\begin{equation}
\mathbf{h}_{\mathrm{out}} = \mathbf{W}_0 \mathbf{h}_{\mathrm{in}} + \alpha \cdot \mathbf{B} \mathbf{A} \mathbf{h}_{\mathrm{in}}
\end{equation}
where $\alpha$ is a scaling coefficient. This lightweight adaptation (0.8\% new parameters) mitigates catastrophic forgetting and maintains the model's baseline generation quality for generic prompts while enabling precise style control through our \texttt{[style]} textual conditioning.

\section{Experiments}

\begin{table}[ht]
    \centering
    \resizebox{1.0\linewidth}{!}{
    \begin{tabular}{lccc|ccc}
        \toprule
        & \multicolumn{3}{c}{\textbf{Training}} & \multicolumn{3}{c}{\textbf{Testing}} \\
        \cmidrule(lr){2-4} \cmidrule(lr){5-7}
          & Scenes  & Floors & Panoramas & Scenes & Floors & Panoramas  \\
        \midrule
        Matterport3D & 61 & 127 & 6177 & 14 & 29 & 1405 \\
        Gibson & 152 & 203 & 5379 & 39 & 76 & 1672 \\
        \bottomrule
    \end{tabular}
    }
    \vspace{-2mm}
    \caption{The numbers of scenes, floors, and panorama images in the training and testing sets of the two datasets}
    \label{tab:datasets}
\end{table}

\subsection{Data Preparation}
For evaluation, we use the Matterport3D~\cite{Matterport3D} and Gibson~\cite{xiazamirhe2018gibsonenv} datasets. Since no existing dataset provides both top-down views and high-quality panoramic images, we generate top-down views from 3D models in these datasets using Blender. Specifically, we import textured 3D meshes into Blender and render top-down views with an orthographic camera.
The top-down view we render closely resembles a floorplan, unlike the perspective-rendered views used in embodied dialog localization~\cite{hahn2020you}. This similarity enhances our model's ability to generalize to floorplan inputs.
To determine the number of floors in each scene, we apply DBSCAN~\cite{ester1996density} clustering to the camera positions within the datasets. We exclude certain scenes, such as airports and large supermarkets, as well as panoramic images depicting outdoor environments to ensure alignment with our task. After processing, the final dataset sizes are summarized in Table~\ref{tab:datasets}.

\begin{table}[t!]
\centering
\resizebox{0.99\linewidth}{!}{
\begin{tabular}{c|lcccc}
\hline
       &\textbf{Method}      & PSNR $\uparrow$ & SSIM $\uparrow$ & FID $\downarrow$& LPIPS $\downarrow$ \\  \hline

        \parbox[t]{2mm}{\multirow{4}{*}{\rotatebox[origin=c]{90}{\textbf{{\scriptsize Matterport3D}}}}} & Sat2Density\cite{10377305}+LDM\cite{rombach2022high}  & 11.27& 0.4221& 100.06& 0.6237 \\
     & Sat2Density\cite{10377305}+ControlNet\cite{zhang2023adding} & 11.42& 0.4222& 85.78&0.6163 \\
     & PanFusion\cite{panfusion2024} & 11.45 & 0.4372 & 85.74 & 0.6153  \\    \cline{2-6}  
      & Top2Pano (Ours)    & \textbf{11.72} & \textbf{0.4409}&\textbf{30.84}& \textbf{0.6029}   \\  \hline

        \parbox[t]{2mm}{\multirow{4}{*}{\rotatebox[origin=c]{90}{\textbf{Gibson}}}} & Sat2Density\cite{10377305}+LDM\cite{rombach2022high}  & 10.54&0.4480 &84.33&0.6462 \\
     & Sat2Density\cite{10377305}+ControlNet\cite{zhang2023adding}  &10.97  & 0.4582& 85.21 & 0.6645 \\
     & PanFusion\cite{panfusion2024} & 11.36 & 0.4744&79.53&0.6634  \\    \cline{2-6}  
      & Top2Pano (Ours)    & \textbf{11.58} & \textbf{0.4851} &\textbf{28.68}&\textbf{0.6282}   \\  \hline
\end{tabular}
}
\vspace{-2mm}
\caption{Quantitative comparison with existing methods on the Matterport3D~\cite{Matterport3D} and Gibson~\cite{xiazamirhe2018gibsonenv} datasets.}
\label{tab:results_baseline}
\vspace{-1.5mm}
\end{table}

\subsection{Evaluation Metrics}
\vspace{-0.2em}
To assess the quality of the generated panoramas, we employ both pixel-based and perceptual evaluation metrics. For pixel-level assessment, we utilize peak signal-to-noise ratio (PSNR) and structural similarity index measure (SSIM) to quantify image fidelity. Additionally, we incorporate perceptual metrics such as Fréchet Inception Distance (FID)~\cite{heusel2017gans} and Learned Perceptual Image Patch Similarity (LPIPS)~\cite{zhang2018perceptual} to capture higher-level visual realism.

\subsection{Implement Details}
Our code runs on an NVIDIA RTX A6000 GPU with 48GB of memory. The model has 3.3 billion parameters and is trained with a batch size of 21 for 100 epochs. On average, each experiment takes approximately two days to complete on both the Matterport3D and Gibson datasets. We optimize our model using the Adam optimizer \cite{kingma2014adam} with default parameters ($\beta_1 = 0.9$, $\beta_2 = 0.999$, $\epsilon = 10^{-8}$) and a learning rate of $10^{-5}$. 

\subsection{Comparison with Previous Methods}
\begin{figure*}[ht]
    \centering
    \footnotesize
    \setlength{\tabcolsep}{1.5pt}
    \begin{tabular}{ccccc}
    \centering
        {\includegraphics[width=0.106\linewidth]{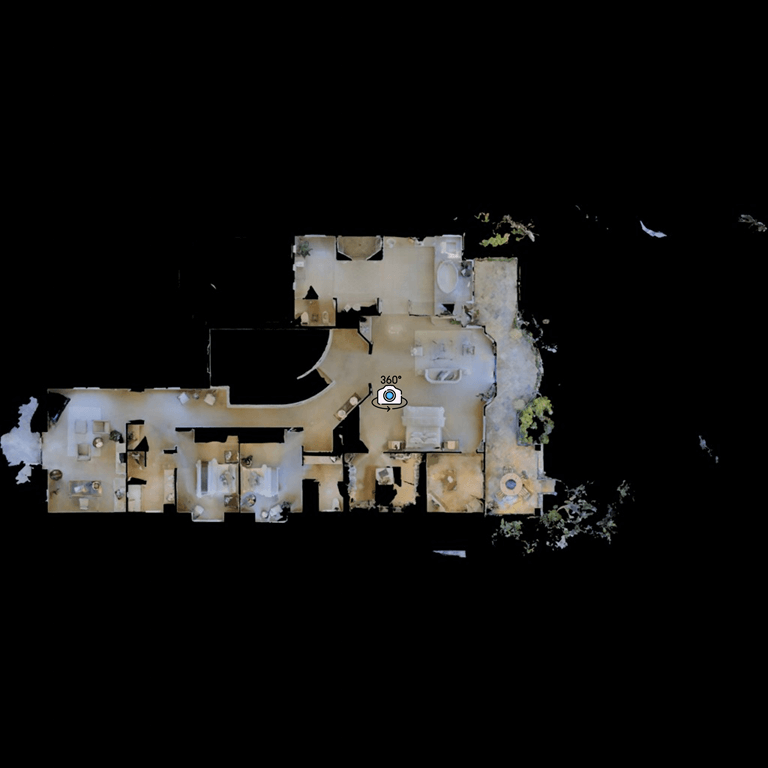}} &  
        {\includegraphics[width=0.212\linewidth]{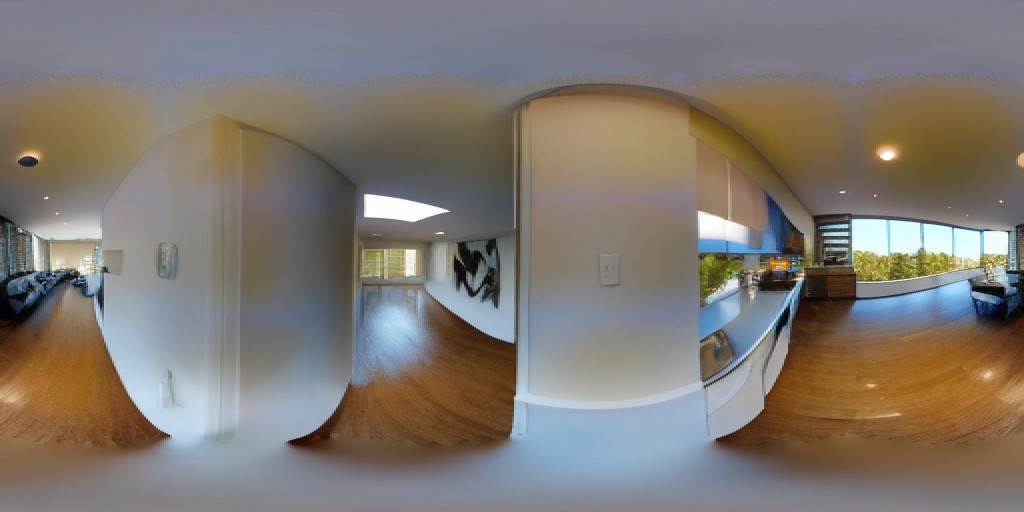}} &  
        {\includegraphics[width=0.212\linewidth]{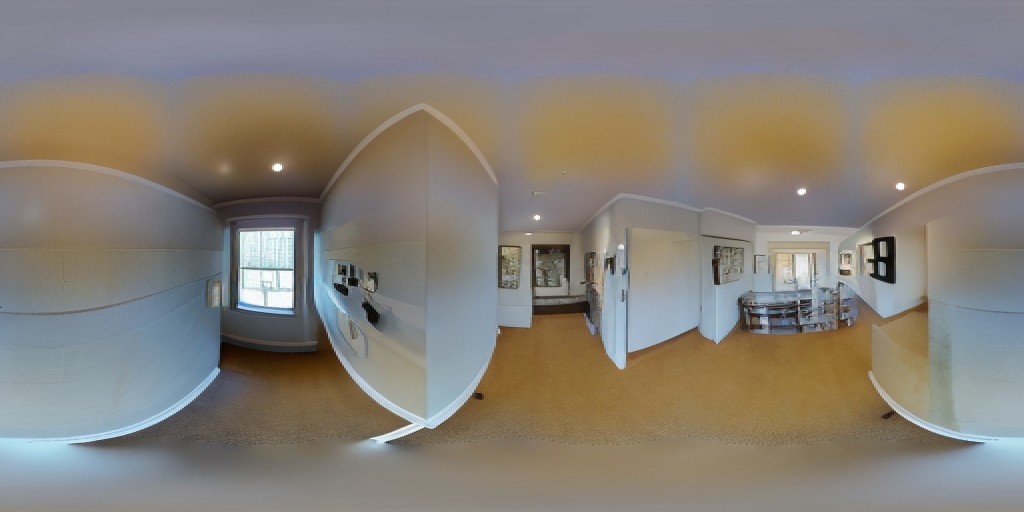}} &  
        {\includegraphics[width=0.212\linewidth]{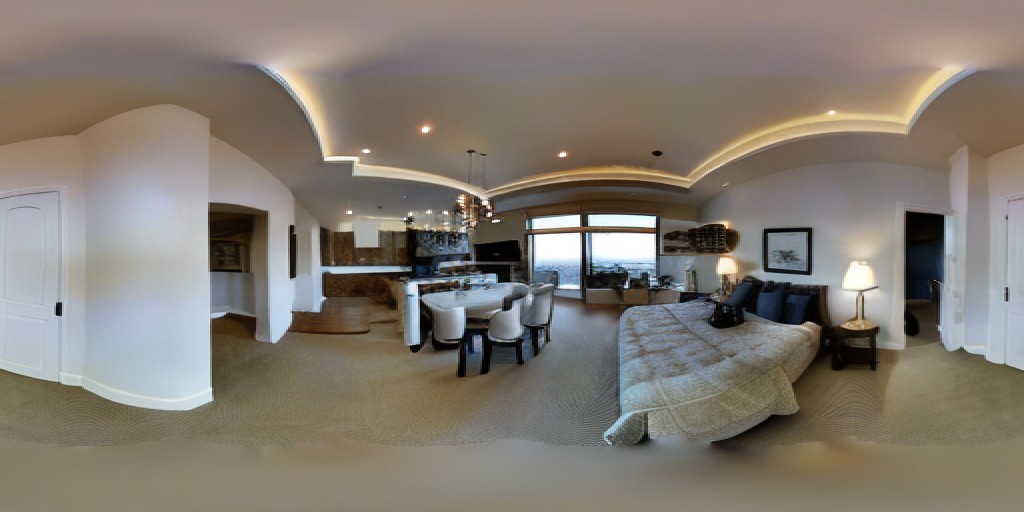}} &  
        {\includegraphics[width=0.212\linewidth]{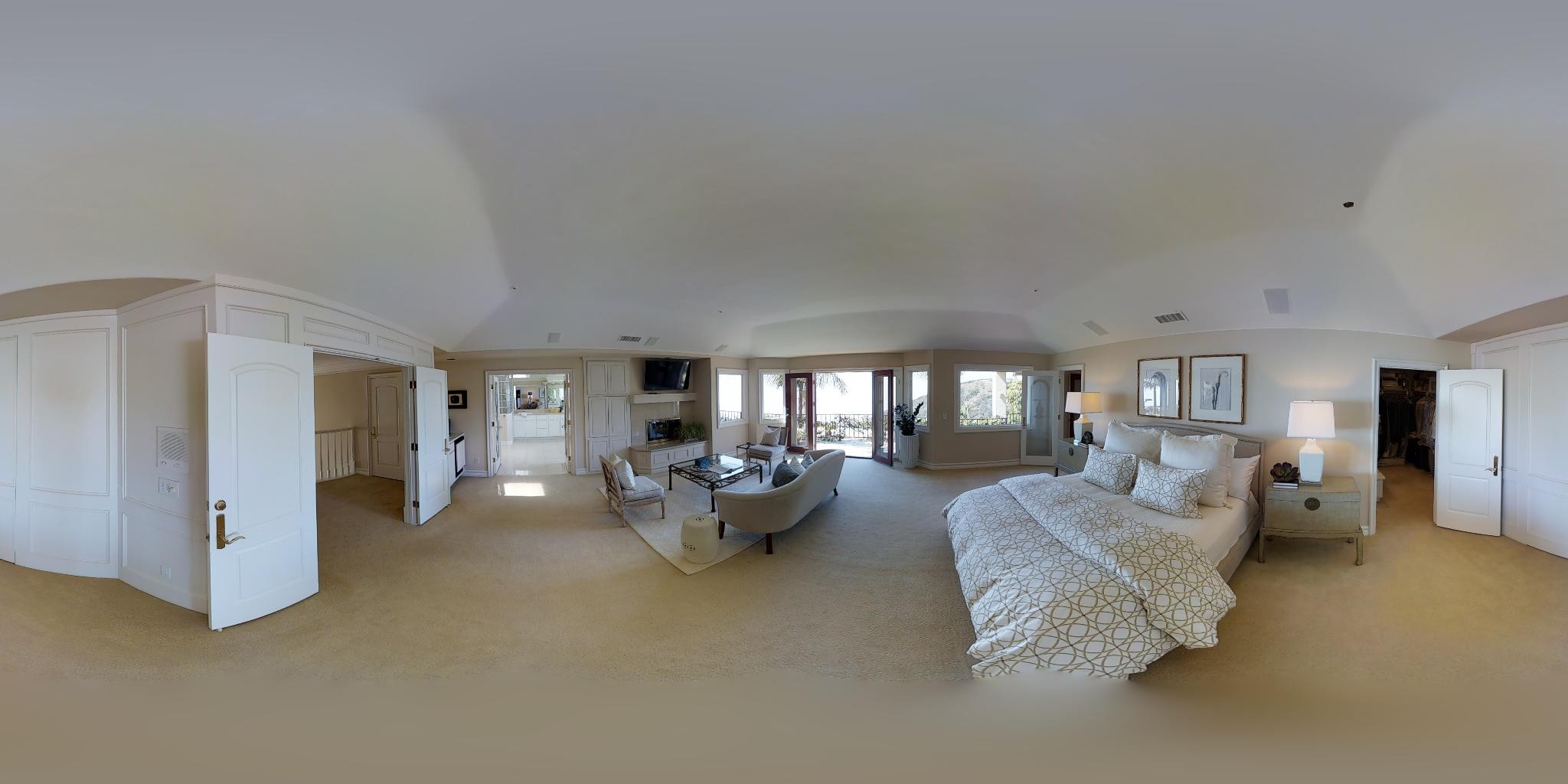}} \\
        {\includegraphics[width=0.106\linewidth]{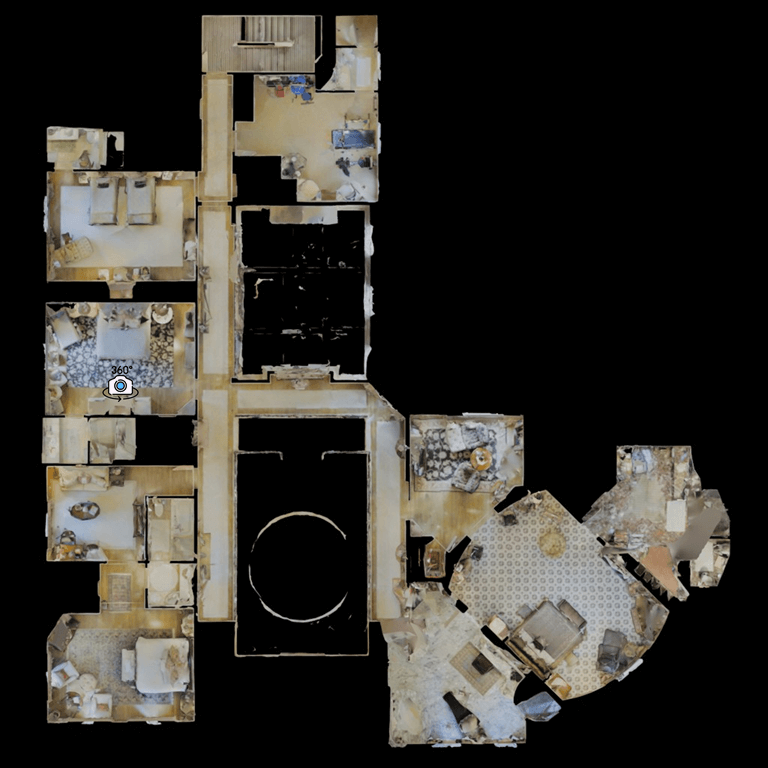}} &  
        {\includegraphics[width=0.212\linewidth]{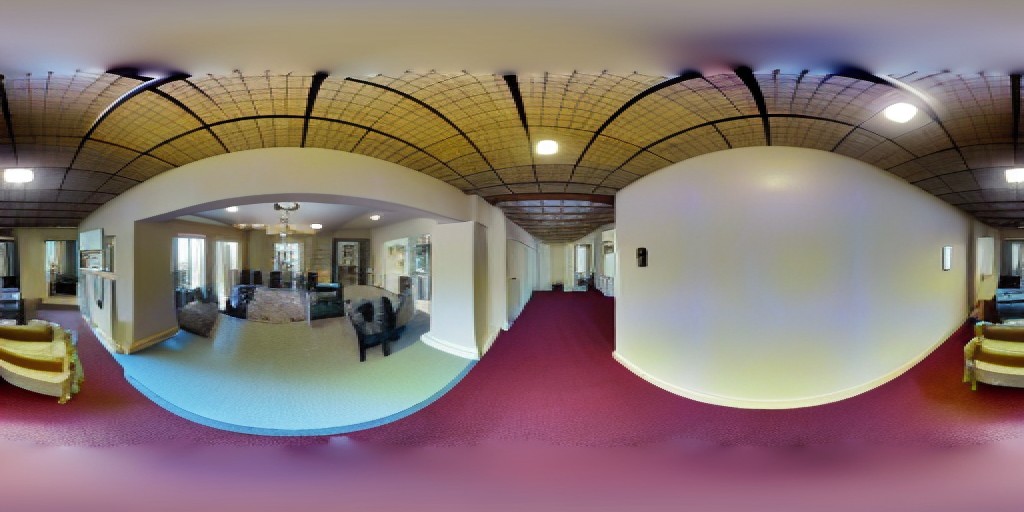}} &  
        {\includegraphics[width=0.212\linewidth]{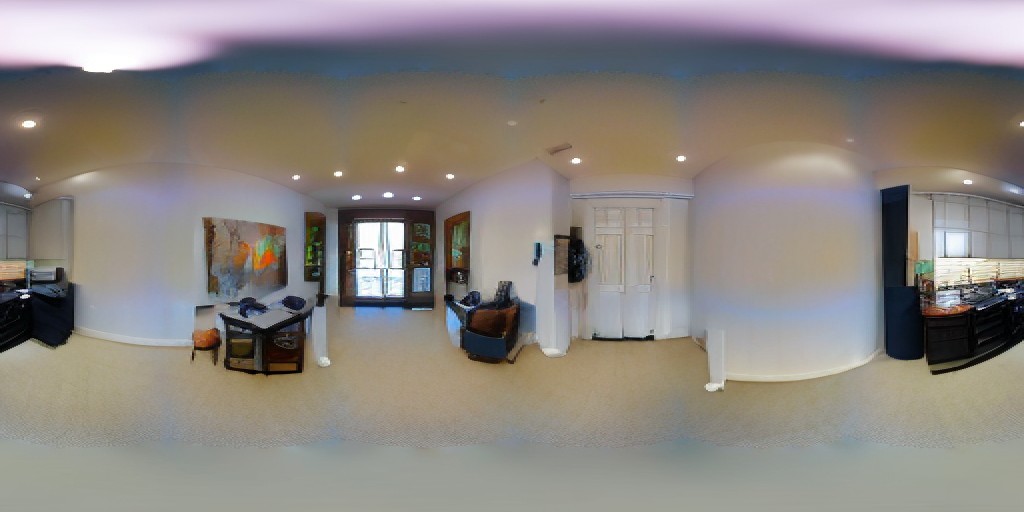}} &  
        {\includegraphics[width=0.212\linewidth]{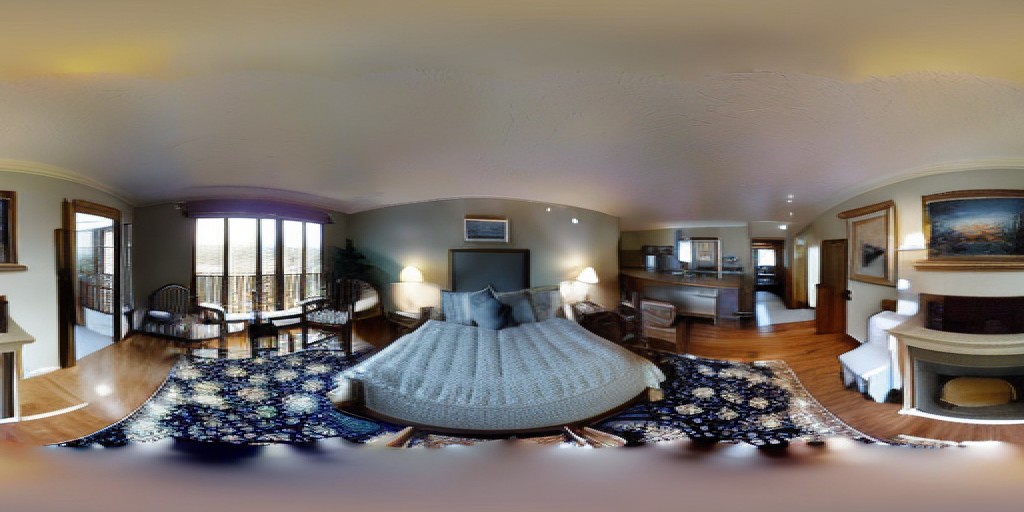}} &  
        {\includegraphics[width=0.212\linewidth]{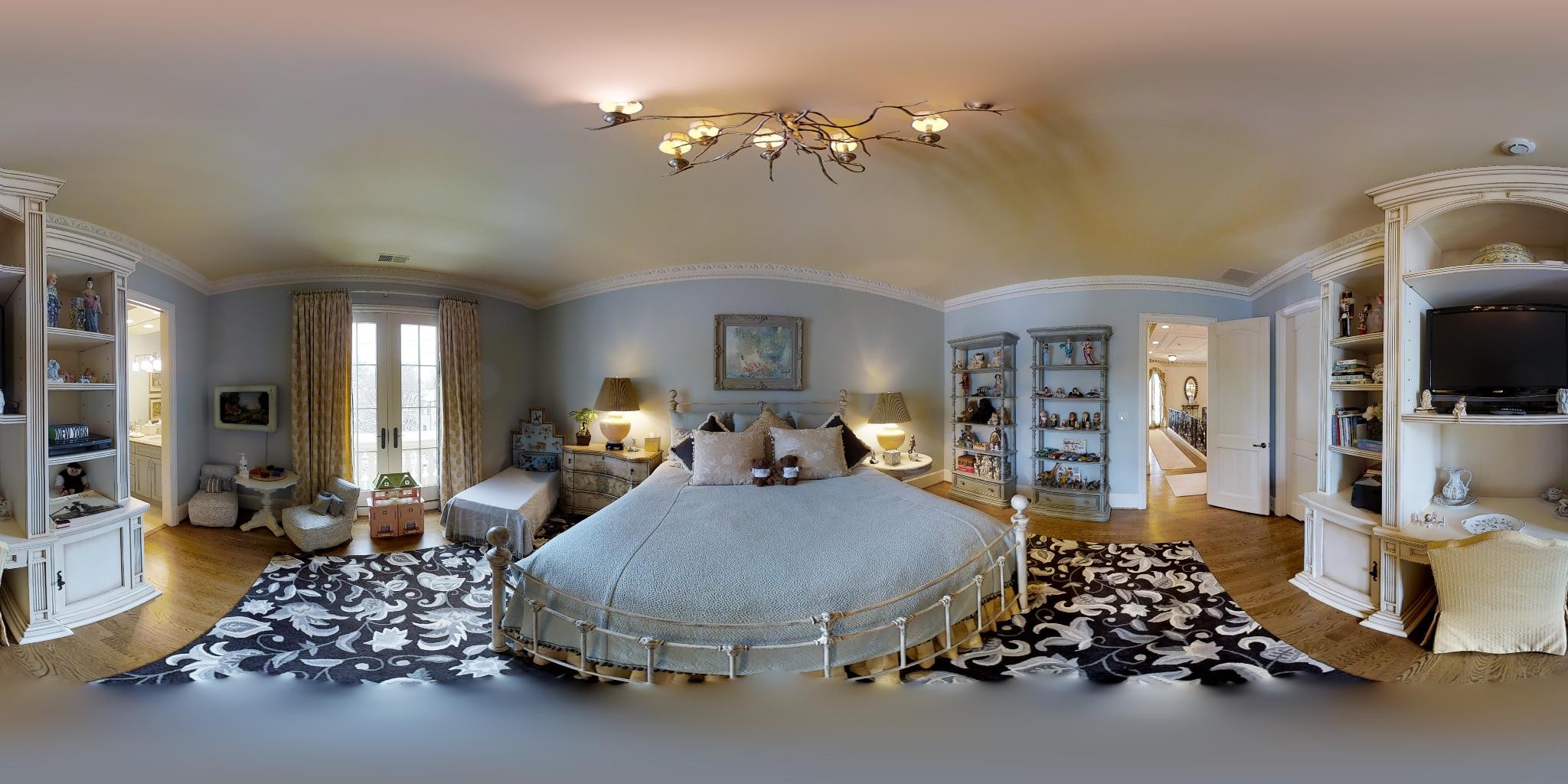}} \\
        {\includegraphics[width=0.106\linewidth]{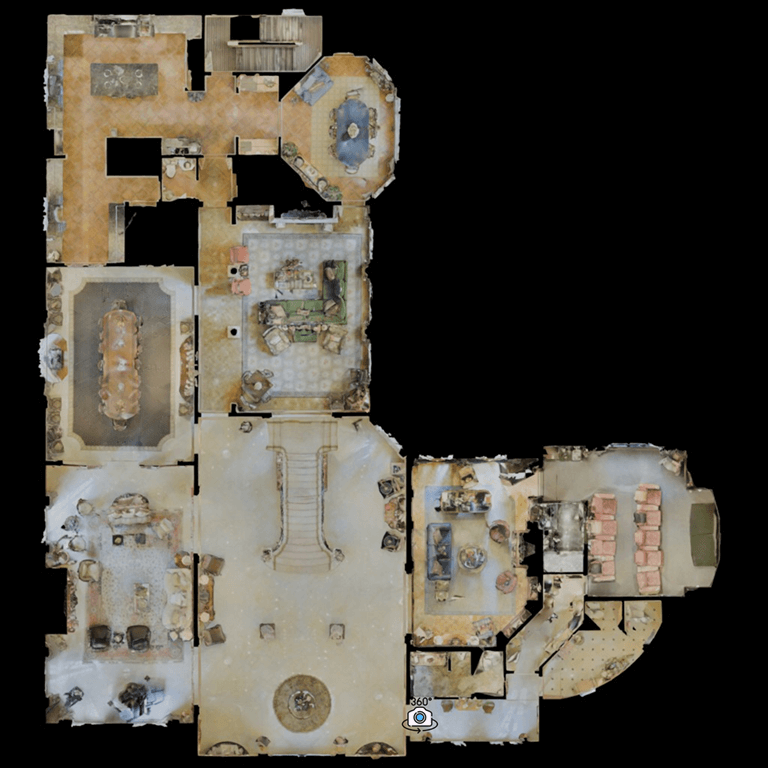}} &  
        {\includegraphics[width=0.212\linewidth]{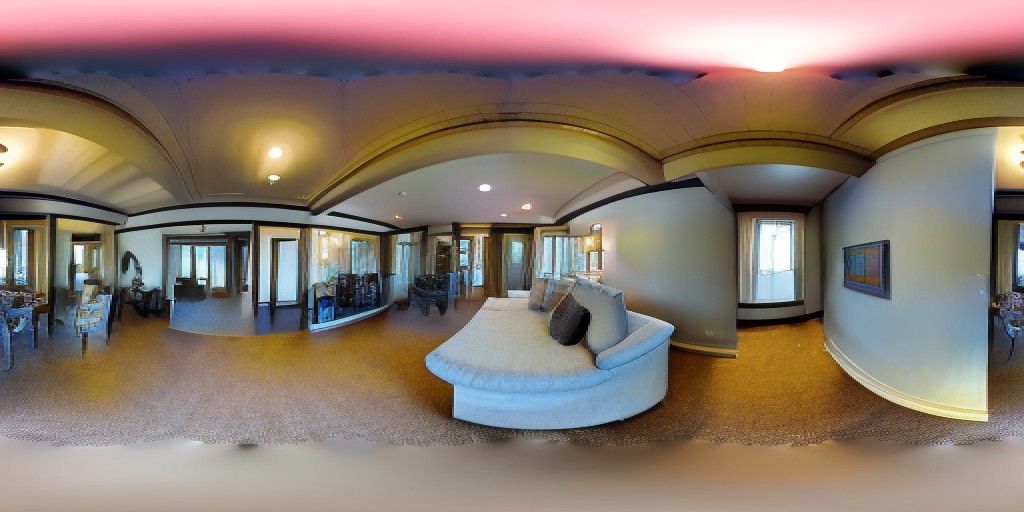}} &  
        {\includegraphics[width=0.212\linewidth]{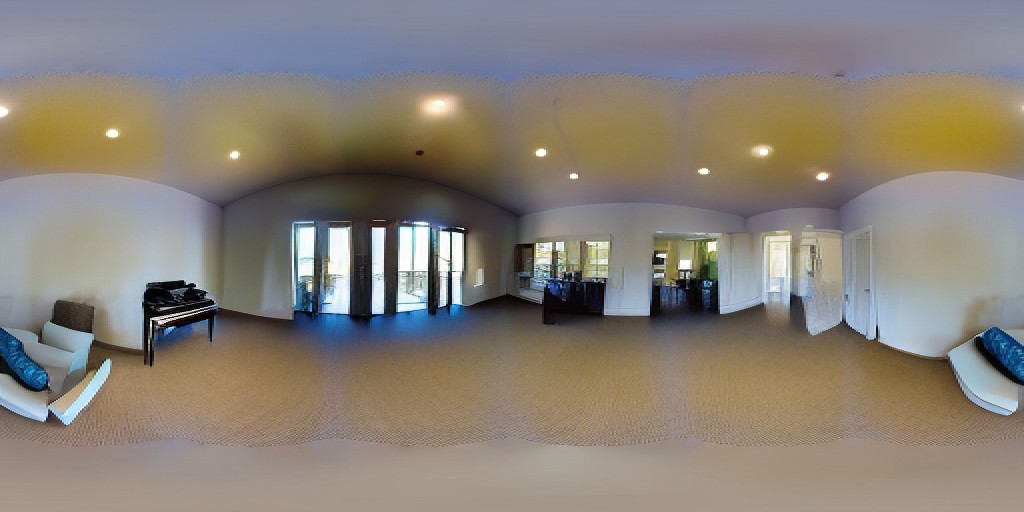}} &  
        {\includegraphics[width=0.212\linewidth]{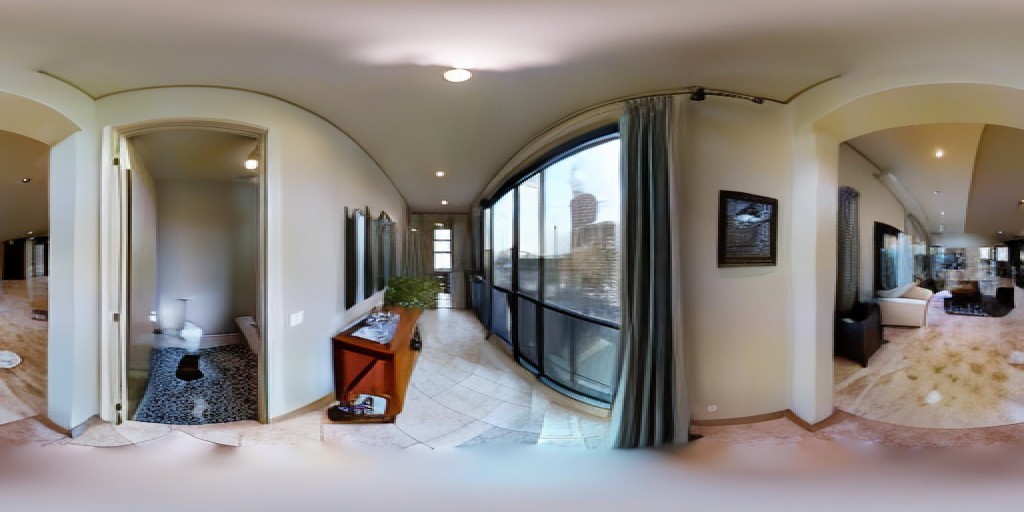}} &  
        {\includegraphics[width=0.212\linewidth]{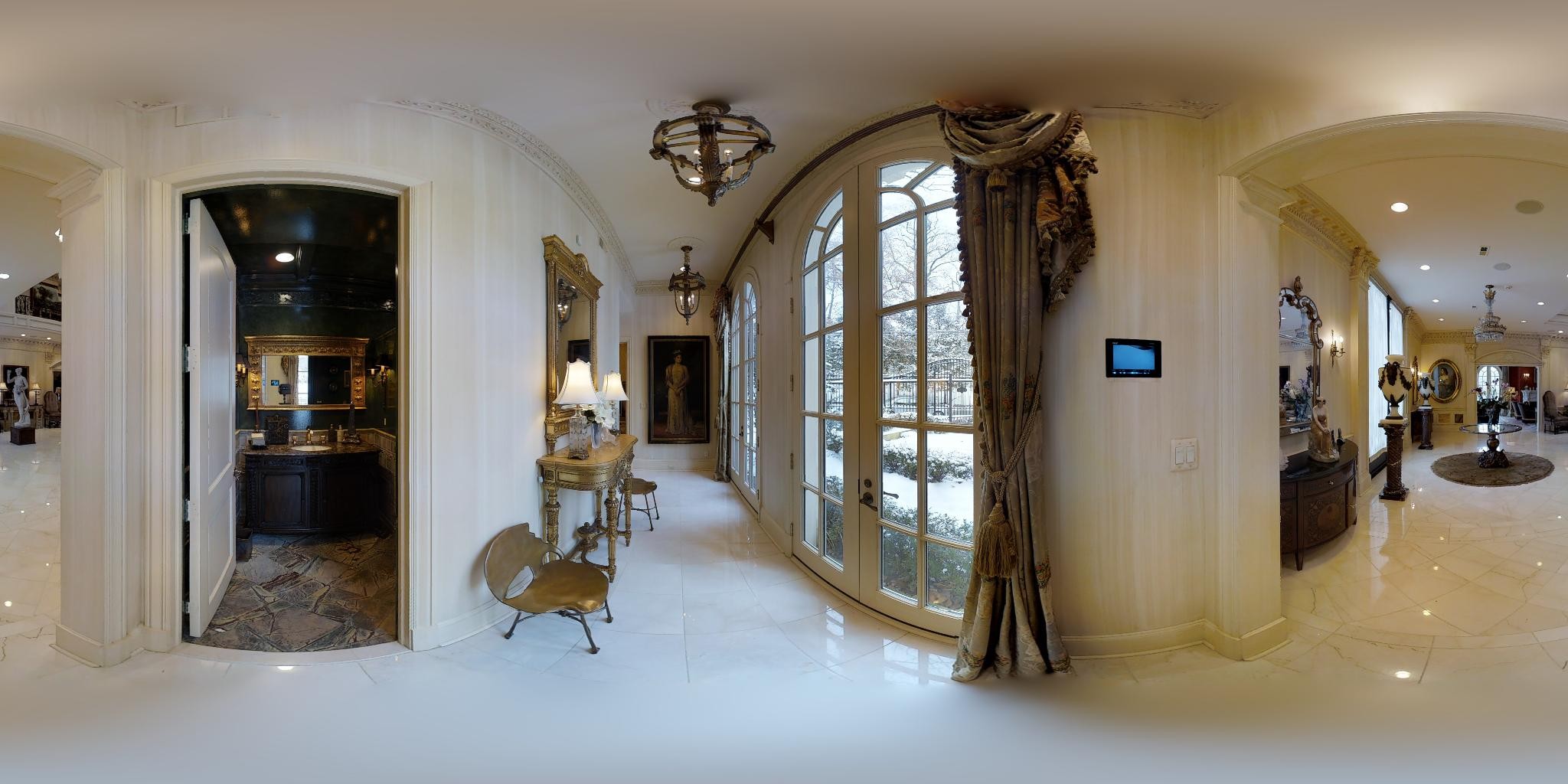}} \\
        {\includegraphics[width=0.106\linewidth]{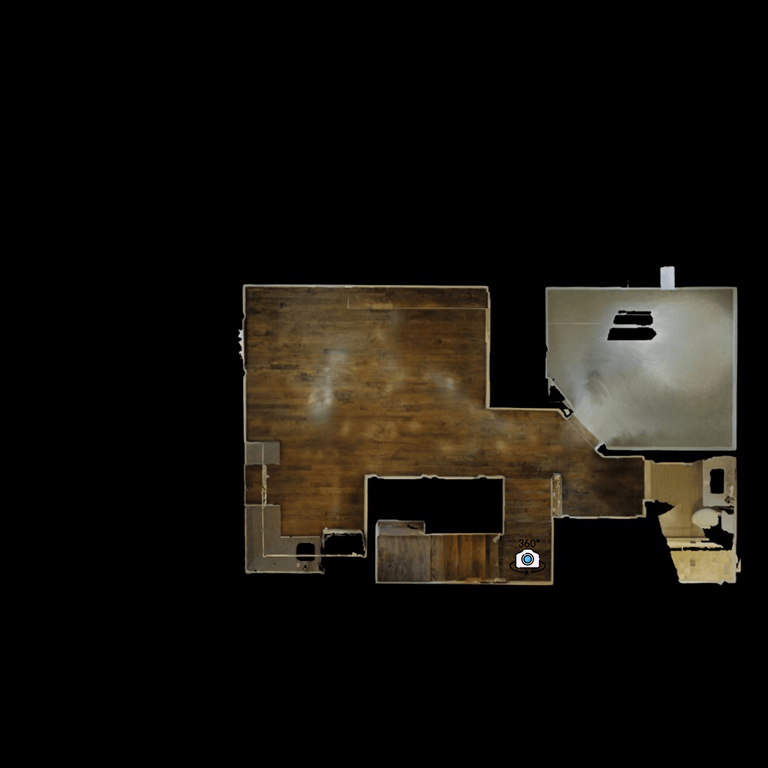}} &  
        {\includegraphics[width=0.212\linewidth]{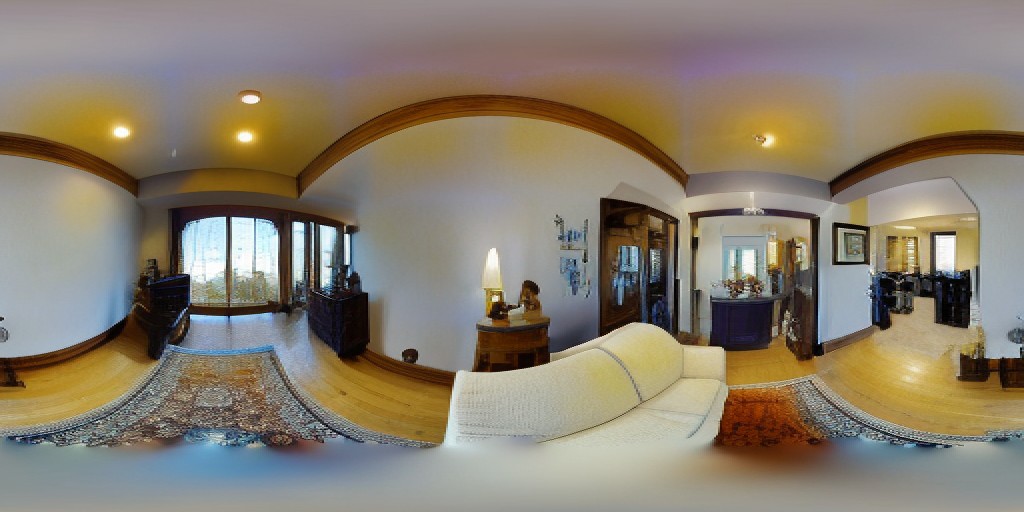}} &  
        {\includegraphics[width=0.212\linewidth]{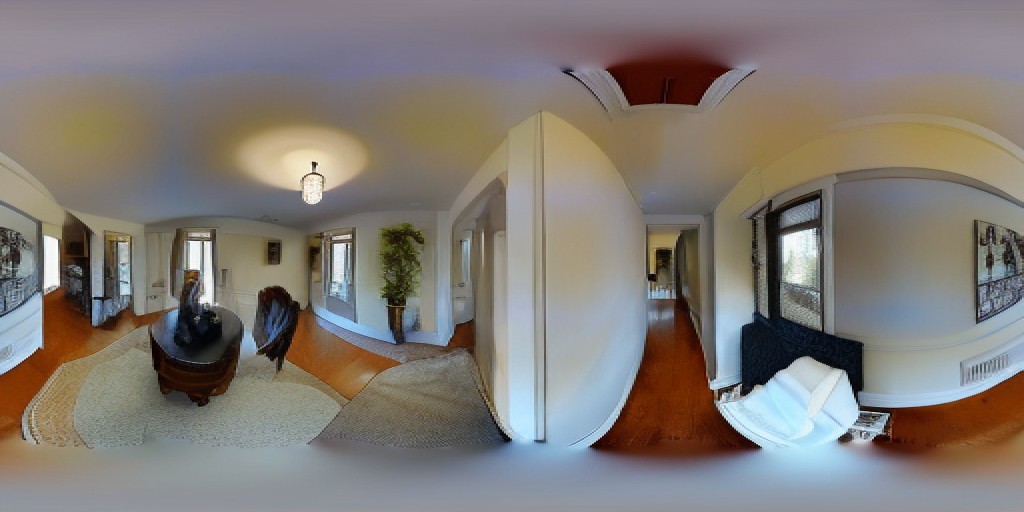}} &  
        {\includegraphics[width=0.212\linewidth]{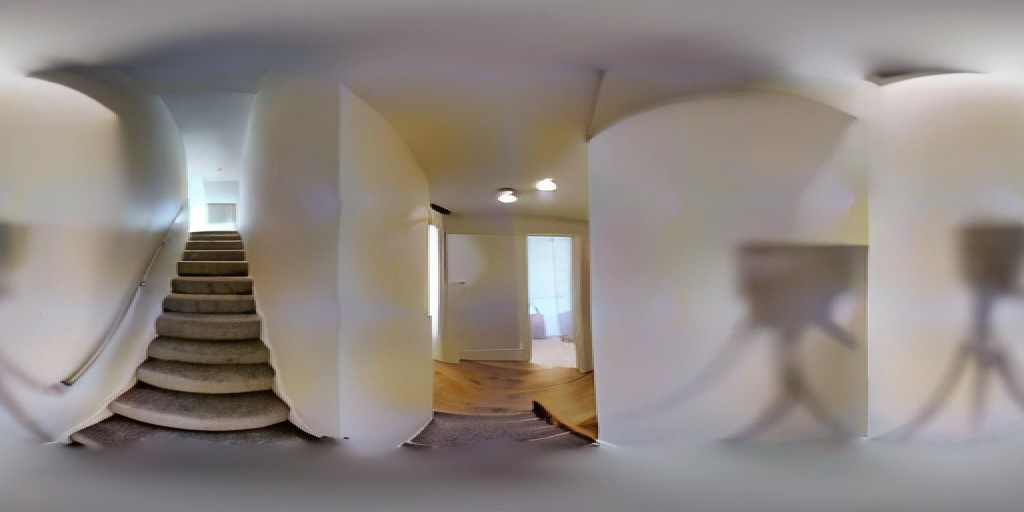}} &  
        {\includegraphics[width=0.212\linewidth]{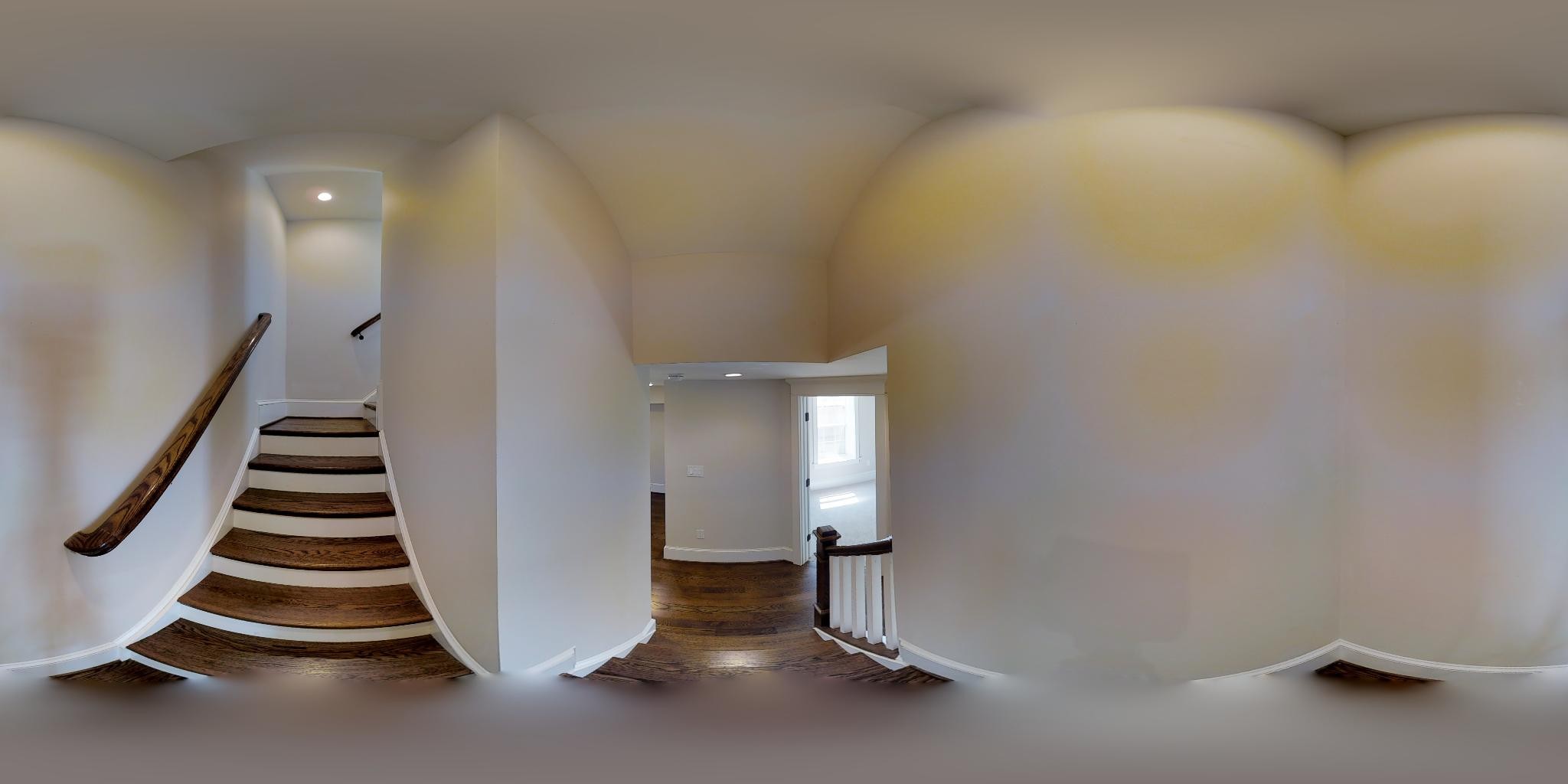}} \\
        {\includegraphics[width=0.106\linewidth]{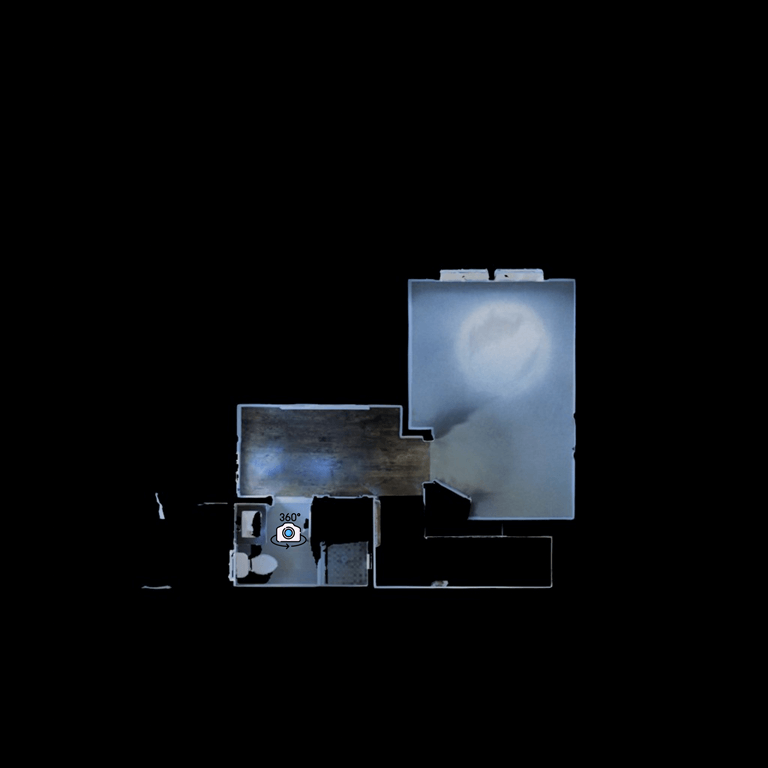}} &  
        {\includegraphics[width=0.212\linewidth]{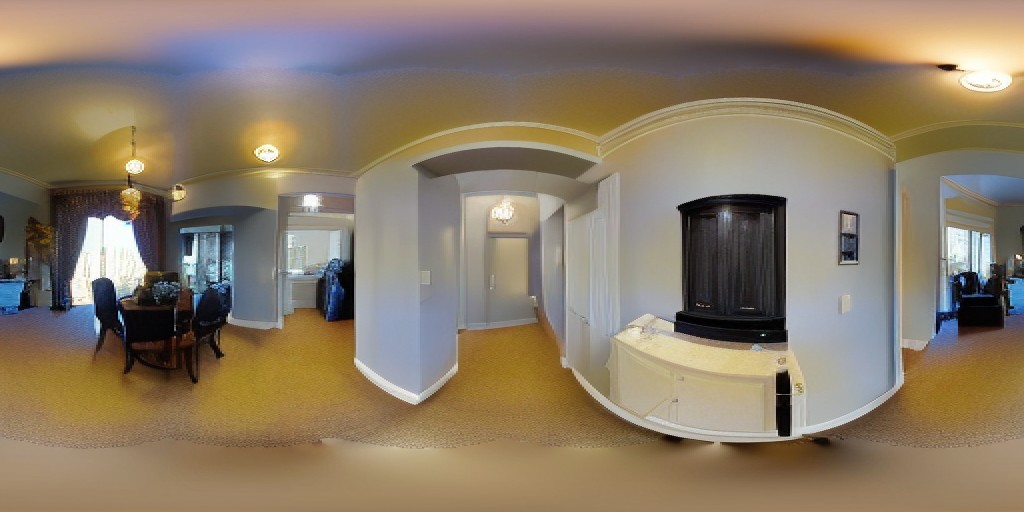}} &  
        {\includegraphics[width=0.212\linewidth]{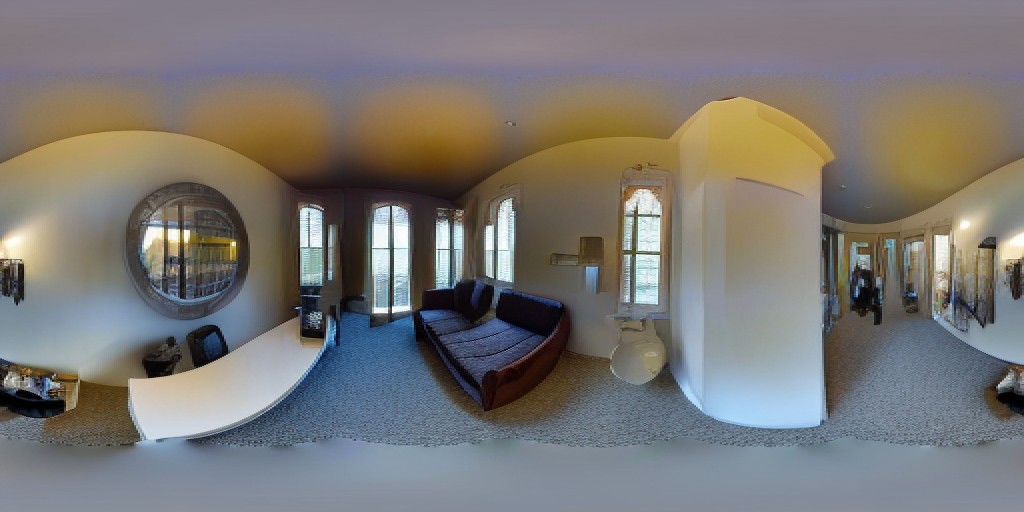}} &  
        {\includegraphics[width=0.212\linewidth]{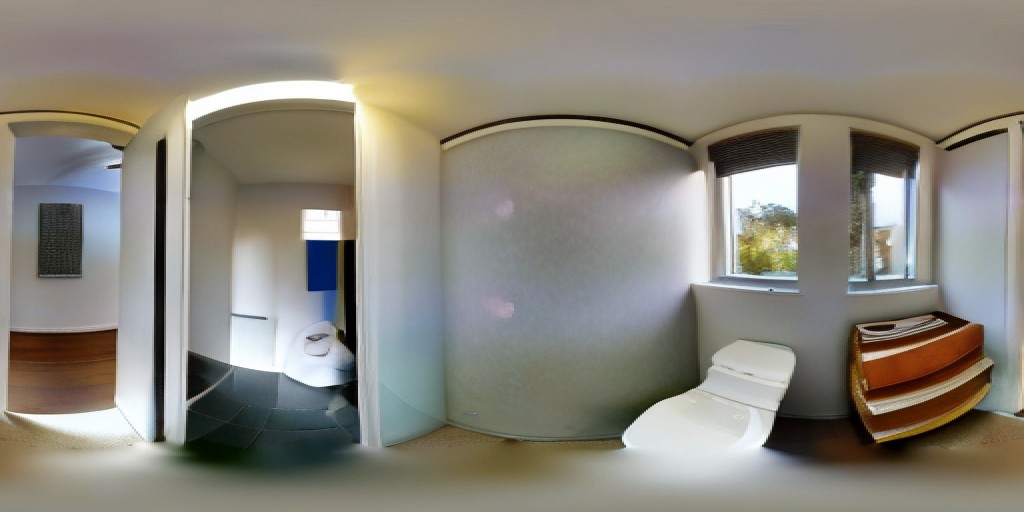}} &  
        {\includegraphics[width=0.212\linewidth]{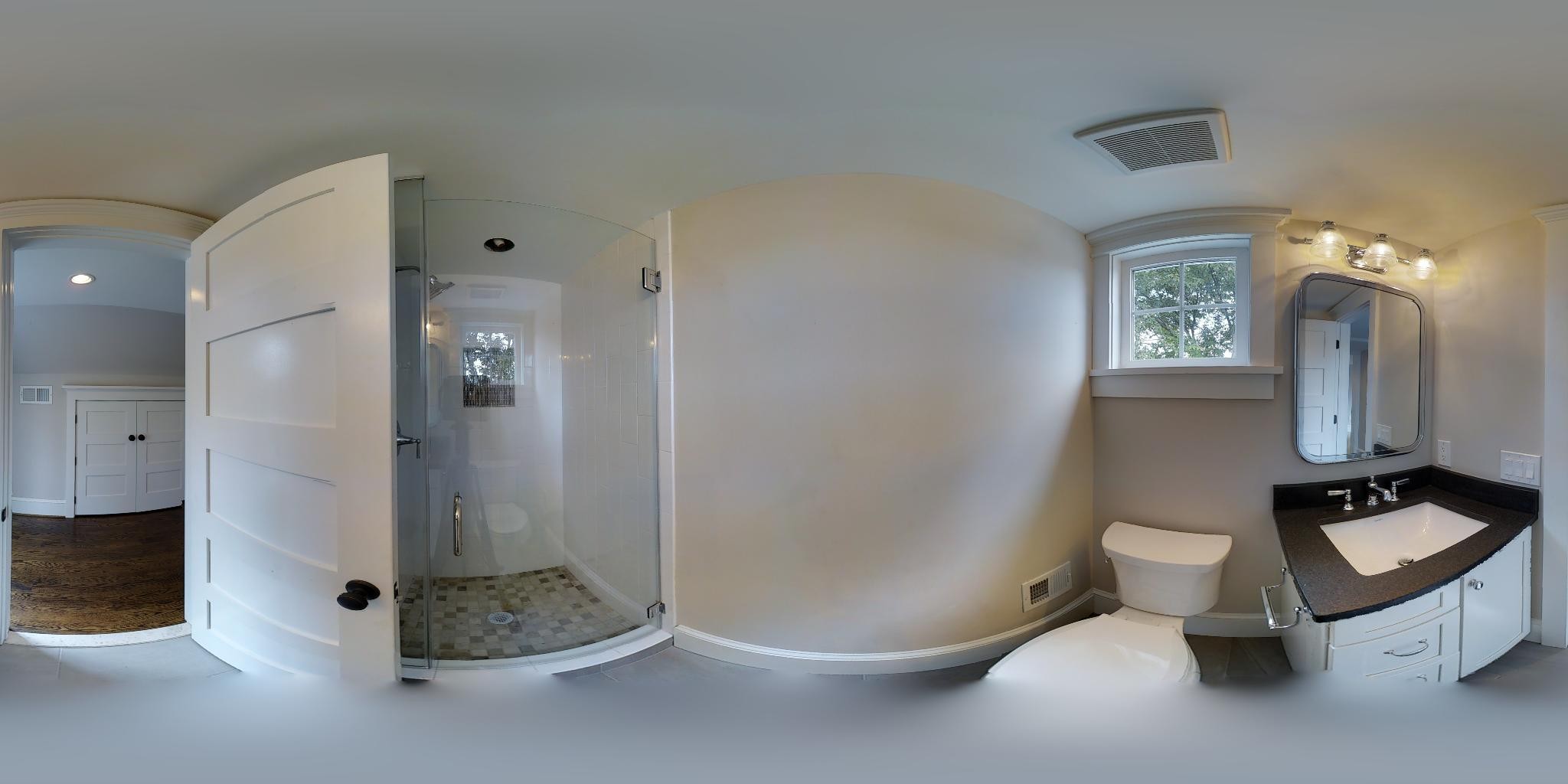}} \\
        (a) Top-down & (b) Sat2Density~\cite{10377305} & (c) PanFusion~\cite{panfusion2024} & (d) Ours & (e) Ground Truth\\ 
    \end{tabular}
    \vspace{-2mm}
    \caption{Qualitative comparisons on the Matterport3D dataset.}
    \label{fig:Matterport_Baseline}
\end{figure*}

\begin{figure*}[ht]
    \centering
    \footnotesize
    \setlength{\tabcolsep}{1.5pt}
    \begin{tabular}{ccccc}
    \centering
        {\includegraphics[width=0.106\linewidth]{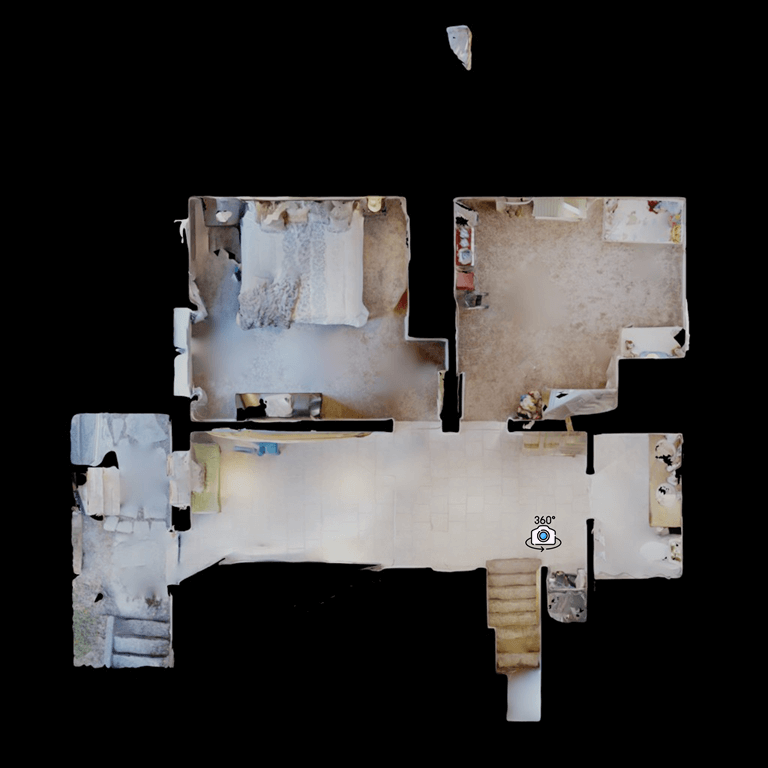}} &  
        {\includegraphics[width=0.212\linewidth]{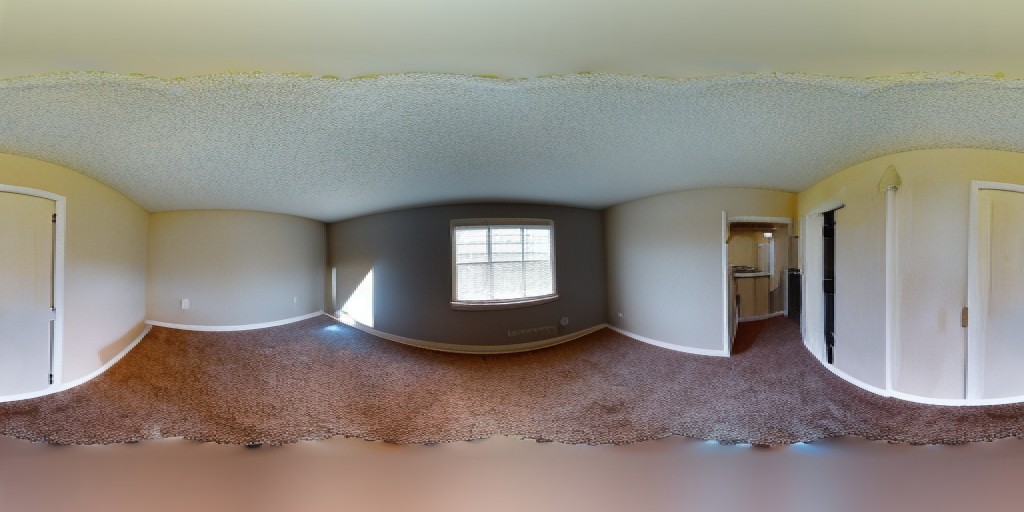}} &  
        {\includegraphics[width=0.212\linewidth]{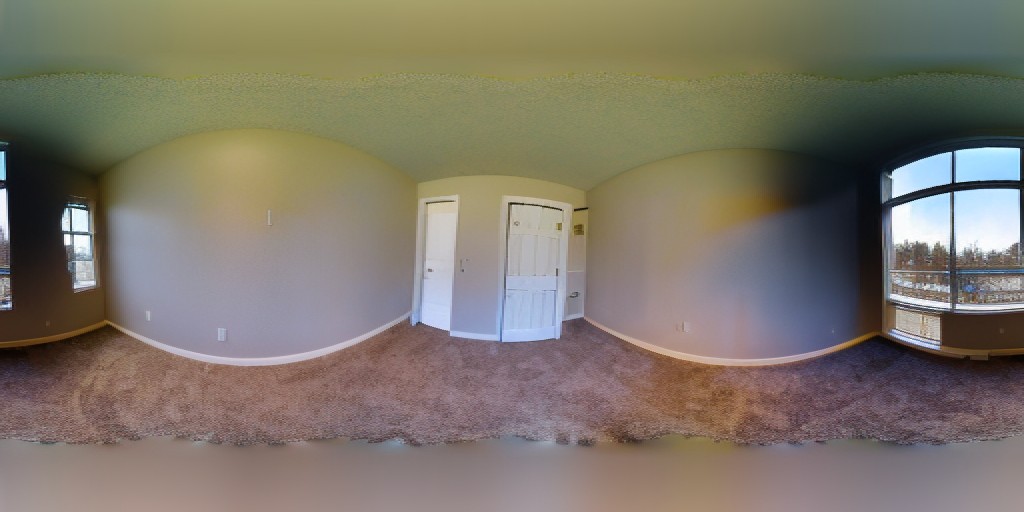}} &  
        {\includegraphics[width=0.212\linewidth]{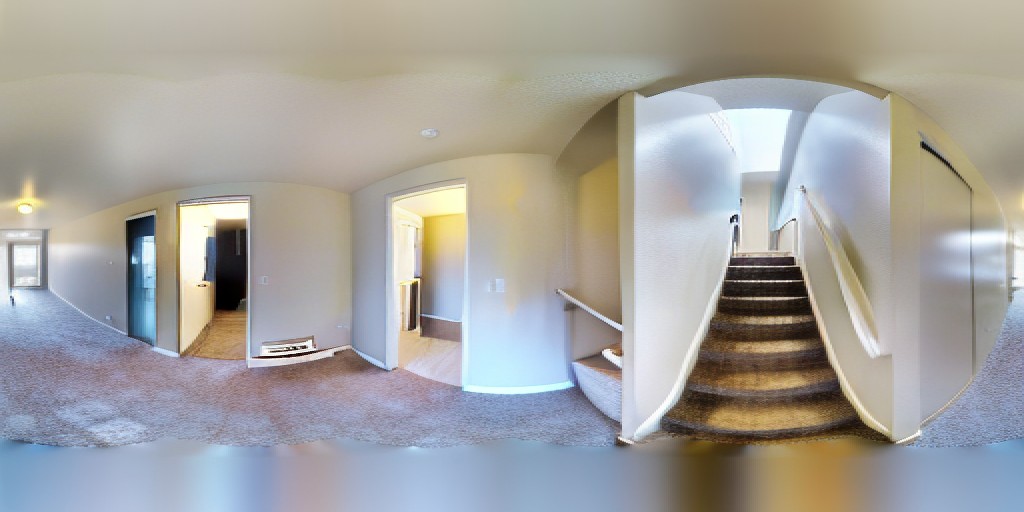}} &  
        {\includegraphics[width=0.212\linewidth]{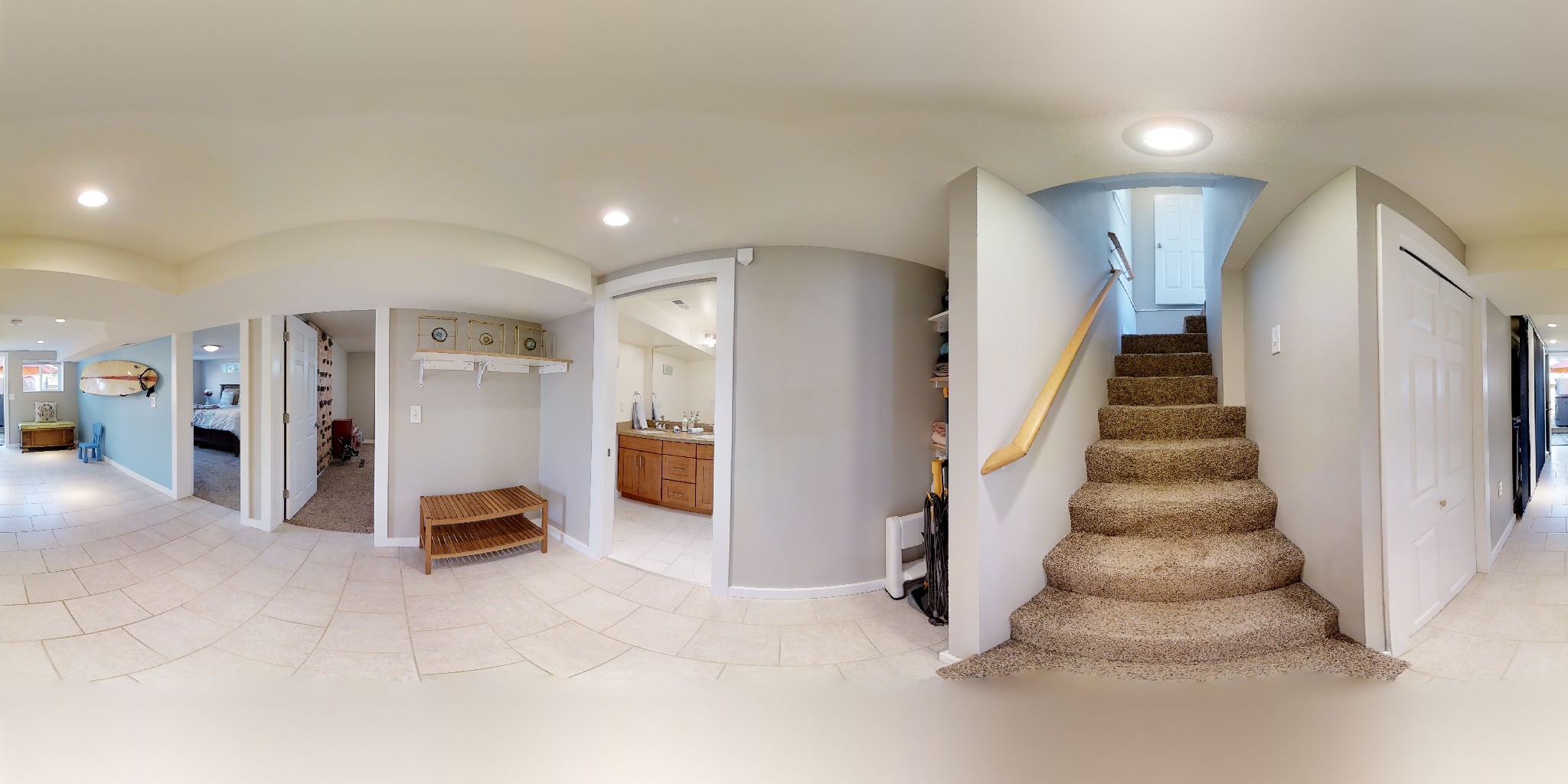}} \\
        {\includegraphics[width=0.106\linewidth]{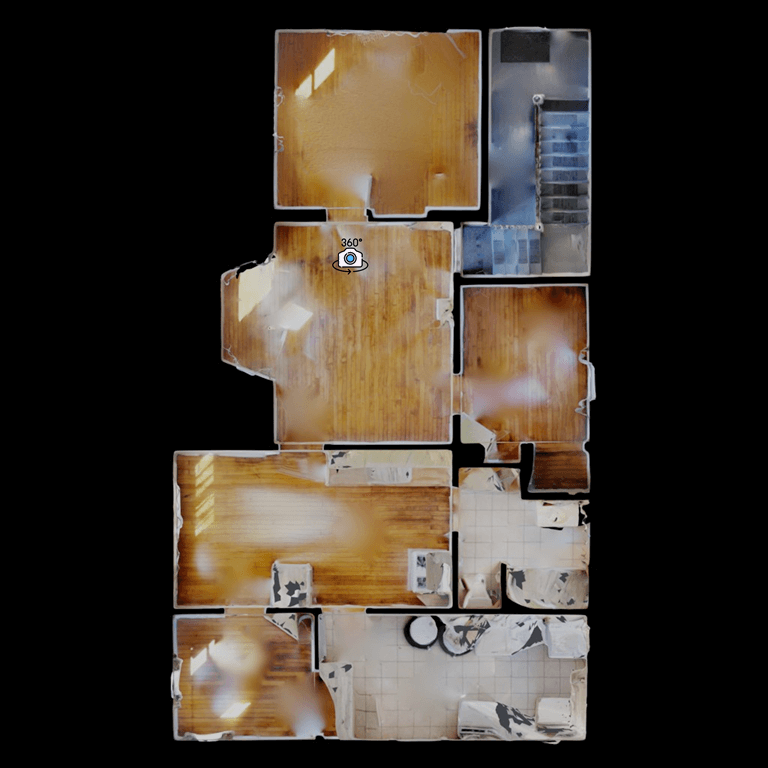}} &  
        {\includegraphics[width=0.212\linewidth]{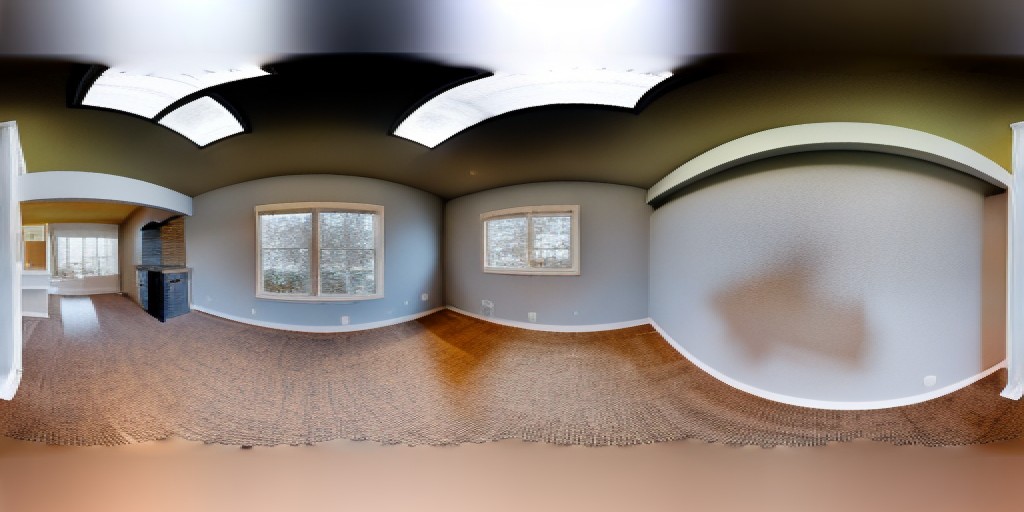}} &  
        {\includegraphics[width=0.212\linewidth]{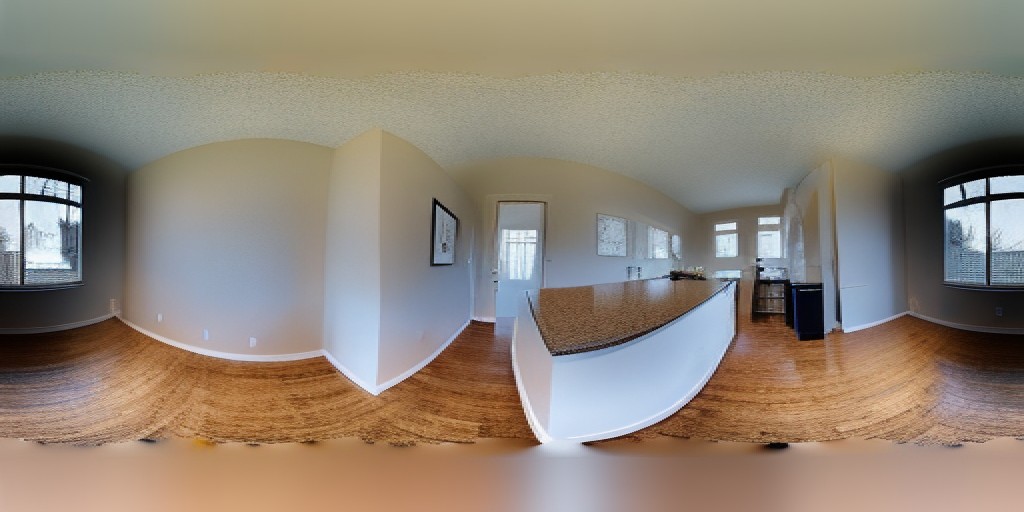}} &  
        {\includegraphics[width=0.212\linewidth]{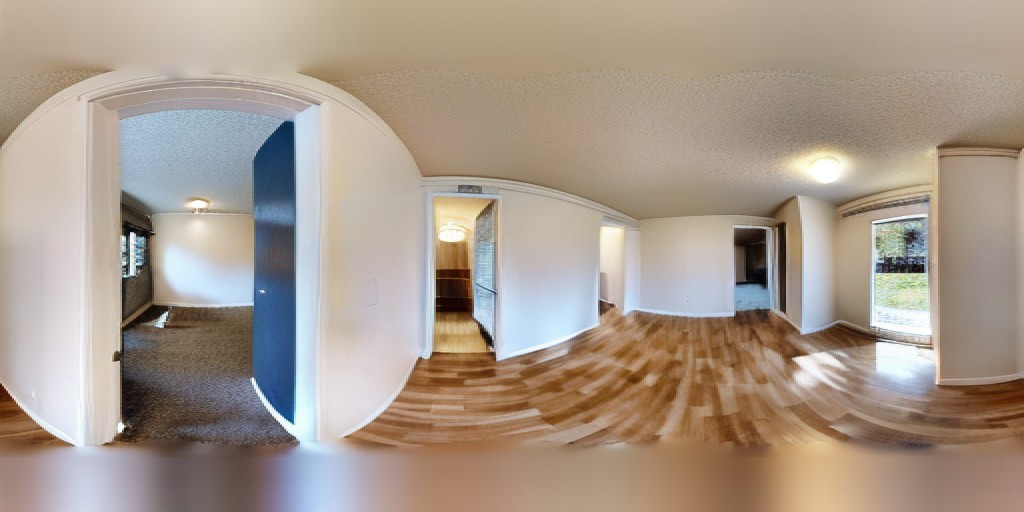}} &  
        {\includegraphics[width=0.212\linewidth]{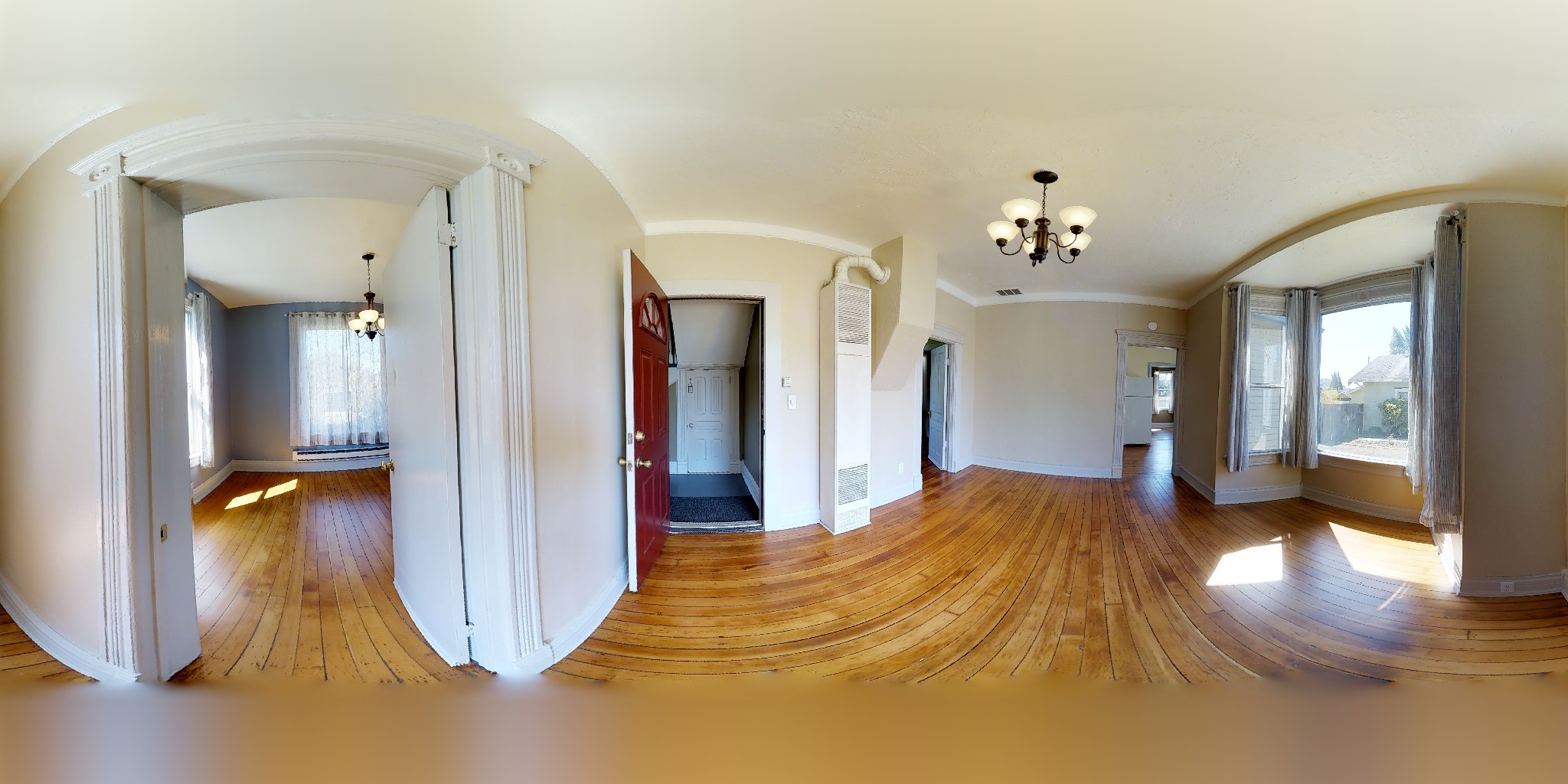}} \\
        {\includegraphics[width=0.106\linewidth]{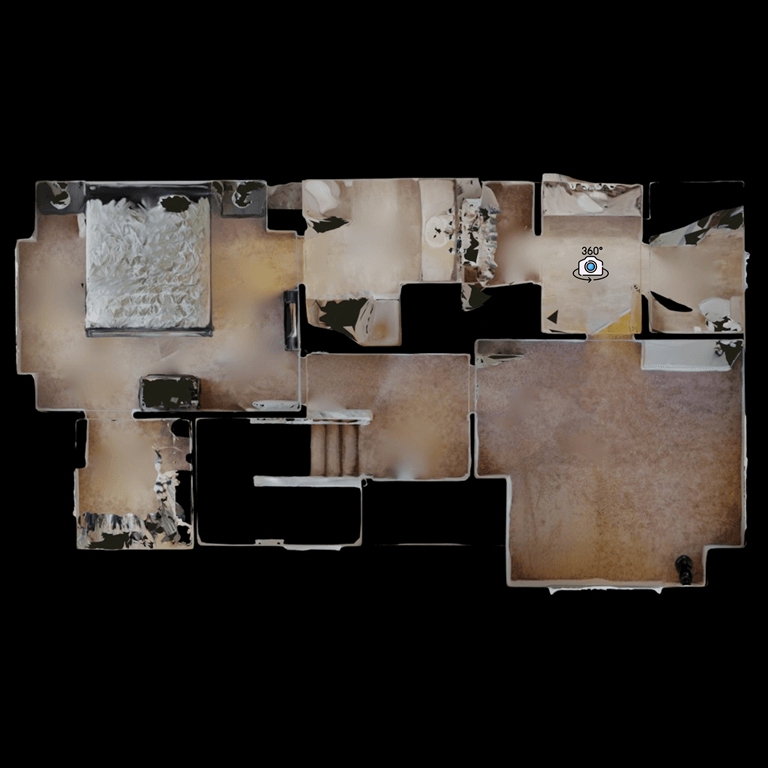}} &  
        {\includegraphics[width=0.212\linewidth]{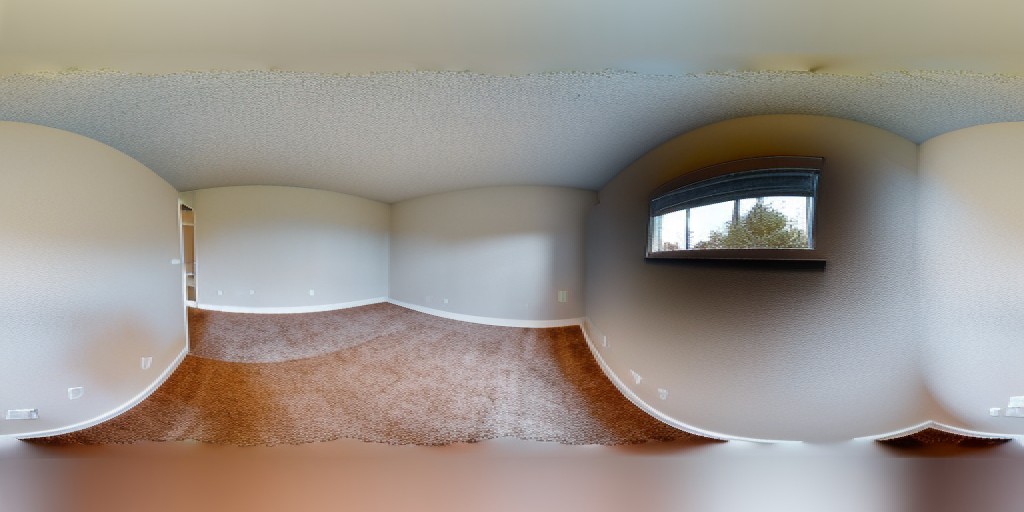}} &  
        {\includegraphics[width=0.212\linewidth]{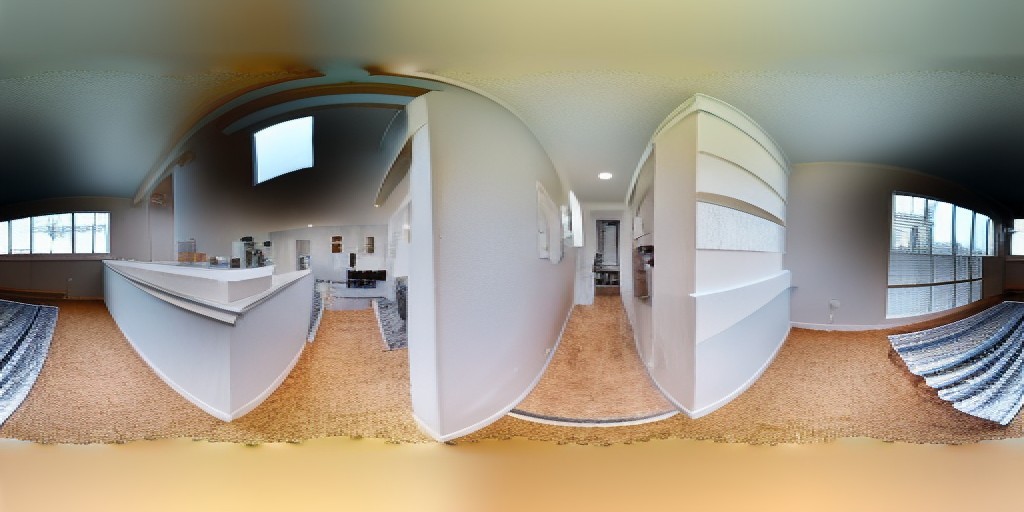}} &  
        {\includegraphics[width=0.212\linewidth]{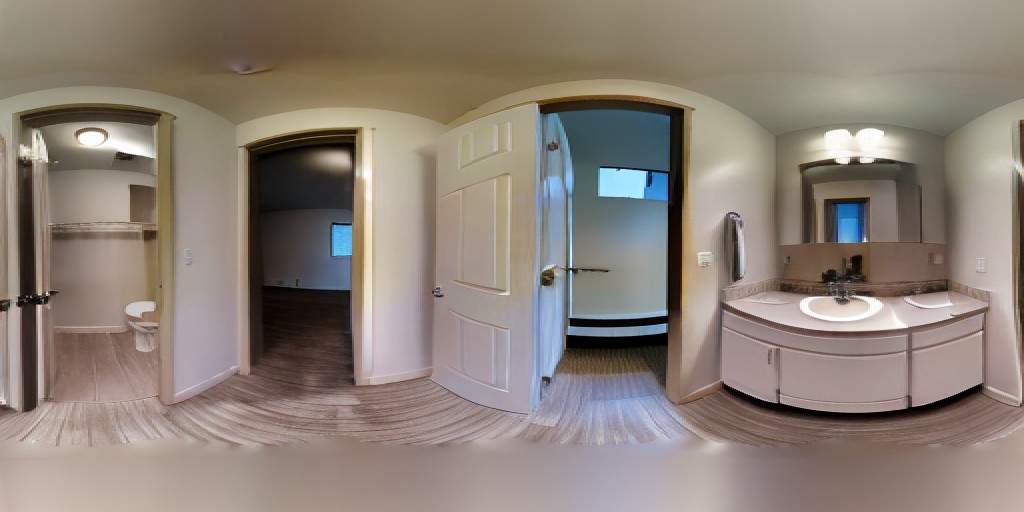}} &  
        {\includegraphics[width=0.212\linewidth]{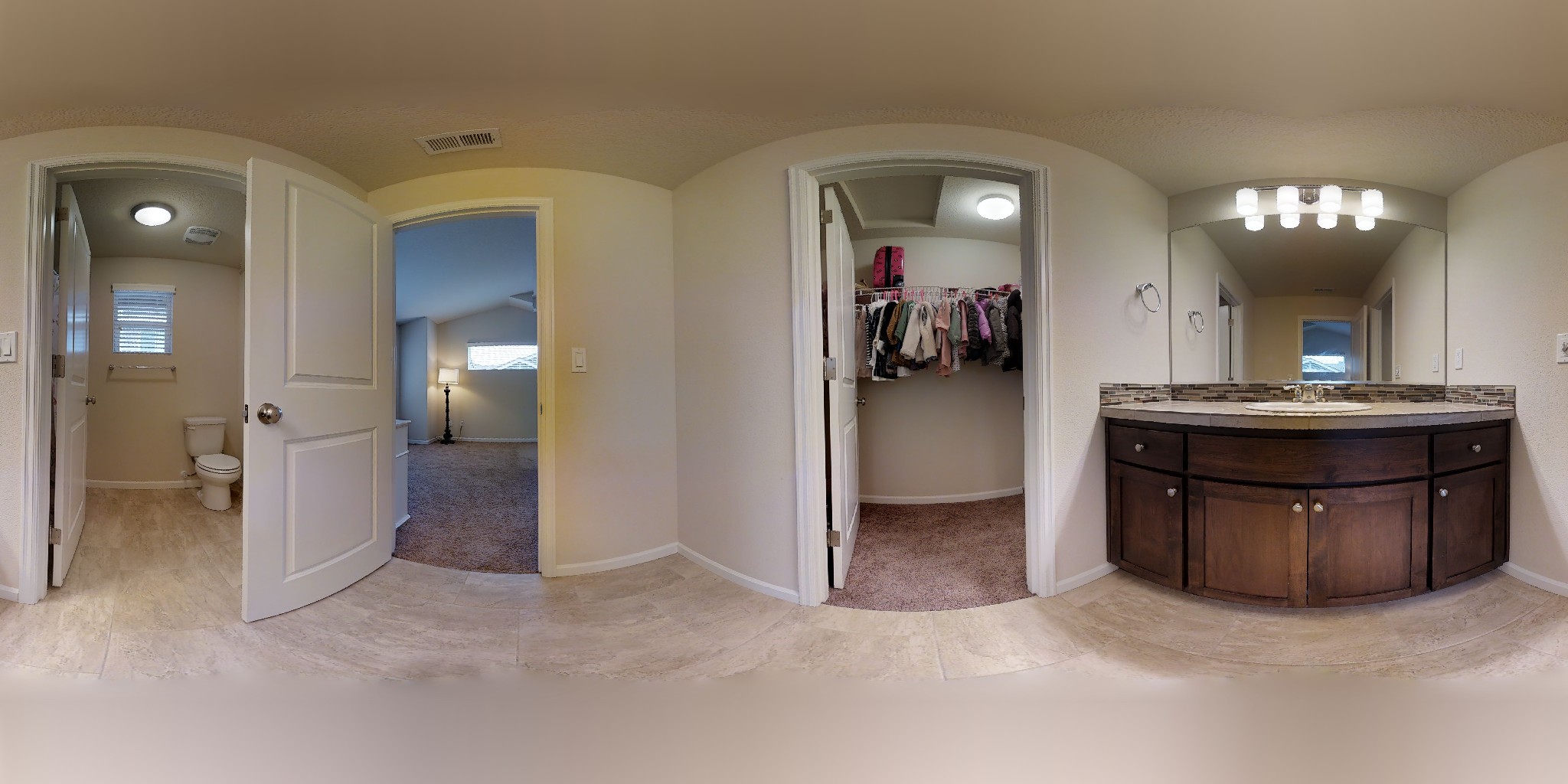}} \\
        {\includegraphics[width=0.106\linewidth]{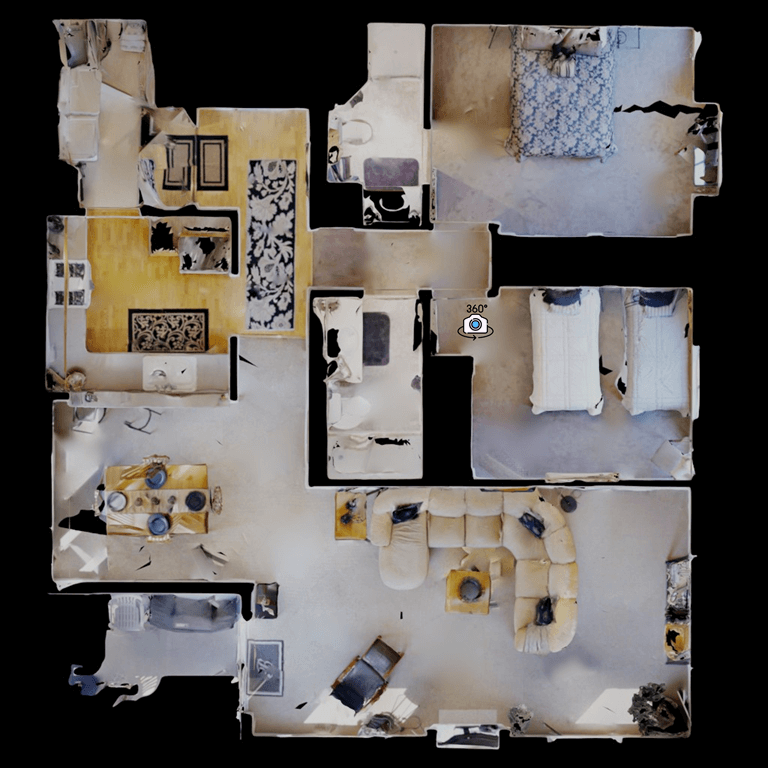}} &  
        {\includegraphics[width=0.212\linewidth]{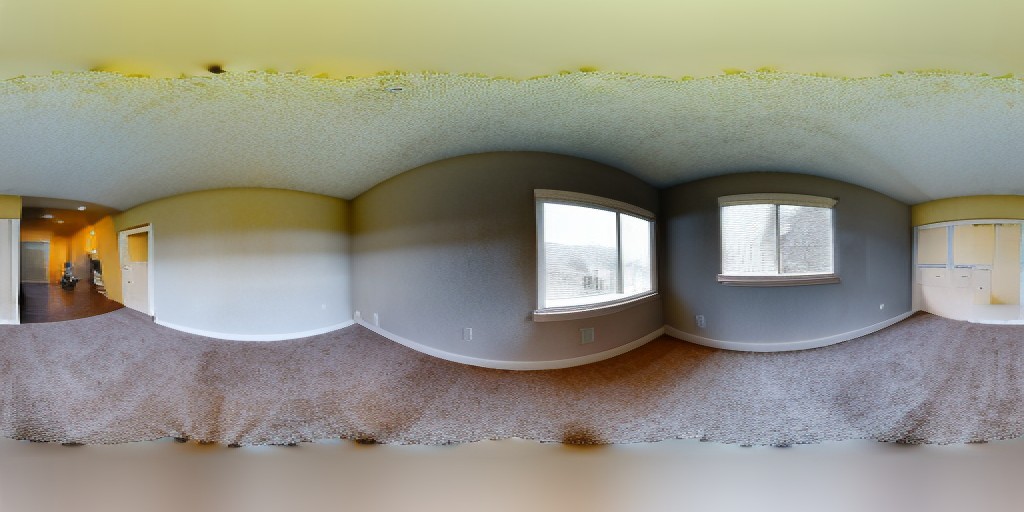}} &  
        {\includegraphics[width=0.212\linewidth]{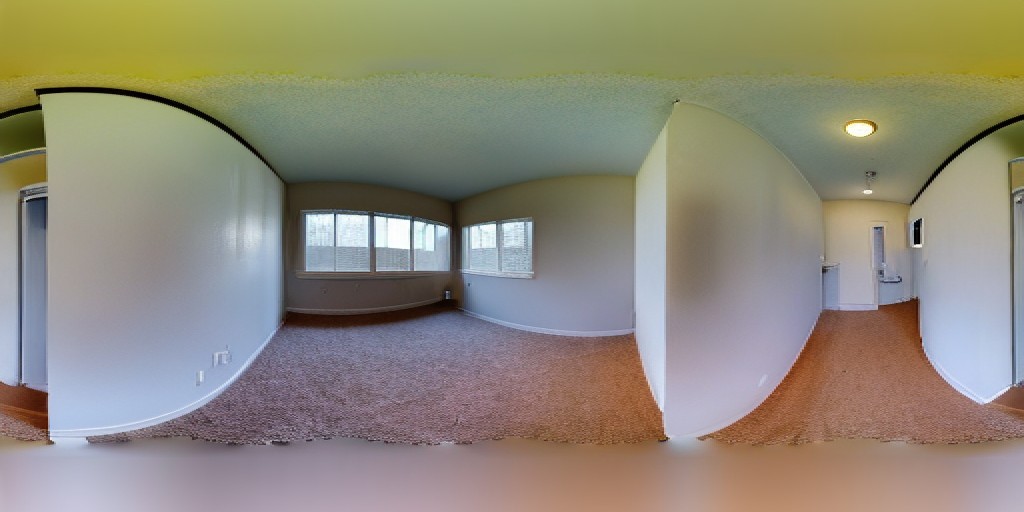}} &  
        {\includegraphics[width=0.212\linewidth]{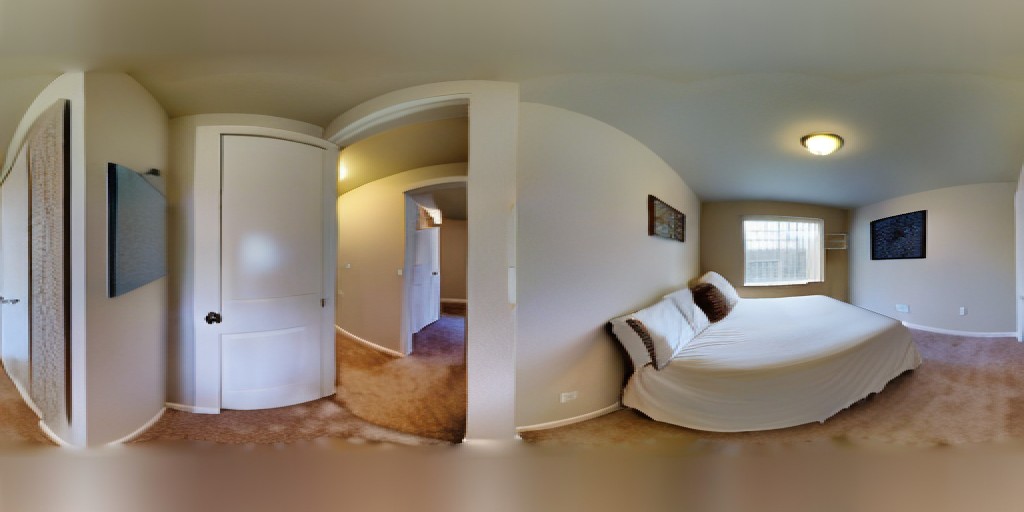}} &  
        {\includegraphics[width=0.212\linewidth]{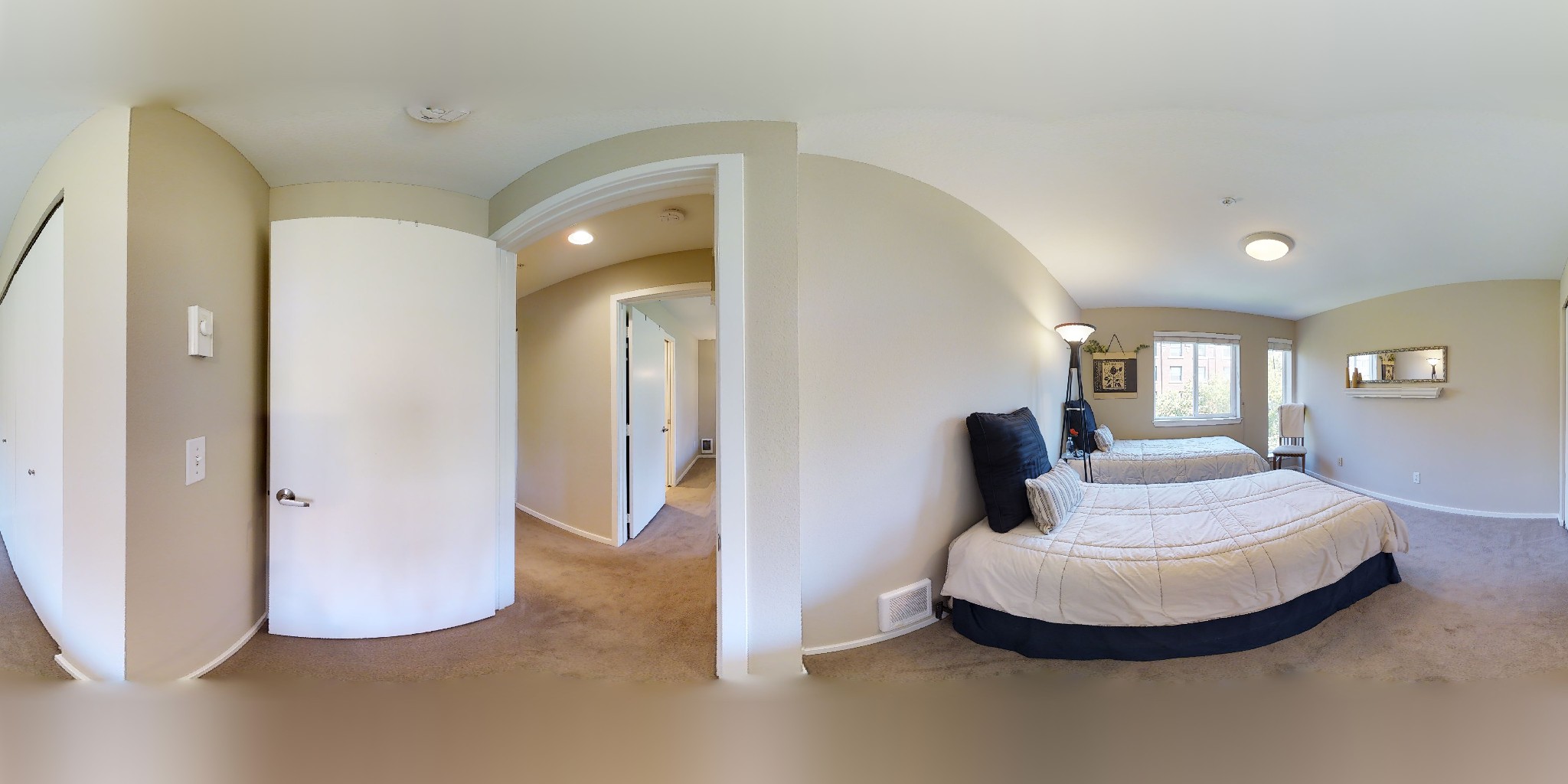}} \\
        {\includegraphics[width=0.106\linewidth]{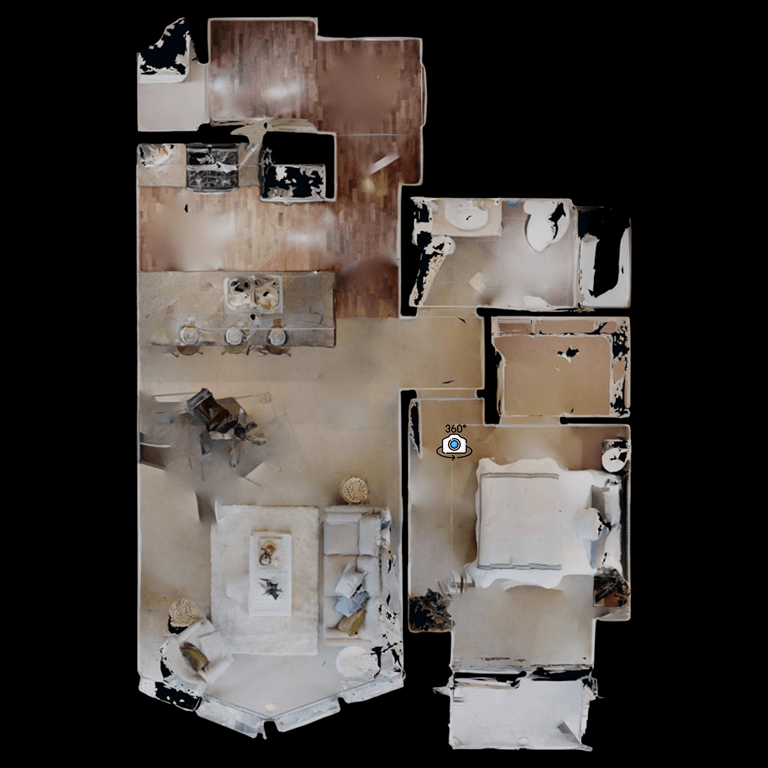}} &  
        {\includegraphics[width=0.212\linewidth]{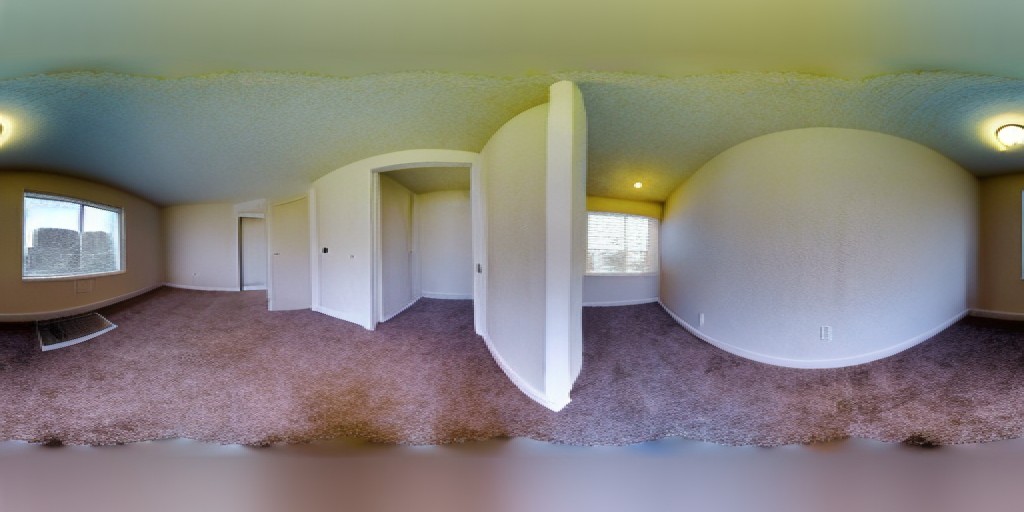}} &  
        {\includegraphics[width=0.212\linewidth]{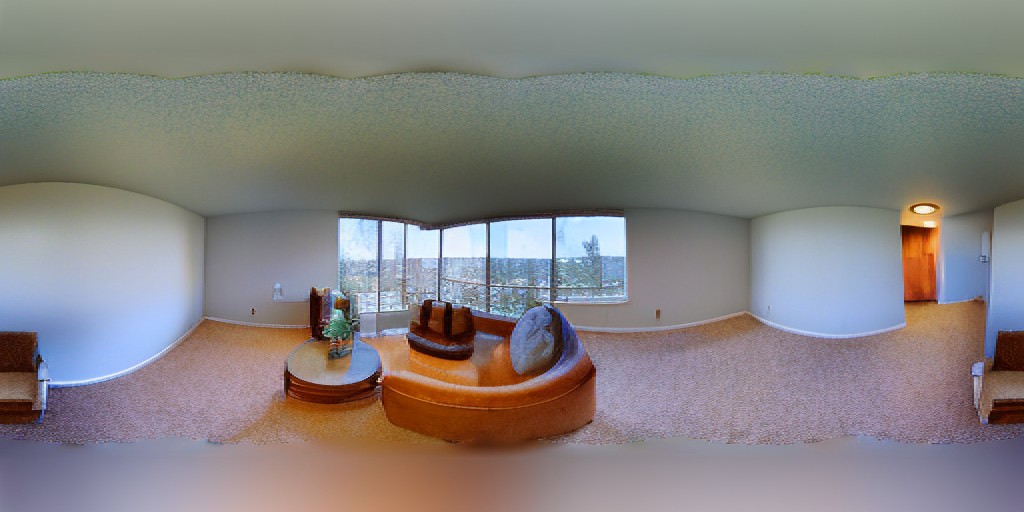}} &  
        {\includegraphics[width=0.212\linewidth]{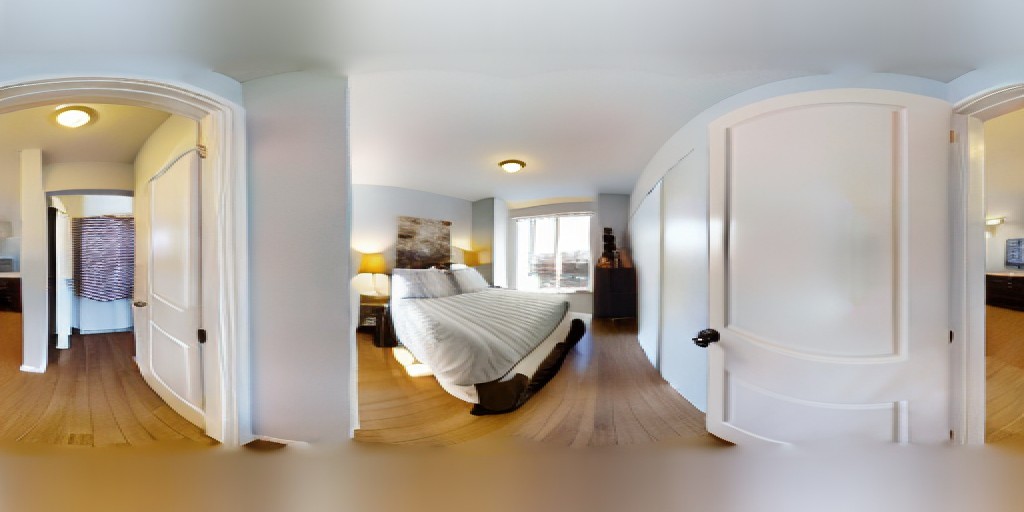}} &  
        {\includegraphics[width=0.212\linewidth]{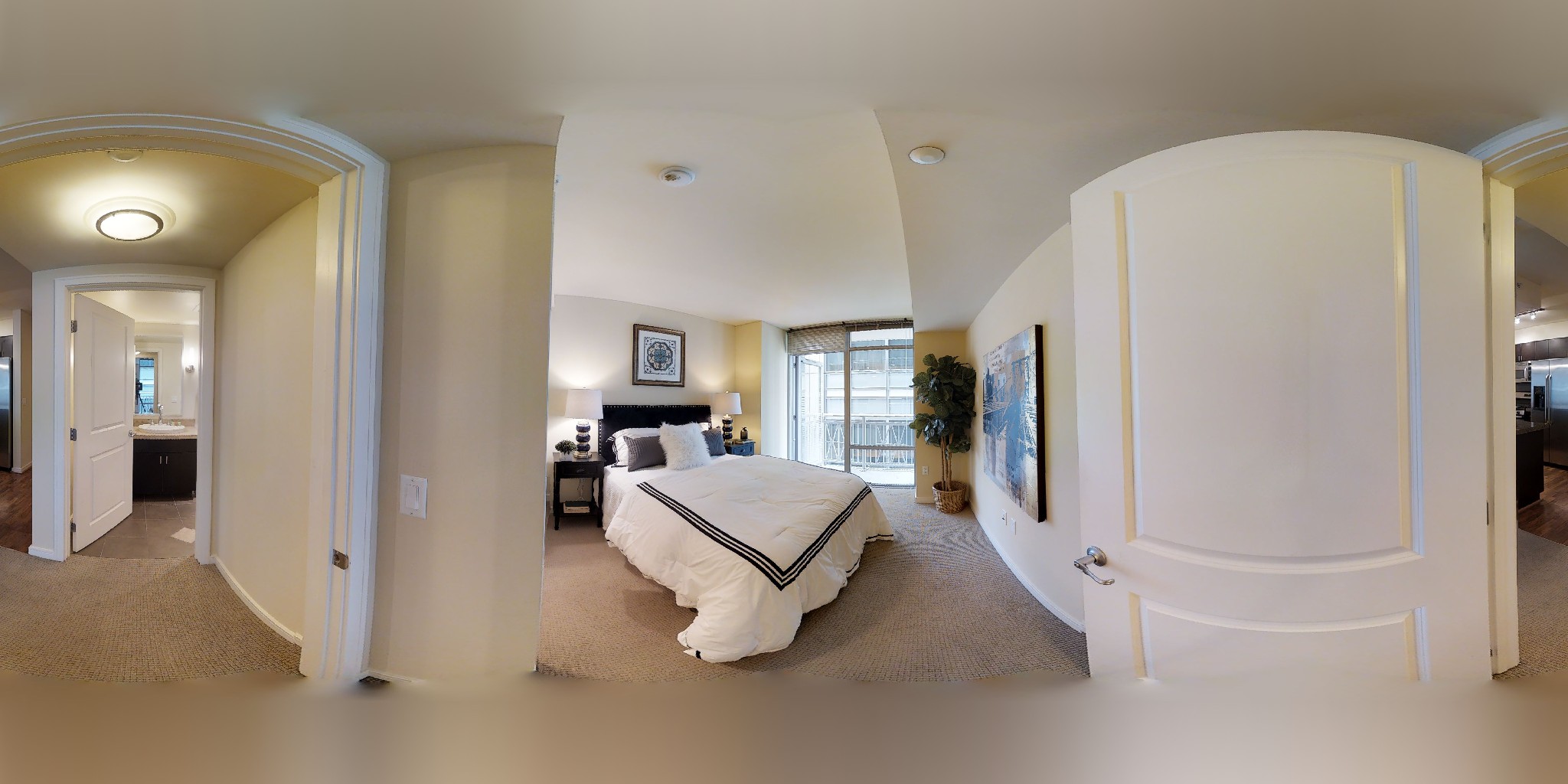}} \\      
        (a) Top-down & (b) Sat2Density~\cite{10377305} & (c) PanFusion~\cite{panfusion2024} & (d) Ours & (e) Ground Truth\\ 
    \end{tabular}
    \vspace{-2mm}
    \caption{Qualitative comparisons on the Gibson dataset.}
    \label{fig:Gibson_Baseline}
\end{figure*}

As shown in Table~\ref{tab:results_baseline}, we compare our method against three baseline approaches across four evaluation metrics. Sat2Density~\cite{10377305} is a satellite-to-ground panorama synthesis method, which we adapt for indoor panorama generation using latent diffusion model (LDM)~\cite{rombach2022high} and ControlNet~\cite{zhang2023adding}. PanFusion~\cite{panfusion2024} is a text-to-panorama generation framework that also incorporates layout-conditioned generation via ControlNet.  
Table~\ref{tab:results_baseline} shows that our method outperforms all baselines across all four metrics on both datasets, demonstrating its effectiveness. Furthermore, qualitative comparisons in Figures~\ref{fig:Matterport_Baseline} and~\ref{fig:Gibson_Baseline} highlight that our approach generates more realistic and structurally accurate house reconstructions, including furniture placement.

\begin{table}[t!]
    \centering
    \setlength{\tabcolsep}{2pt}
    \resizebox{0.99\linewidth}{!}{
    \begin{tabular}{ccccc|cccc|cccc}
        \toprule
        \multicolumn{5}{c}{\textbf{Modules}} & \multicolumn{4}{c}{\textbf{Matterport3D}} & \multicolumn{4}{c}{\textbf{Gibson}} \\
        \cmidrule(lr){1-5} \cmidrule(lr){6-9} \cmidrule(lr){10-13}
         seg & floor & wall & depth & color 
                       & PSNR$\uparrow$ & SSIM$\uparrow$ & FID$\downarrow$ & LPIPS$\downarrow$
                       & PSNR$\uparrow$ & SSIM$\uparrow$ & FID$\downarrow$ & LPIPS$\downarrow$\\
        \midrule
         
             &  &  & \checkmark & \checkmark 
            & 11.08 & 0.4122 & 104.03 & 0.6463 
            & 11.07 & 0.4551 & 117.08 & 0.6574 \\

            \checkmark &  &  & \checkmark & \checkmark
            & 11.26 & 0.4206 & 102.84 & 0.6471 
            & 11.10 & 0.4564 & 115.88 & 0.6568 \\

           \checkmark & \checkmark &  &  & \checkmark
            & 10.78 & 0.4091 & 55.73 & 0.6161
            & 11.19 & 0.4668 & 32.28 & 0.6310 \\

            \checkmark &  & \checkmark & \checkmark & 
            & 11.22 & 0.4207 & 54.44 & 0.6193
            & 11.28 & 0.4613 & 83.31 & 0.6429 \\

            \checkmark &  & \checkmark & \checkmark & \checkmark
            & 11.57 & 0.4378 & 43.99 & 0.6067
            & 11.16 & 0.4733 & 81.87 & 0.6549 \\

            \checkmark & \checkmark &  & \checkmark & \checkmark
            & 10.98 & 0.4196 & 53.34 & 0.6115
            & 11.33 & 0.4601 & 33.89 & 0.6307 \\

             & \checkmark & \checkmark & \checkmark & \checkmark
            & 11.21 & 0.4199 & 35.08 & 0.6131
            & 11.38 & 0.4641 & 36.29 & 0.6384 \\

             \checkmark & \checkmark & \checkmark &  & \checkmark
            & 11.26 & 0.4299 & 34.61 & 0.6060
            & 11.49 & 0.4673 & 34.36 & 0.6336 \\

             \checkmark & \checkmark & \checkmark & \checkmark & 
            & 11.59 & 0.4381 & 67.02 & 0.6192
            & 11.38 & 0.4761 & 65.66 & 0.6557 \\

             \checkmark & \checkmark & \checkmark & \checkmark & \checkmark
            & \textbf{11.72} & \textbf{0.4409} & \textbf{30.84} & \textbf{0.6029}
            & \textbf{11.58} & \textbf{0.4851} & \textbf{28.68} & \textbf{0.6282} \\
        \bottomrule
    \end{tabular}
    }
    \vspace{-2mm}
    \caption{Ablation study on five designs in our Top2Pano model (top-down view \textbf{seg}mentation, \textbf{floor} reinforcement, \textbf{wall} reinforcement, coarse \textbf{depth} panorama, and coarse \textbf{color}ed panorama).}
    \label{tab:results}
\end{table}

\subsection{Ablation Study}
We conducted comprehensive ablation studies to analyze and validate the contribution of each component in our model. Specifically, we performed experiments comparing several model variants against the original. These experiments involved removing key elements such as the structural reinforcement of the floor and walls, the segmentation input to the OccRecon module, and the coarse depth and colored panoramas as conditional inputs to the PanoGen module.

As shown in Table~\ref{tab:results}, our original model achieves the highest overall scores across all four metrics, with any modification leading to some degree of performance degradation. Removing coarse colored panoramas or the embedded floor significantly disrupts furniture placement and color accuracy. While the room structure remains mostly intact, furniture positions become unreliable, and color representations appear distorted. Conversely, excluding coarse depth panoramas or embedded walls maintains color and furniture accuracy but compromises the spatial understanding and overall quality of room structure reconstruction. These effects are further illustrated in the qualitative results in the supp. materials.

\subsection{Generalization, Stylization, Manipulation}

We employed cross-dataset evaluation to assess our model's generalization capability. As shown in Table~\ref{tab:results_cross}, although there is a slight decline in metric scores, our model maintains strong performance. Additionally, the trained parameters tend to generate photorealistic walls that reflect characteristics of the training dataset.

\begin{figure}[t]
\centering
\includegraphics[width=0.8\linewidth]{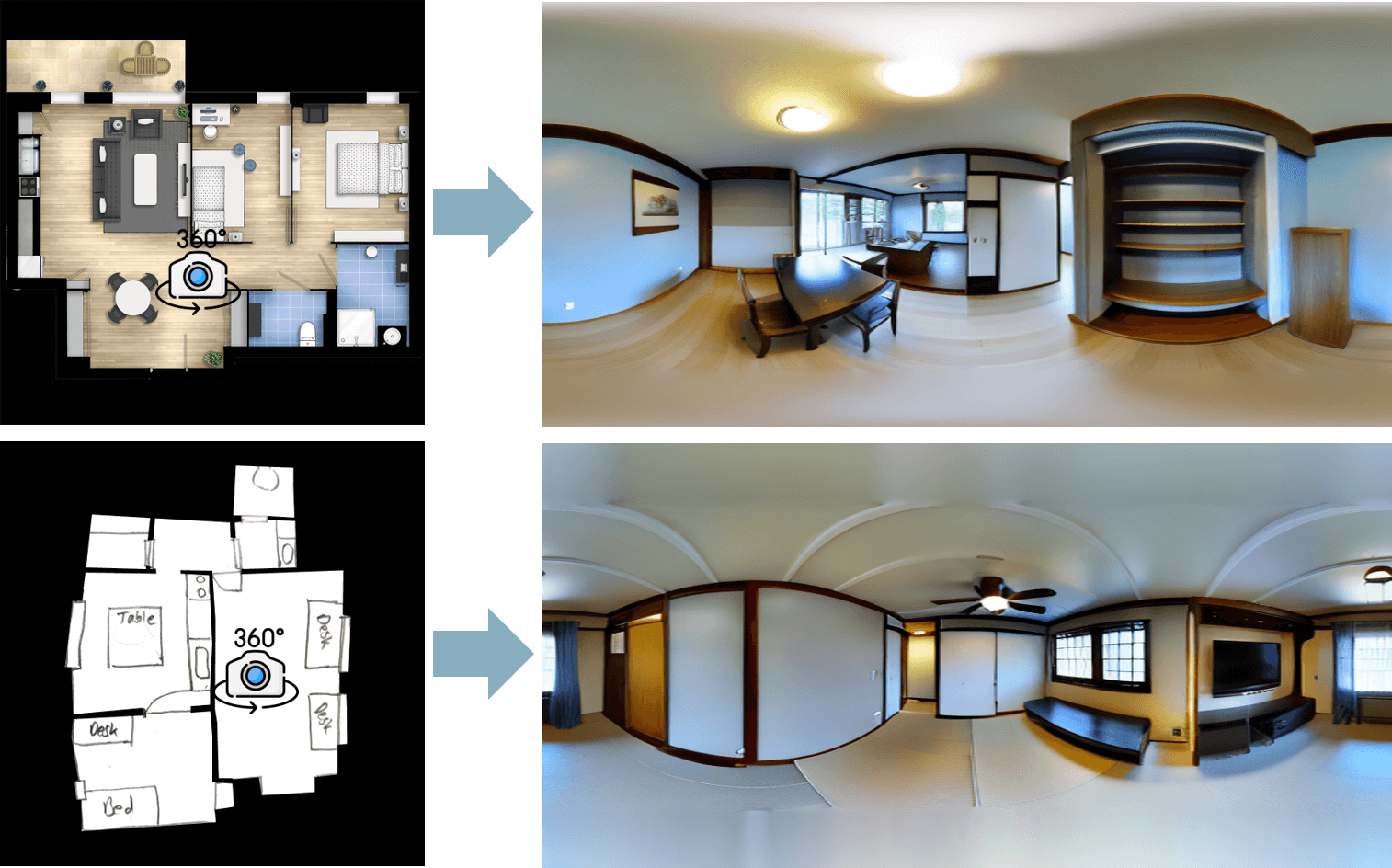}
\vspace{-2mm} 
\caption{Our Top2Pano model generalizes to both textured floorplans (first row) and hand-drawn sketched floorplans (second row), incorporating stylized \texttt{[Japanese]} control.}
\label{fig:floorplan_jp}
\end{figure}

\begin{figure}[t]
\centering
\includegraphics[width=0.8\linewidth]{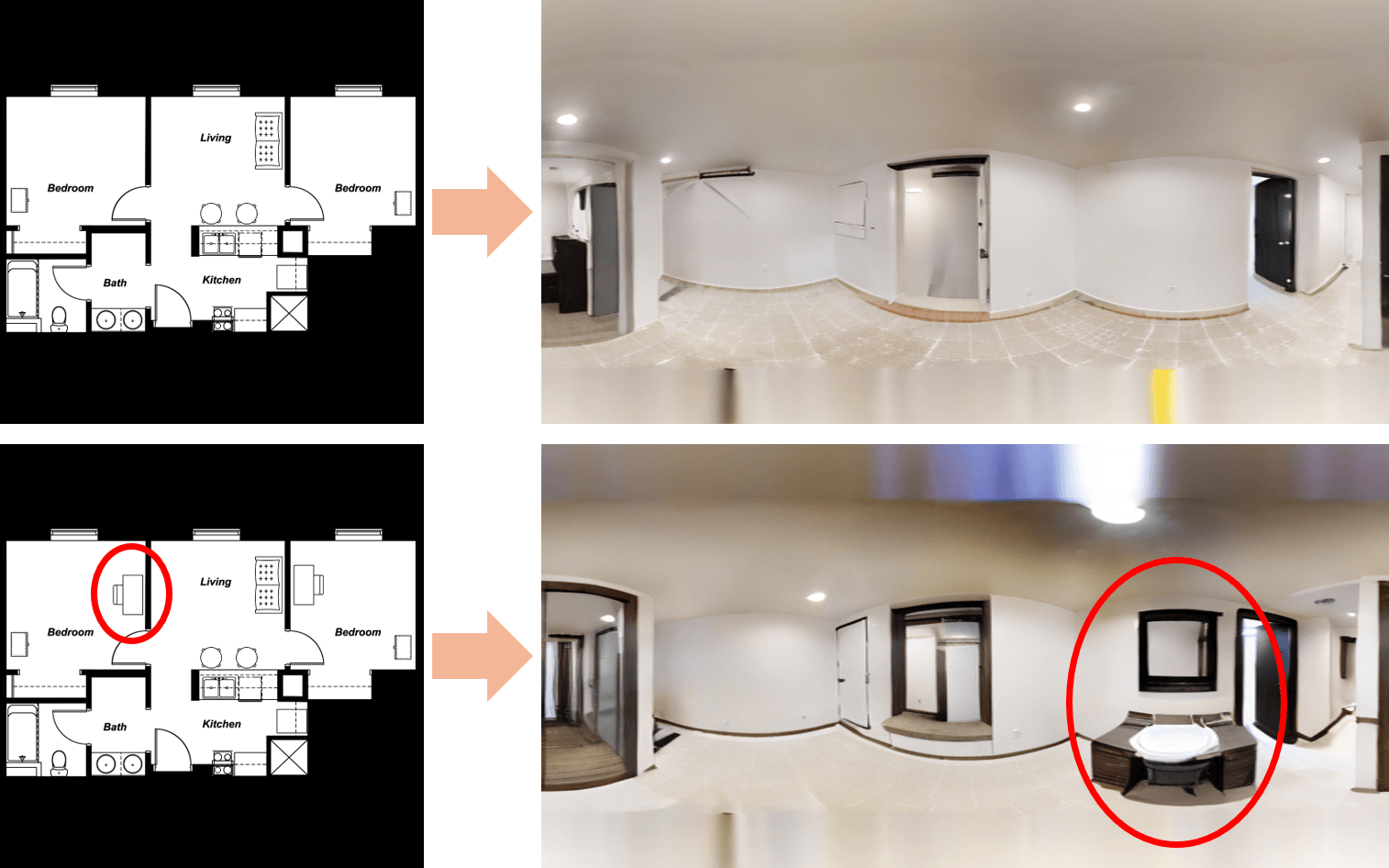}
\vspace{-2mm} 
\caption{
Top2Pano enables panorama manipulation via compositional floorplan editing. In the second row, adding a rectangular object to the floorplan (compared to the first row) leads the model to generate a washstand with a mirror in the panorama.
}
\label{fig:manipulate}
\end{figure}

Our model also generalizes well to floorplans. When given a specific floorplan and camera positions, the model assists users in generating indoor panoramic images. By using different text prompts, users can create various interior design styles, explore the house from a first-person perspective, and modify its style according to their preferences.
We tested our model with three types of floorplans. The first type is a colored floorplan (Figure~\ref{fig:floorplan_jp}, first row), which provides detailed information about room colors and furniture hues, making it highly informative. The second type is a plain floorplan (Figure~\ref{fig:teaser}, bottom), which lacks color information and shows only the room structure and furniture layout. The third type is a hand-drawn floorplan sketch (Figure~\ref{fig:floorplan_jp}, first row), which offers a rough visual representation of the room. Our model successfully generates accurate panoramic images from these floorplans and adapts the visual style based on the provided textual descriptions.
Moreover, our model enables panorama manipulation by editing objects in the floorplan, such as adding new items, as illustrated in Figure~\ref{fig:manipulate}.

\setlength{\textfloatsep}{5pt}
\setlength{\floatsep}{3pt}

\begin{table}[t]
    \centering
    \footnotesize
    \begin{tabular}{lcccc}
        \toprule
         & PSNR $\uparrow$ & SSIM $\uparrow$ & FID $\downarrow$& LPIPS $\downarrow$\\
        \midrule
        $G \rightarrow M$ &11.08 & 0.4397 &  40.74 & 0.6285 \\
        $M \rightarrow G$ & 11.03& 0.4366& 45.87&0.6353\\
        \bottomrule
    \end{tabular}
    \vspace{-2mm} 
    \caption{Cross-dataset evaluation. ``G'' represents Gibson; ``M'' represents Matterport3D.}
    \label{tab:results_cross}
\end{table}

\subsection{Limitations}

\begin{figure}[t]
    \centering
    \footnotesize
    \setlength{\tabcolsep}{1pt}
    \begin{tabular}{ccc}
    \centering
        {\includegraphics[width=0.2\linewidth]{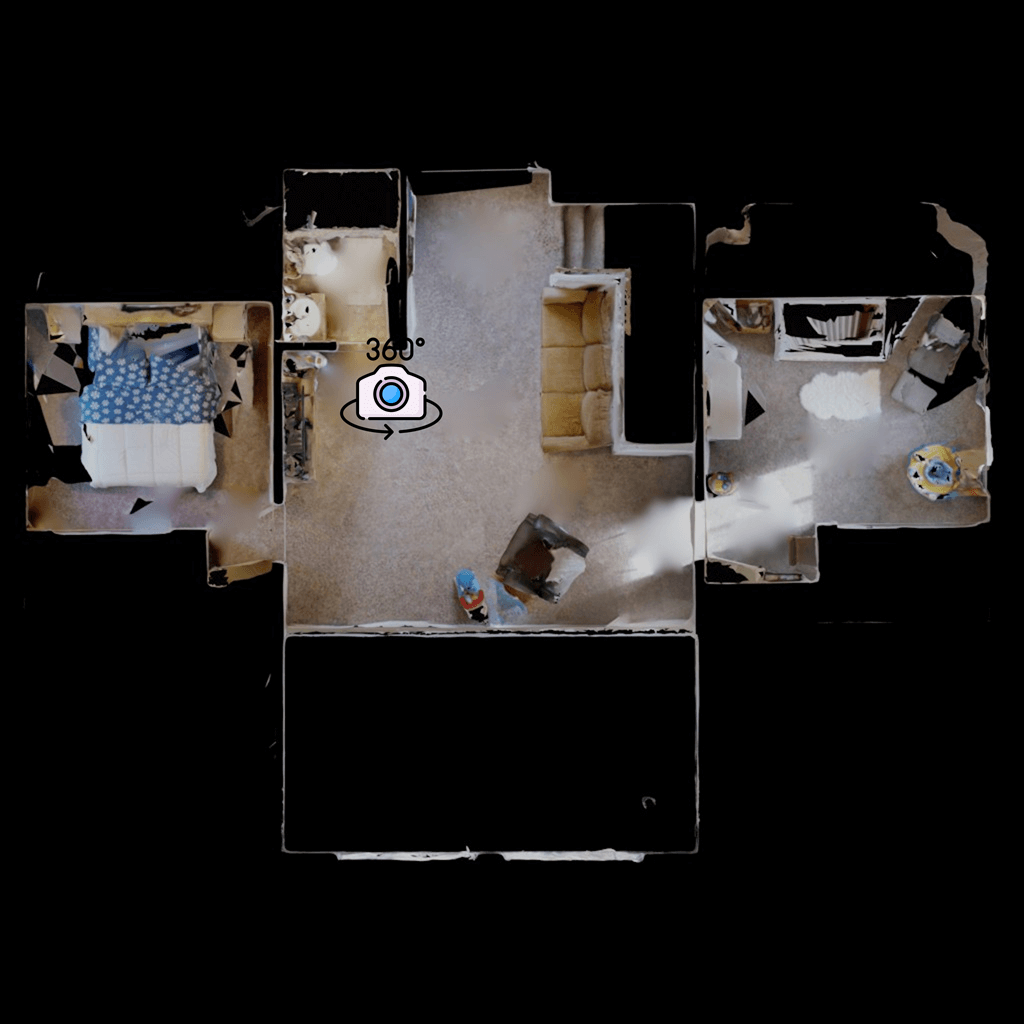}} &  
        {\includegraphics[width=0.4\linewidth]{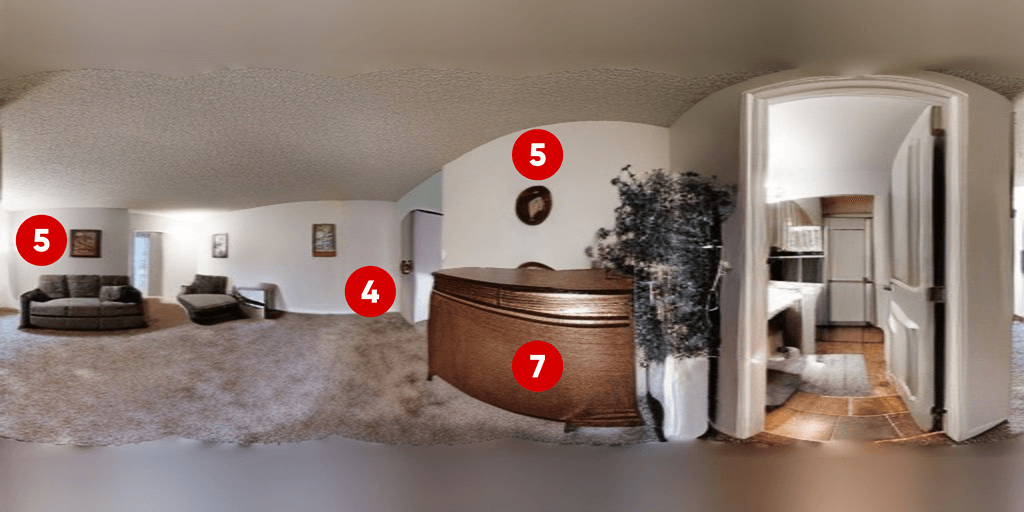}} &  
        {\includegraphics[width=0.4\linewidth]{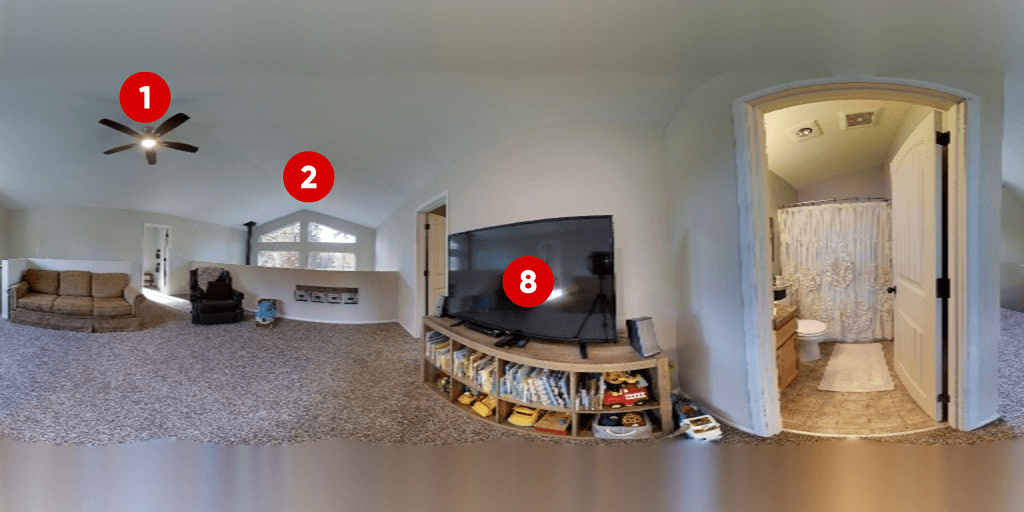}} \\
        {\includegraphics[width=0.2\linewidth]{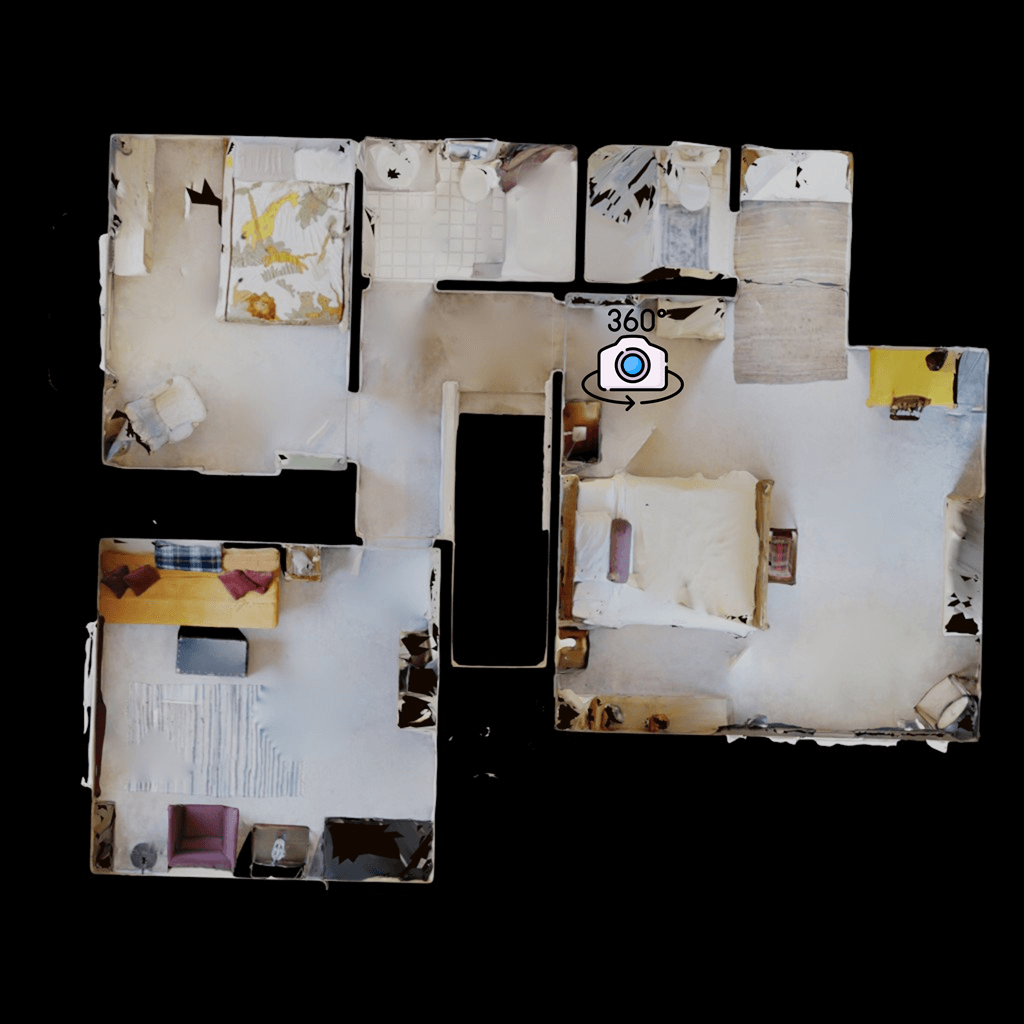}} &  
        {\includegraphics[width=0.4\linewidth]{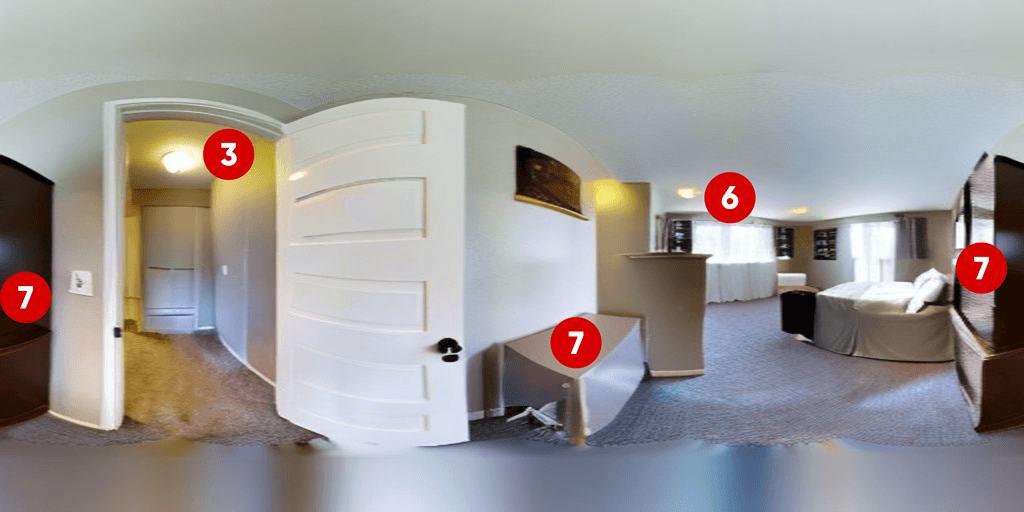}} &  
        {\includegraphics[width=0.4\linewidth]{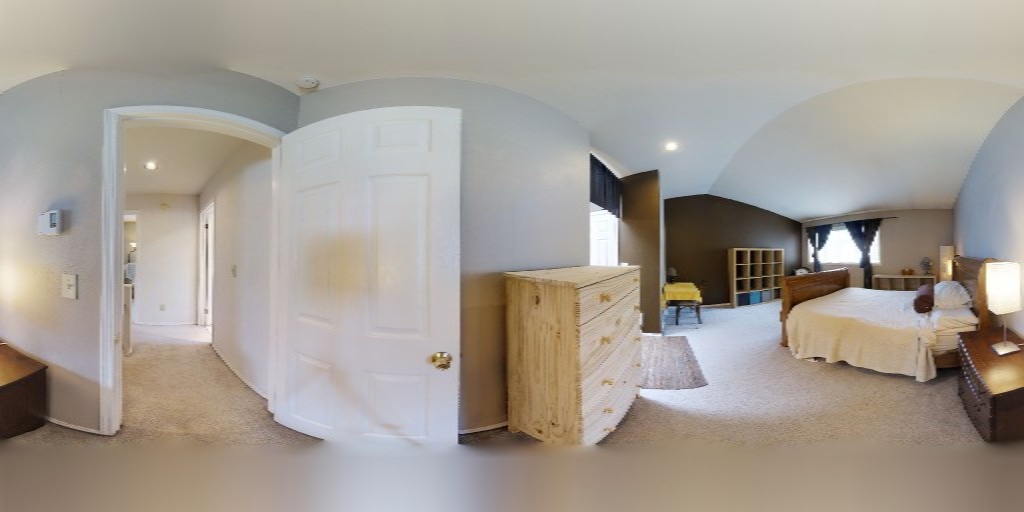}} \\ 
        {\includegraphics[width=0.2\linewidth]{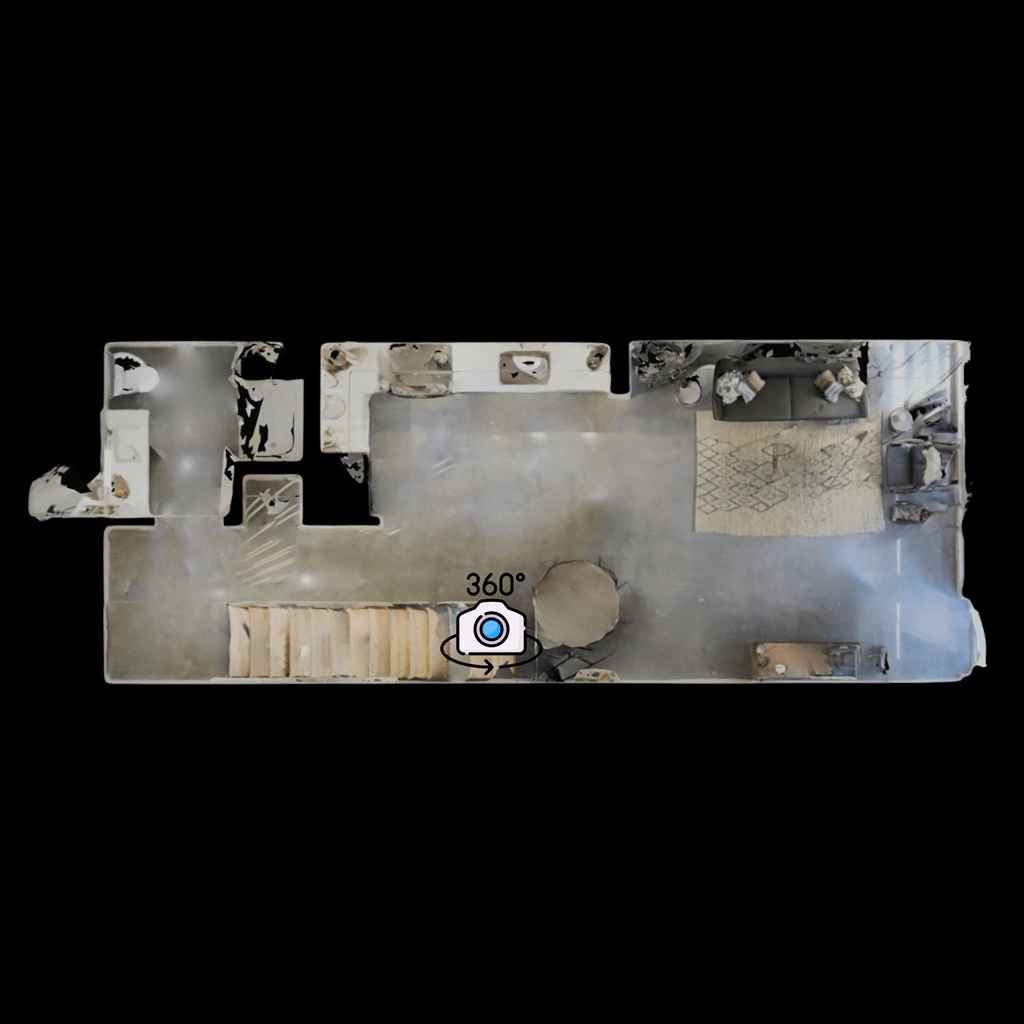}} &  
        {\includegraphics[width=0.4\linewidth]{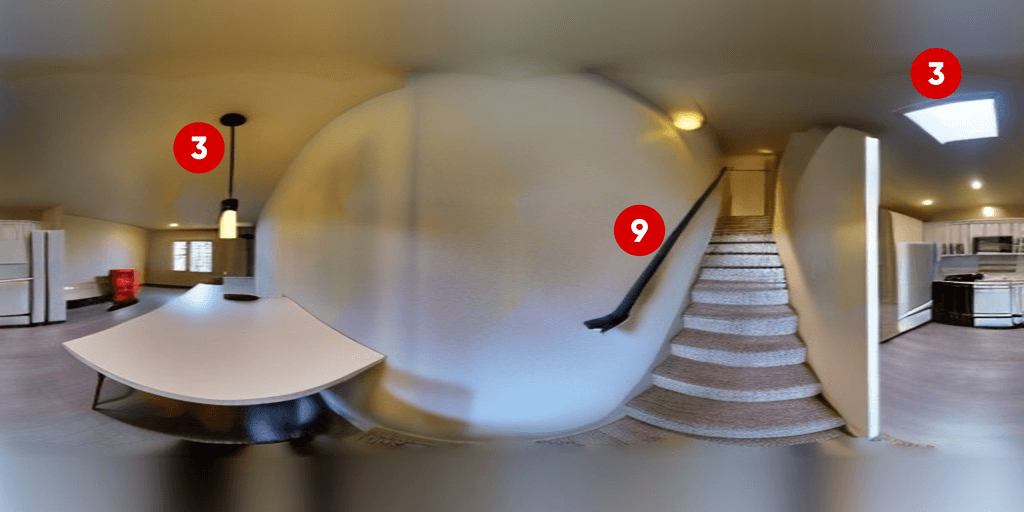}} &  
        {\includegraphics[width=0.4\linewidth]{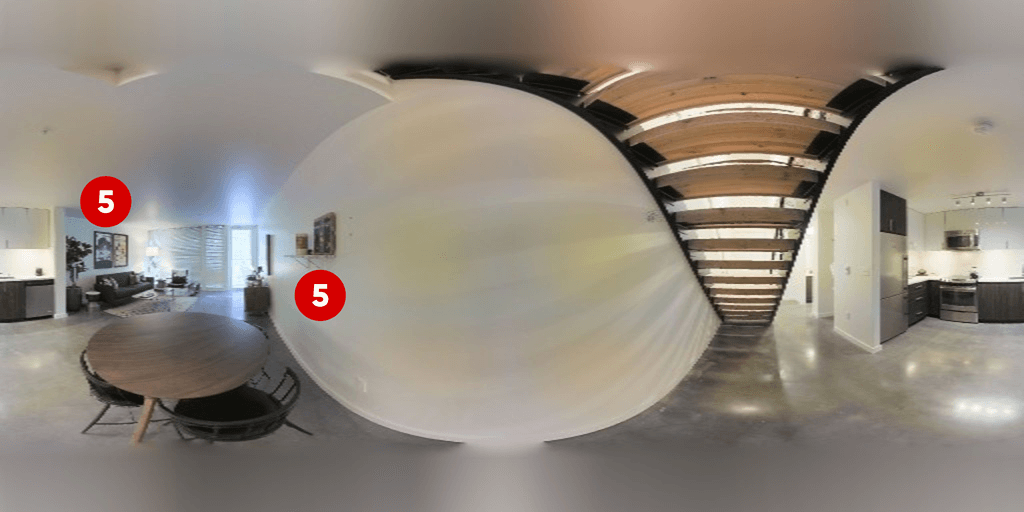}} \\      
        Top-down & Our Prediction & Ground Truth\\ 
    \end{tabular}
    \vspace{-3mm}
    \caption{Failure cases (zoom in to view error types).}
    \label{fig:failure_case}
\end{figure}

\noindent
\textbf{Failure cases.} Figure~\ref{fig:failure_case} shows representative failure cases. We annotate different types of failures with numbered labels:
\begin{itemize}
  \item \textit{Ceiling}: \emoji{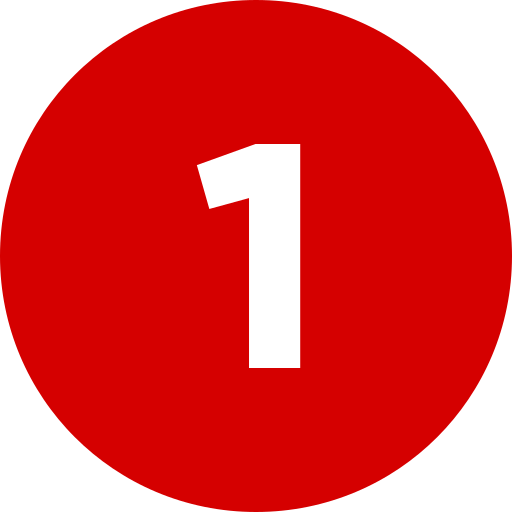} missing fan, \emoji{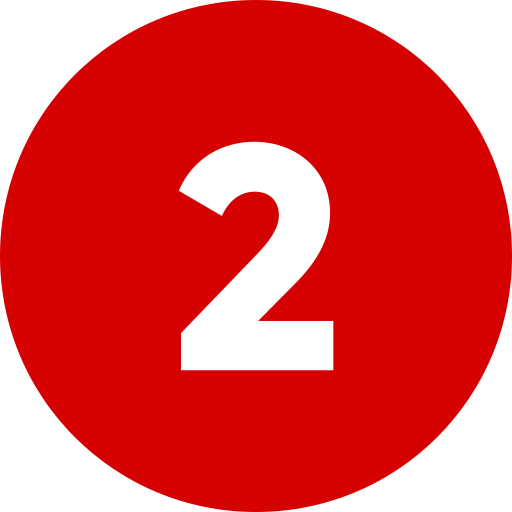} vaulted ceiling, \emoji{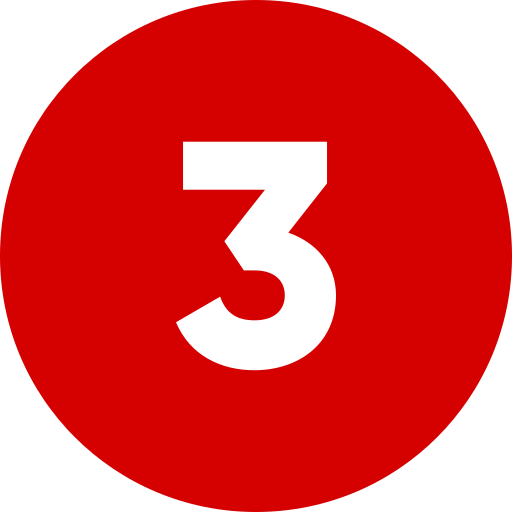} false light;
  \item \textit{Wall}: \emoji{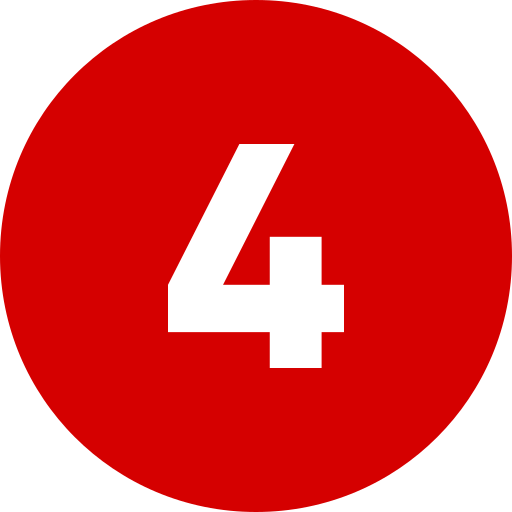} height error, \emoji{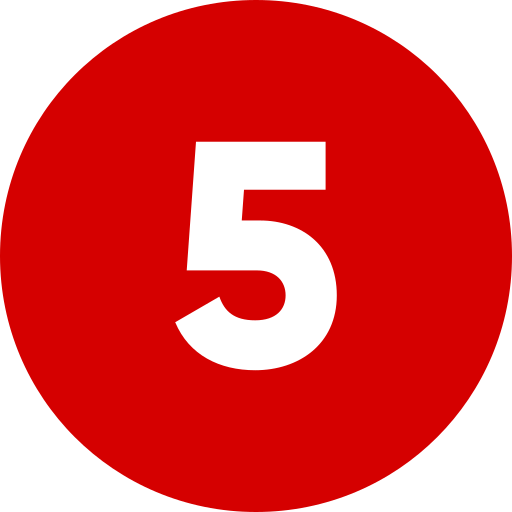} false or missing decorations;
  \item \textit{Window}: \emoji{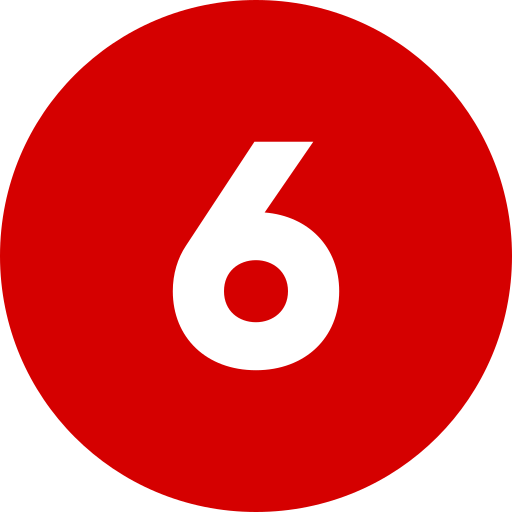}~false window;
  \item \textit{Furniture}: \emoji{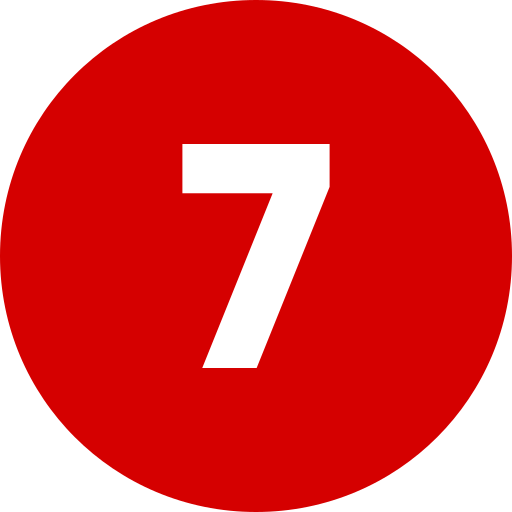} height error;
  \item \textit{Thin object}: \emoji{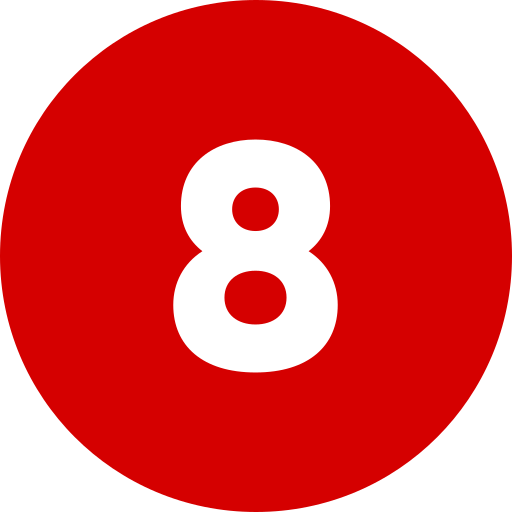} missing flat-screen TV;
  \item \textit{Stairs}: \emoji{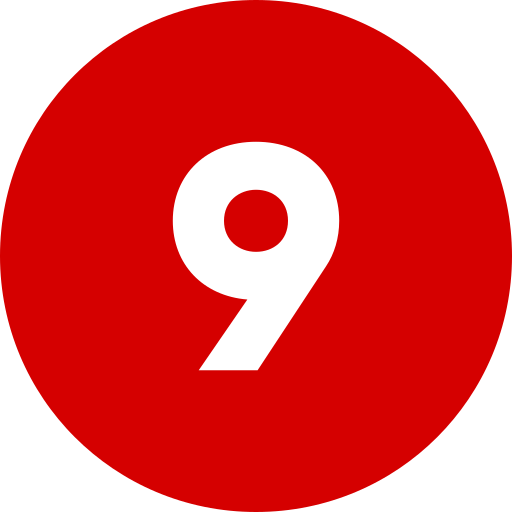} false direction.
\end{itemize}

The failure cases stem from the ambiguity of the 2D input, leading to hallucinated objects that are not observable from the top-down view. The limitations in handling vertical structural details are largely due to the inherent ambiguity of the task.

\smallskip
\noindent
\textbf{Limited Vertical FoV.} 
The generated panoramas exhibit a limited vertical field of view (FoV), reflecting the constraints of the training data. We expect improved performance with future datasets that include full vertical FoV panoramas.

\section{Conclusions}
\vspace{-1mm}

We present \textbf{Top2Pano}, a novel method for generating high-quality 360° indoor panoramas from 2D top-down views. The model first estimates volumetric occupancy to infer 3D structure, then applies volumetric rendering to produce coarse color and depth panoramas. These guide a diffusion-based refinement stage via ControlNet. To our knowledge, this is the first approach to generate panoramas from top-down views. Experiments on two datasets show that Top2Pano outperforms baselines in reconstructing room layouts and realistic furniture.

\clearpage
{
    \small
    \bibliographystyle{ieeenat_fullname}
    \bibliography{main}
}
\clearpage

\appendix

% Redefine figure numbering for the appendix
\renewcommand{\thefigure}{\Alph{section}.\arabic{figure}} 
\setcounter{figure}{0} % Reset the figure counter for the appendix

\onecolumn
\section{Qualitative Results on Ablation Study}
\vspace{-3mm}

\begin{figure*}[h]
    \centering
    \begin{subfigure}[t]{0.13\textwidth}
        \includegraphics[width=\textwidth]{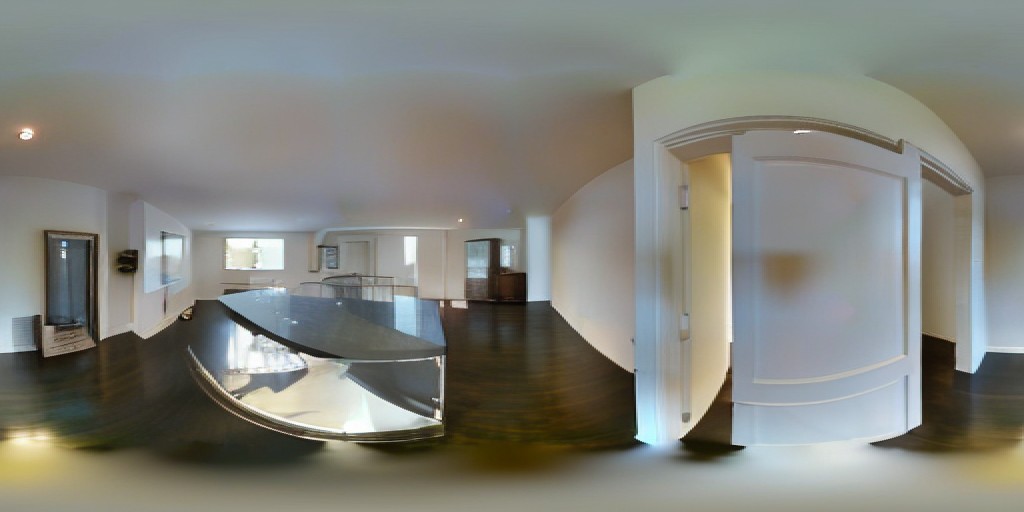}
        \includegraphics[width=\textwidth]{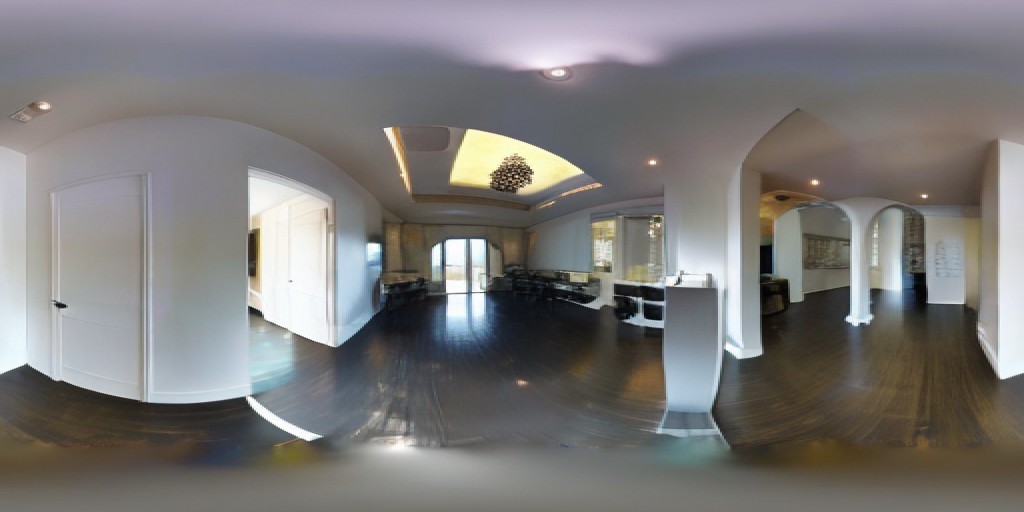}
        \includegraphics[width=\textwidth]{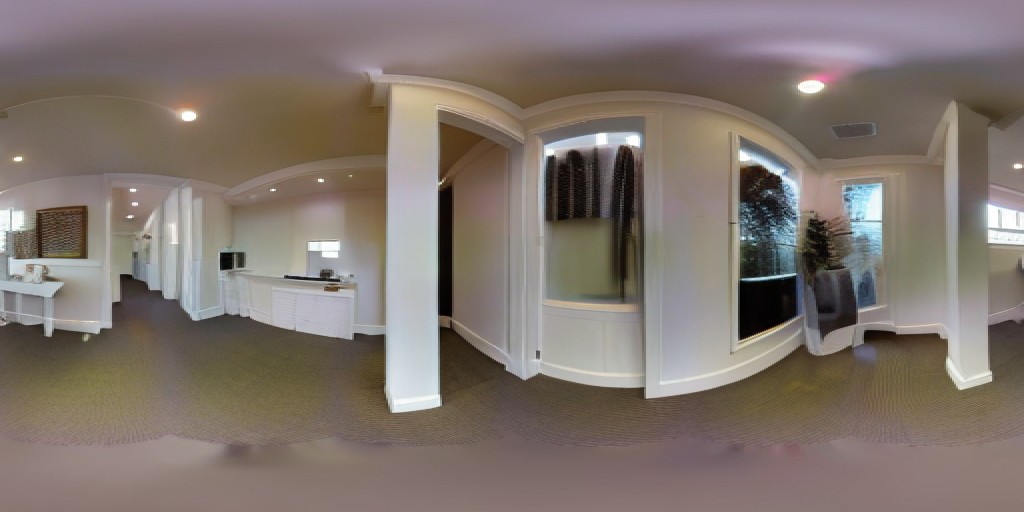}
        \includegraphics[width=\textwidth]{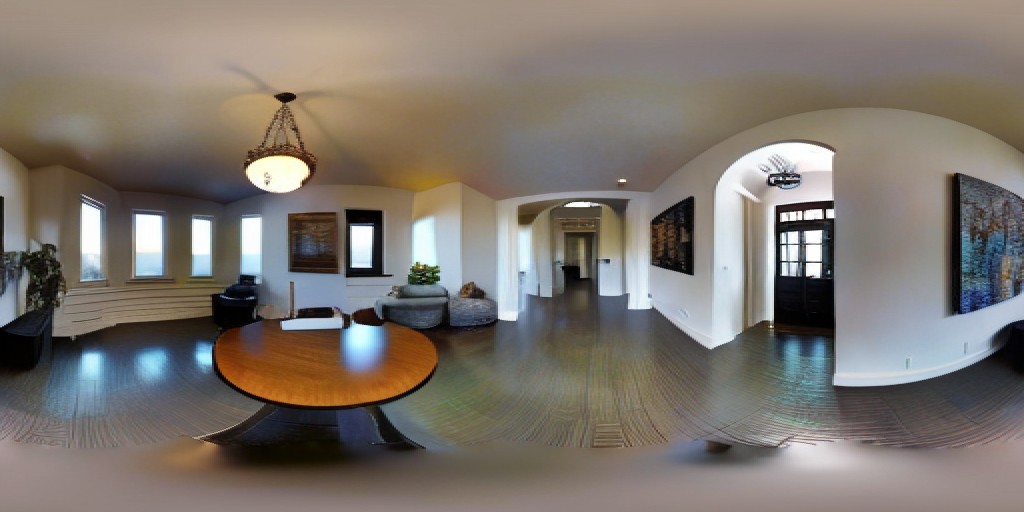}
        \includegraphics[width=\textwidth]{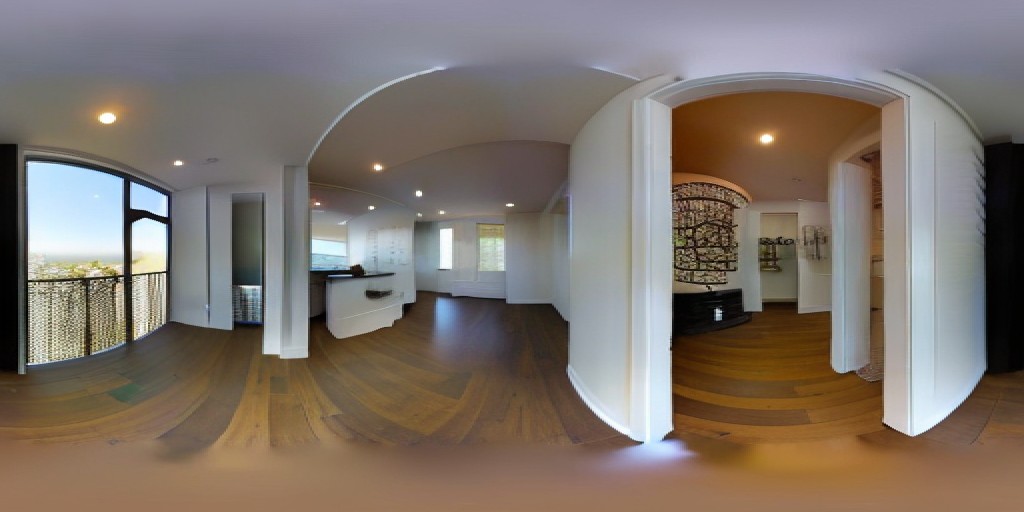}
        \includegraphics[width=\textwidth]{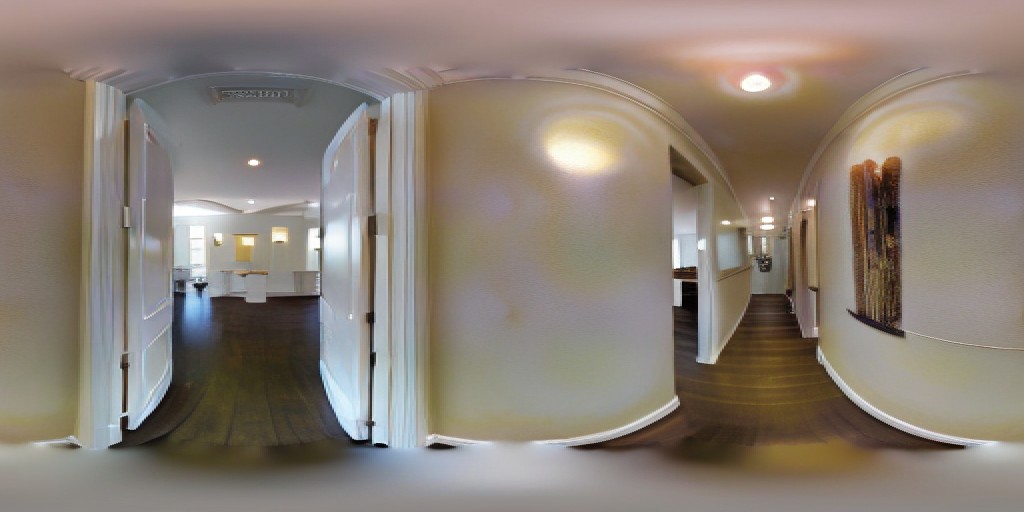}
        \includegraphics[width=\textwidth]{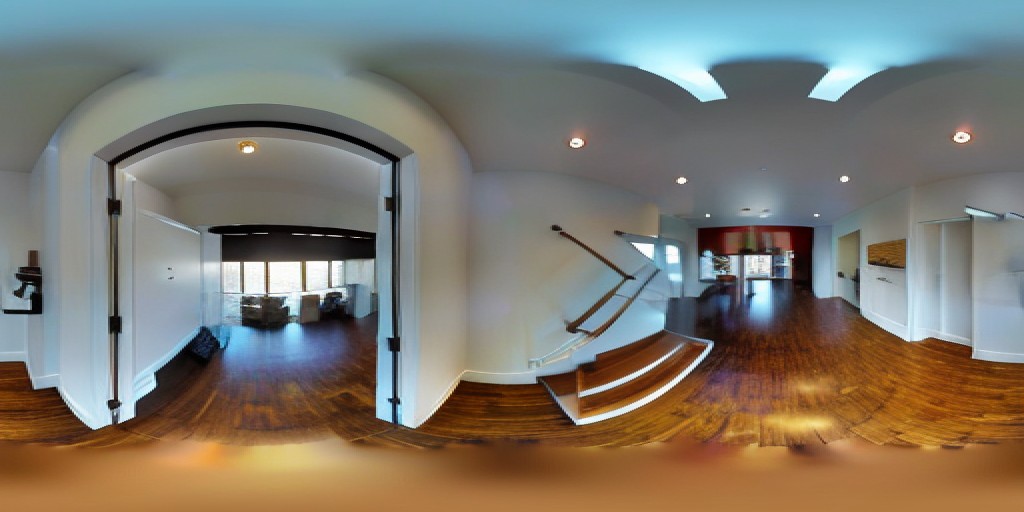}
        \includegraphics[width=\textwidth]{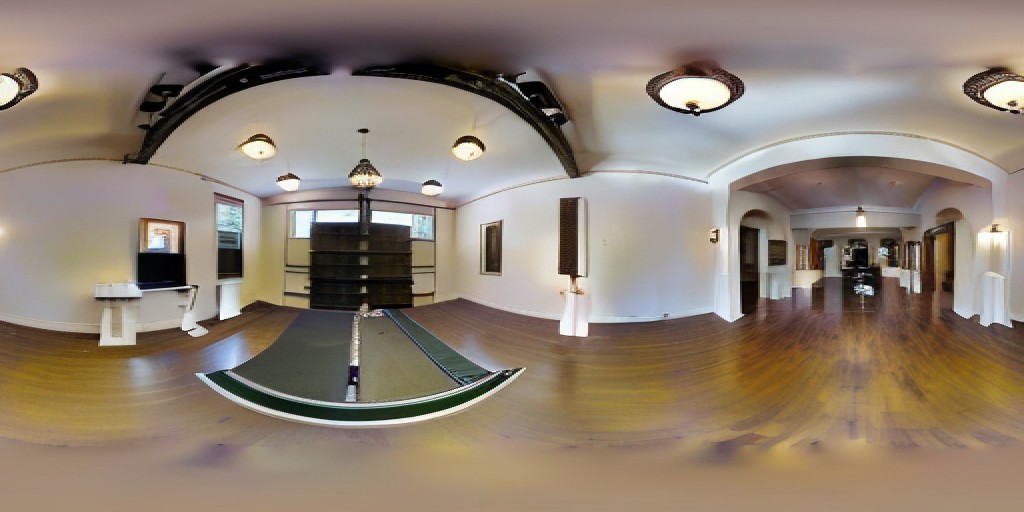}
        \caption{w/o floor}
    \end{subfigure}
    \hfill  
    \begin{subfigure}[t]{0.13\textwidth}
        \includegraphics[width=\textwidth]{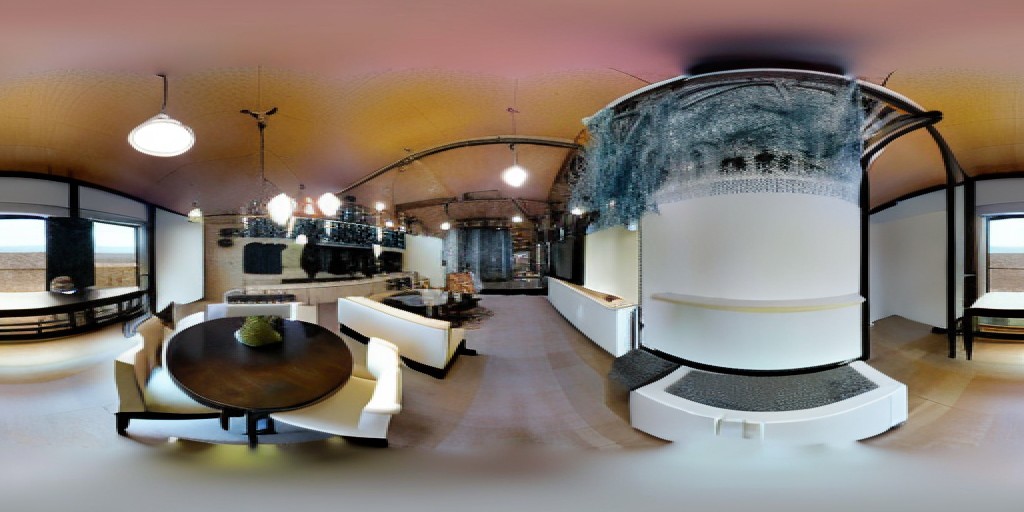}
        \includegraphics[width=\textwidth]{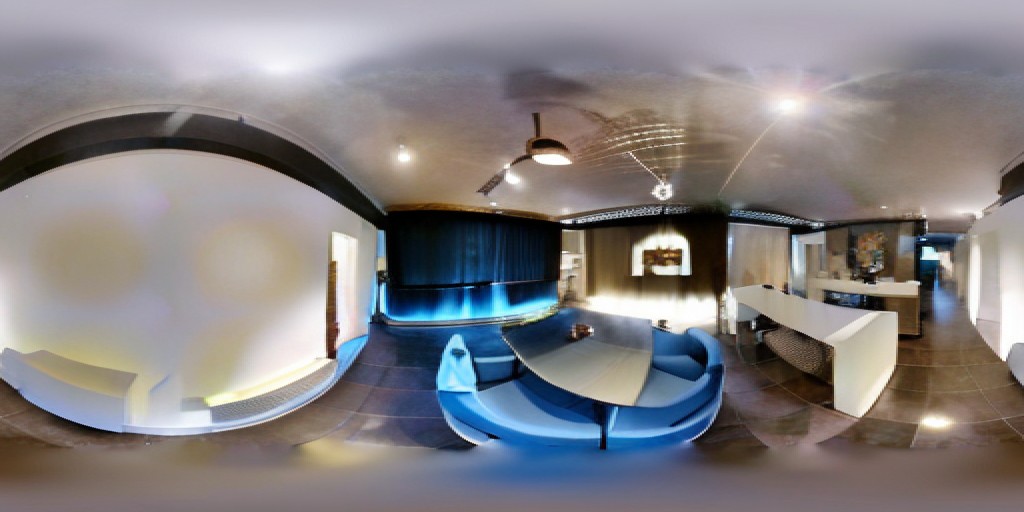}
        \includegraphics[width=\textwidth]{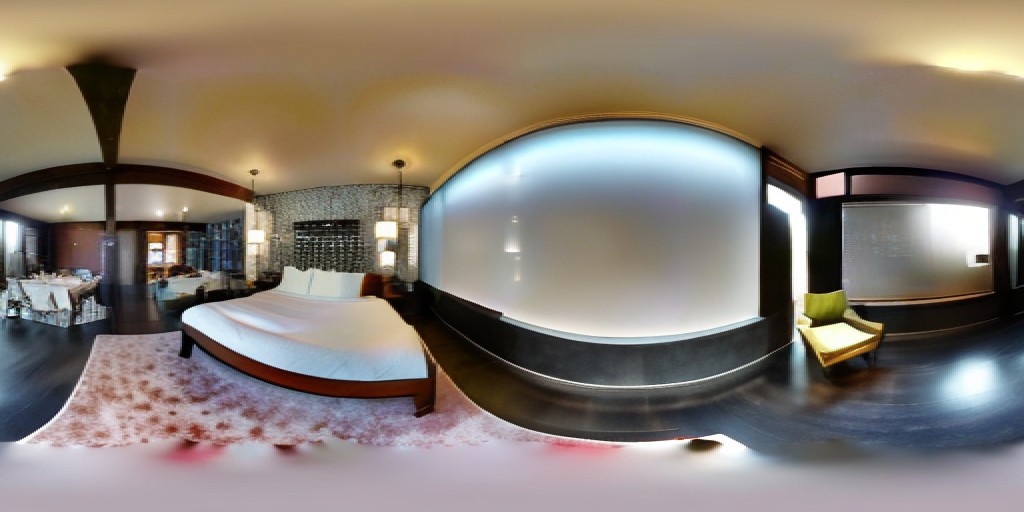}
        \includegraphics[width=\textwidth]{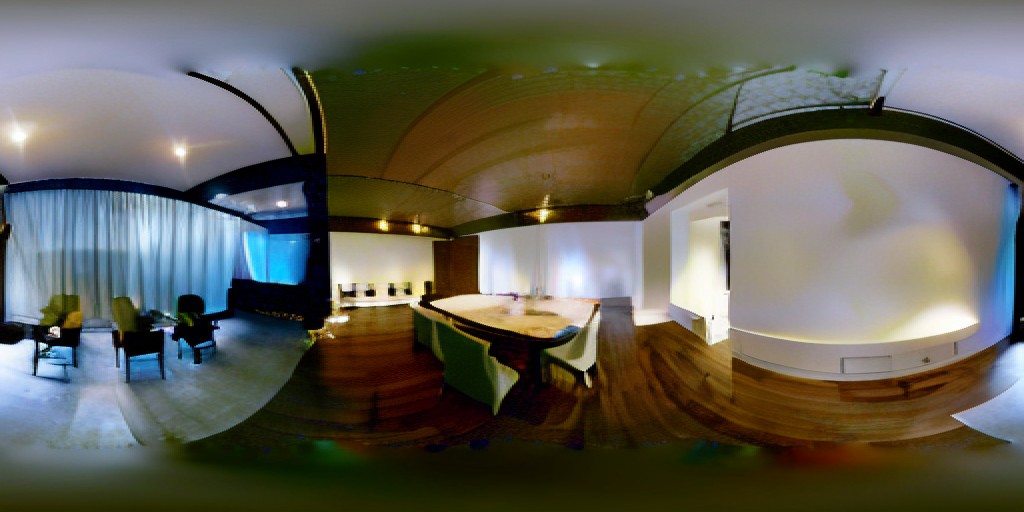}
        \includegraphics[width=\textwidth]{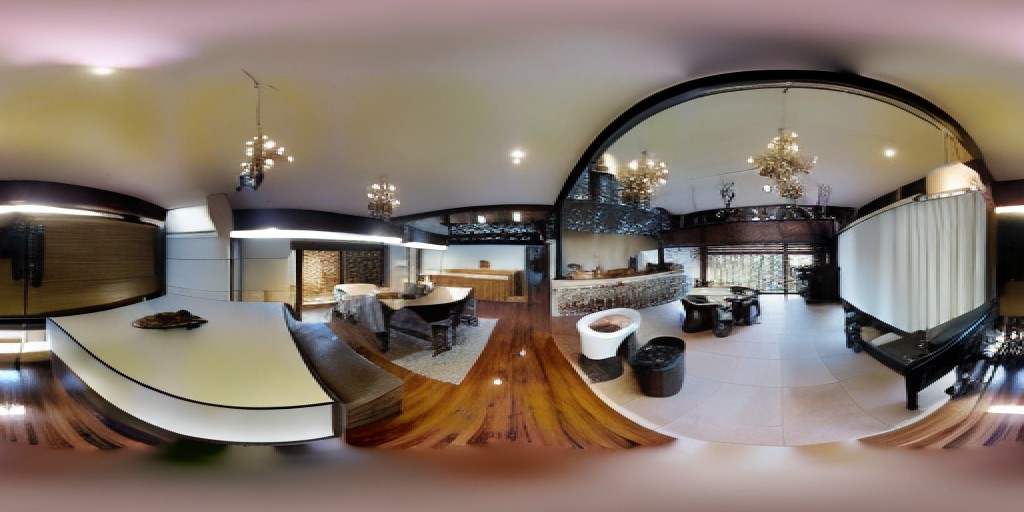}
        \includegraphics[width=\textwidth]{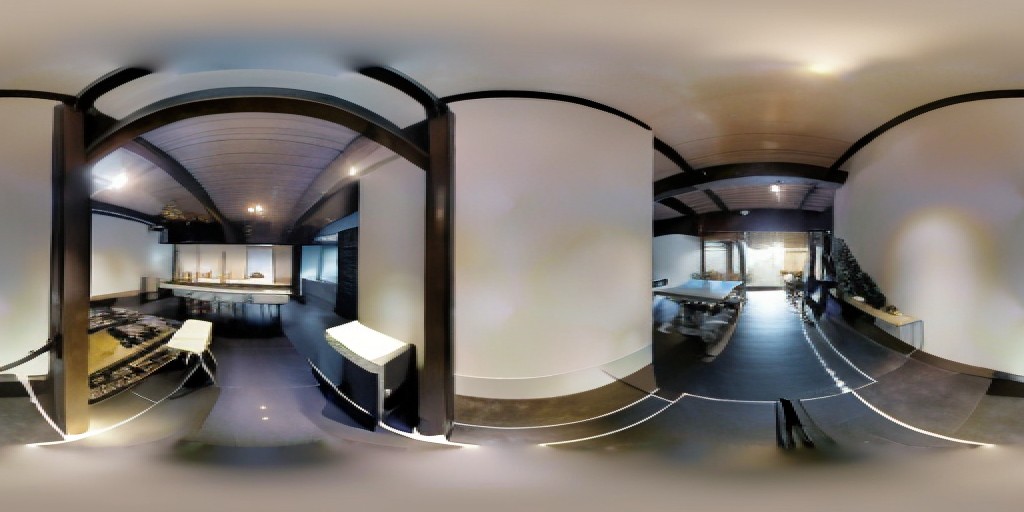}
        \includegraphics[width=\textwidth]{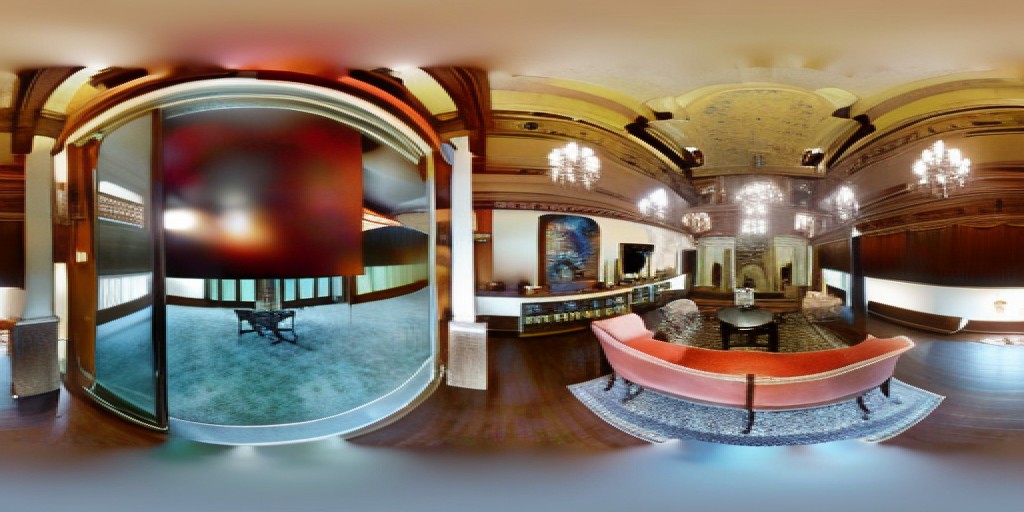}
        \includegraphics[width=\textwidth]{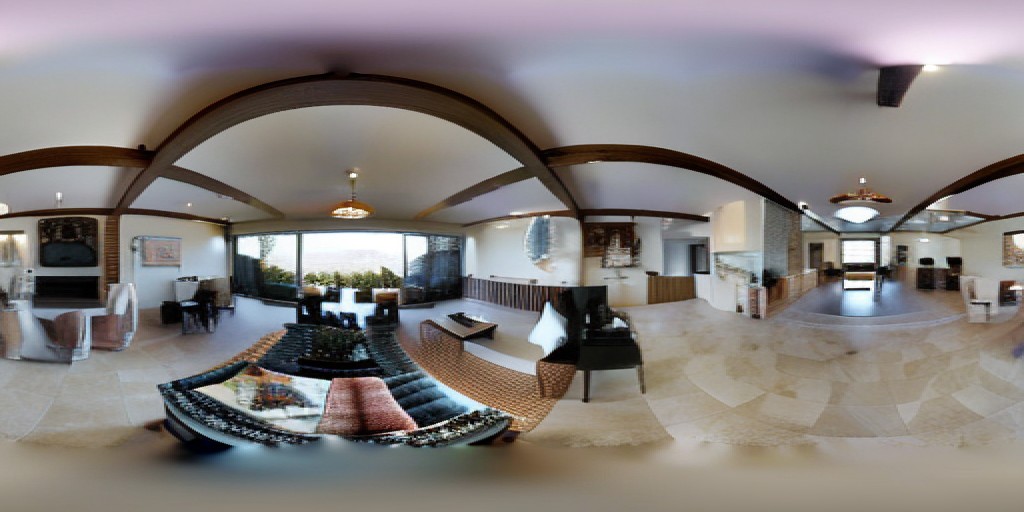}
        \caption{w/o wall}
    \end{subfigure}
    \hfill
    \begin{subfigure}[t]{0.13\textwidth}
        \includegraphics[width=\textwidth]{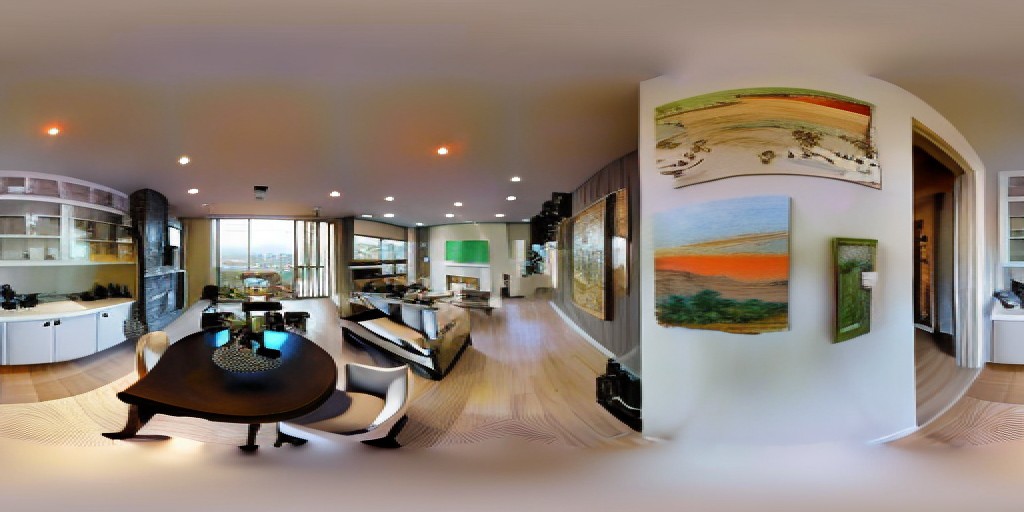}
        \includegraphics[width=\textwidth]{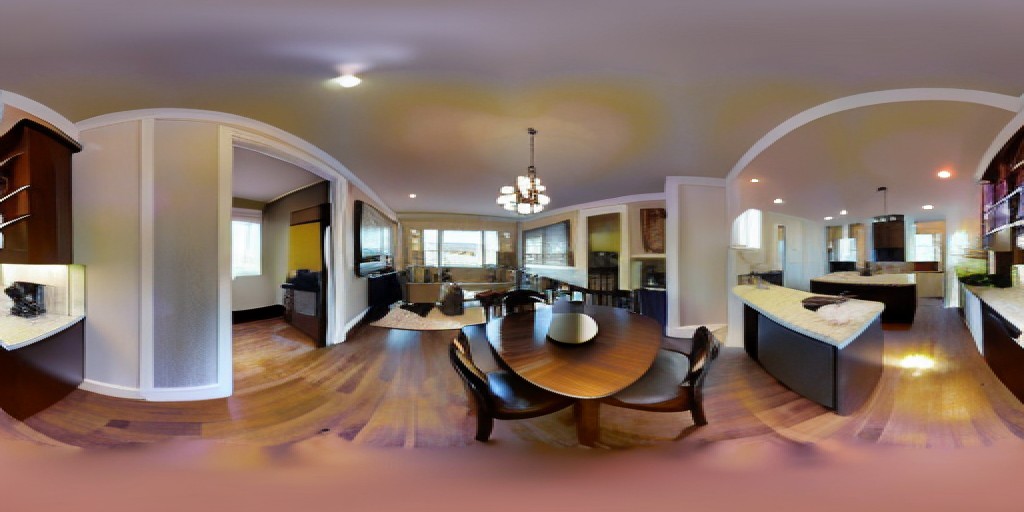}
        \includegraphics[width=\textwidth]{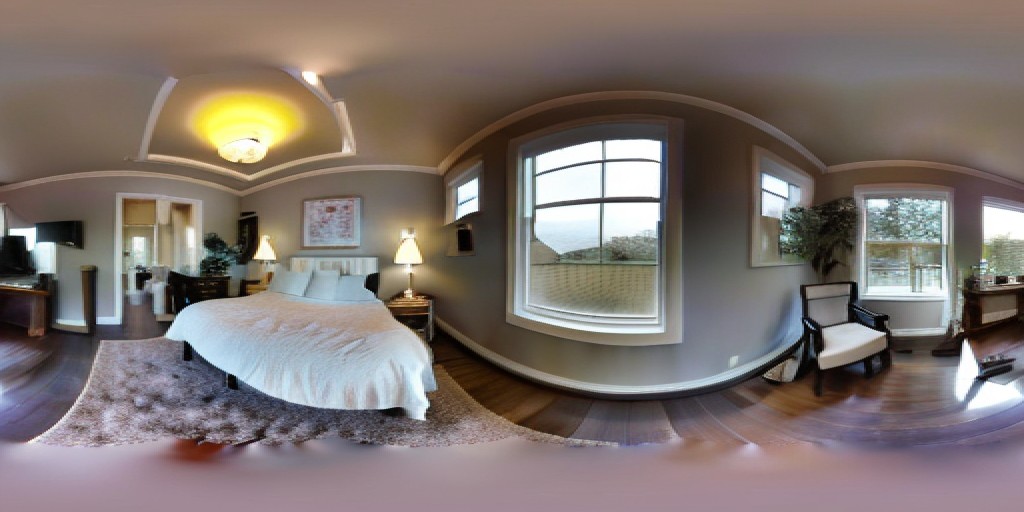}
        \includegraphics[width=\textwidth]{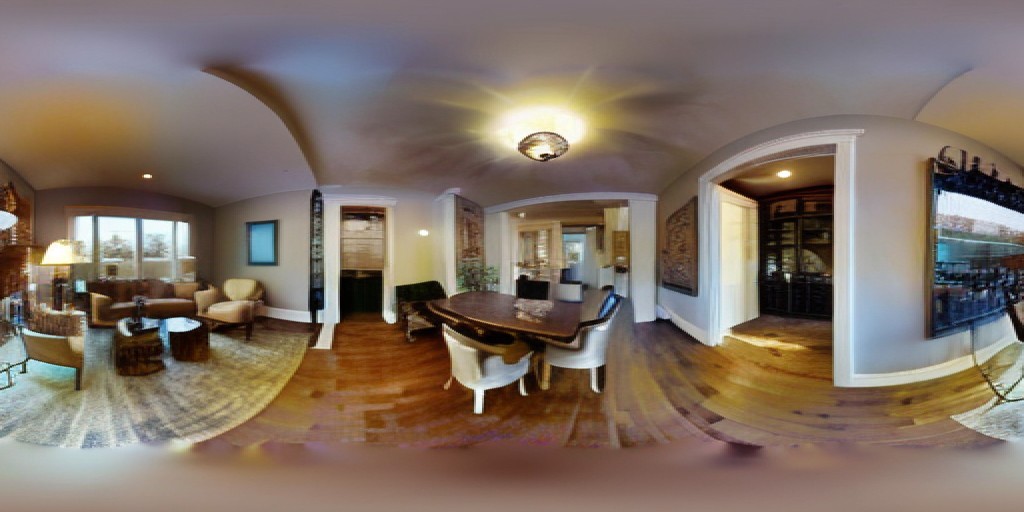}
        \includegraphics[width=\textwidth]{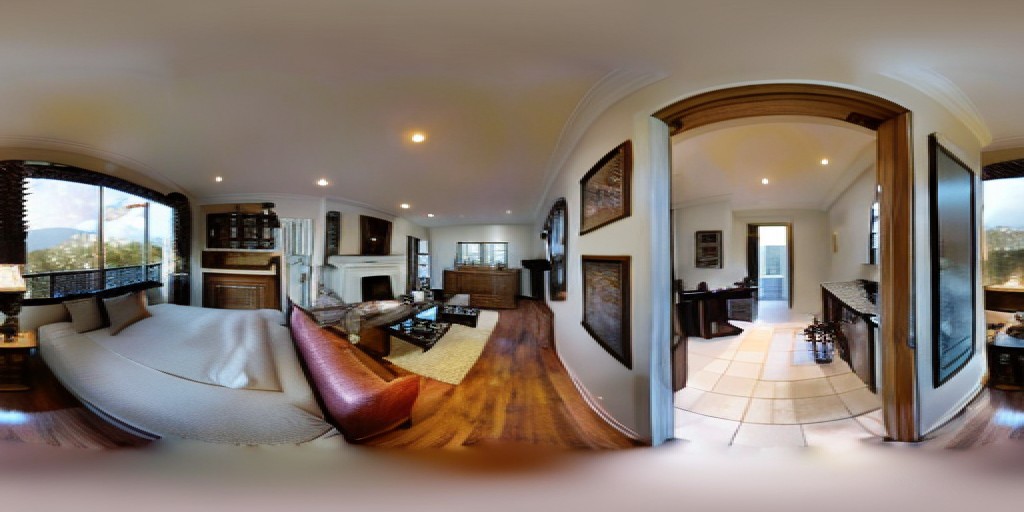}
        \includegraphics[width=\textwidth]{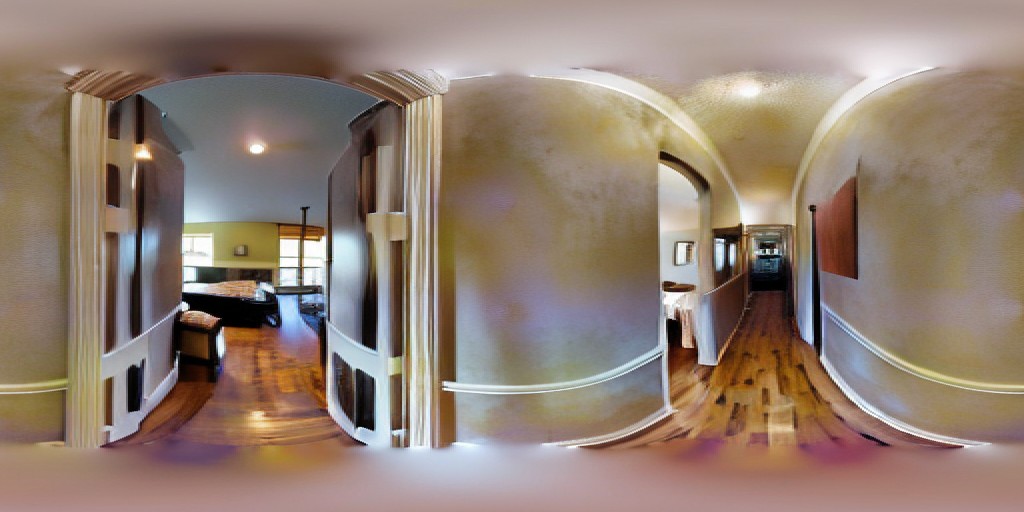}
        \includegraphics[width=\textwidth]{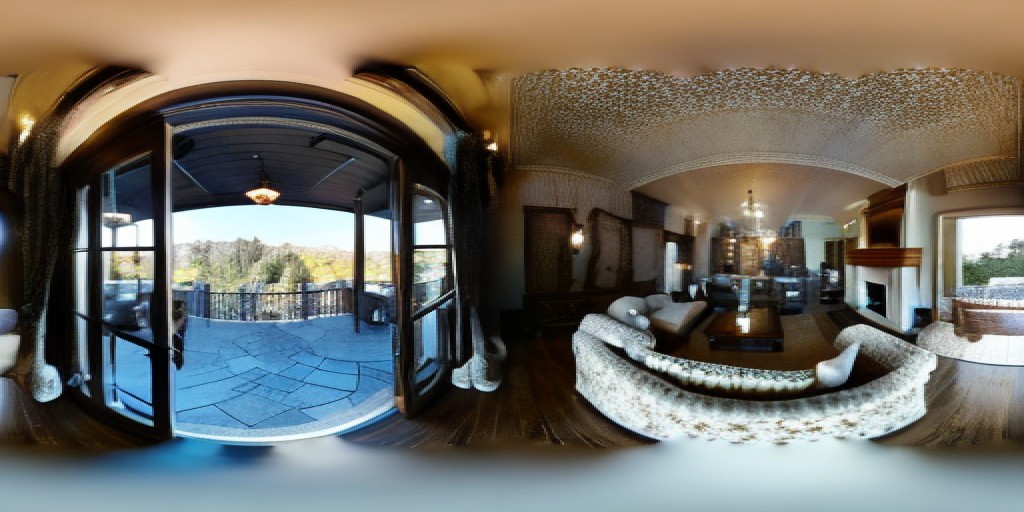}
        \includegraphics[width=\textwidth]{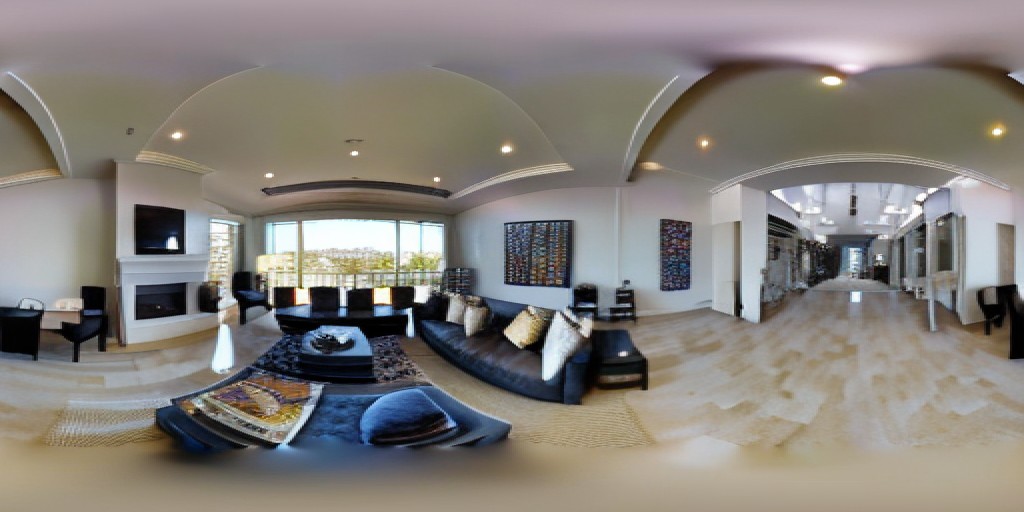}
        \caption{w/o segment}
    \end{subfigure}
    \hfill
    \begin{subfigure}[t]{0.13\textwidth}
       \includegraphics[width=\textwidth]{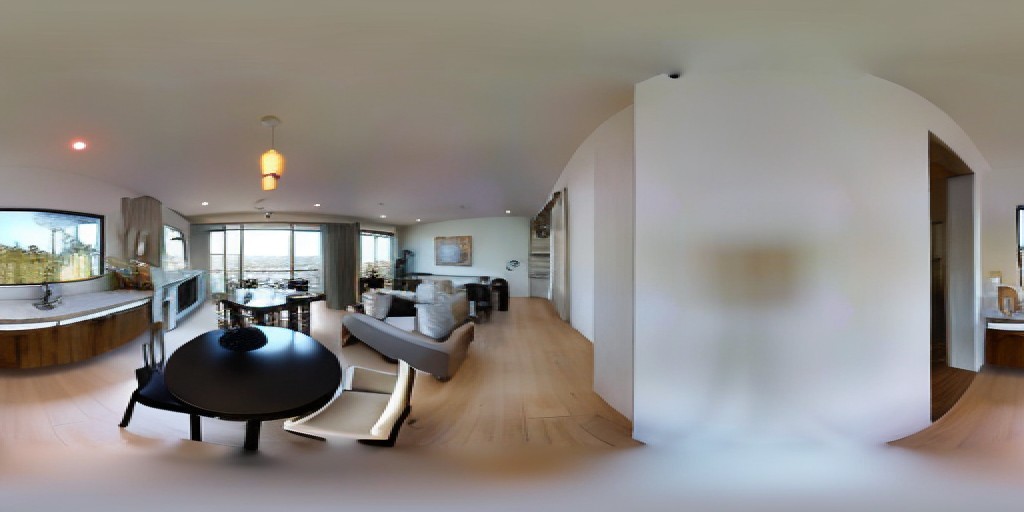}
        \includegraphics[width=\textwidth]{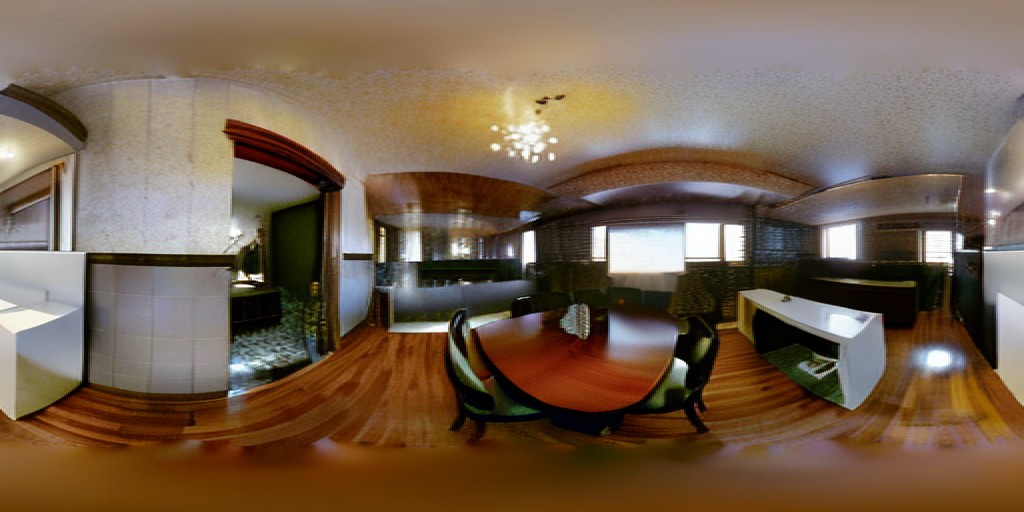}
        \includegraphics[width=\textwidth]{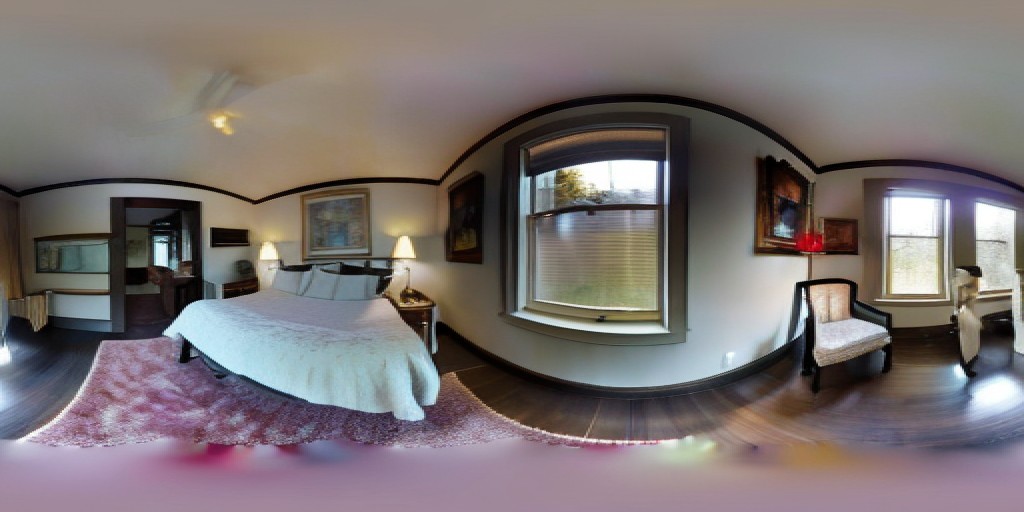}
        \includegraphics[width=\textwidth]{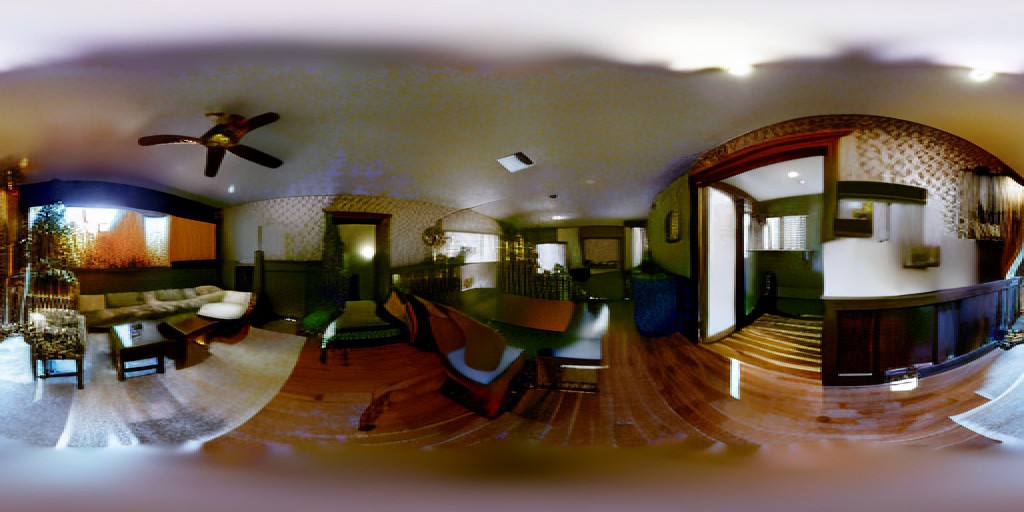}
        \includegraphics[width=\textwidth]{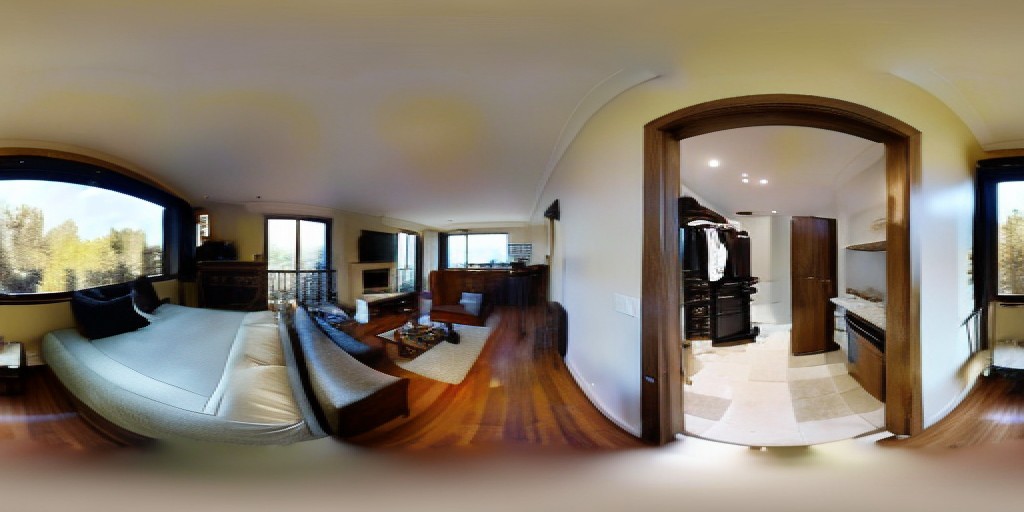}
        \includegraphics[width=\textwidth]{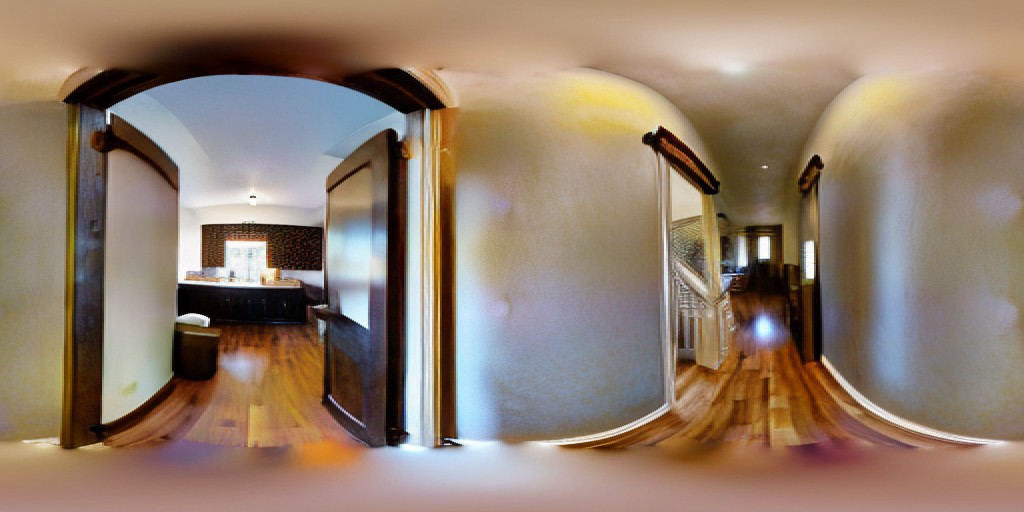}
        \includegraphics[width=\textwidth]{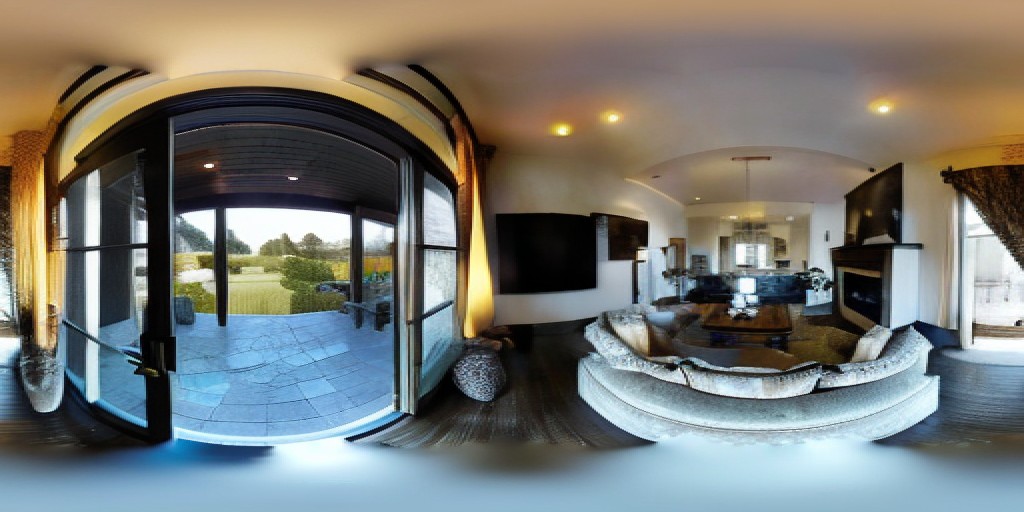}
        \includegraphics[width=\textwidth]{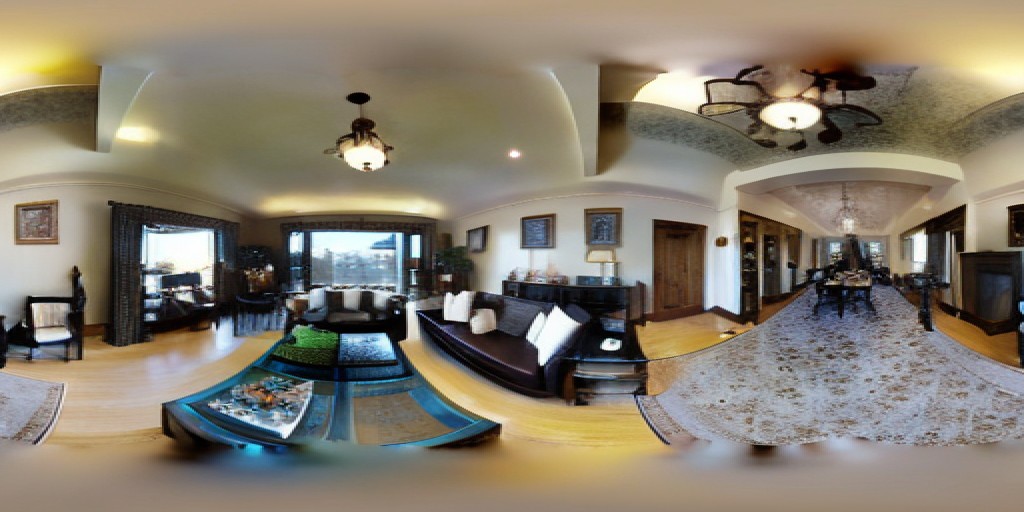}
        \caption{w/o depth}
    \end{subfigure}
    \hfill
    \begin{subfigure}[t]{0.13\textwidth}
        \includegraphics[width=\textwidth]{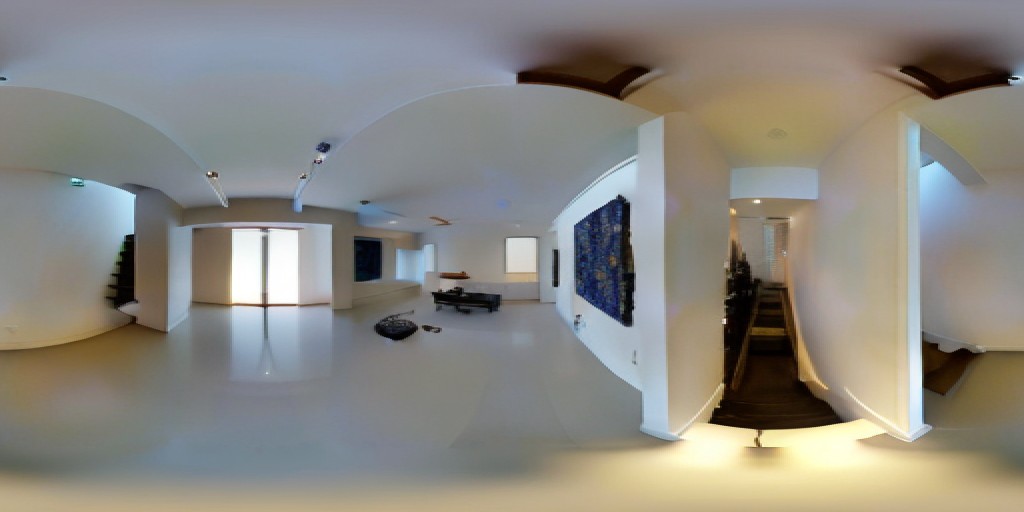}
        \includegraphics[width=\textwidth]{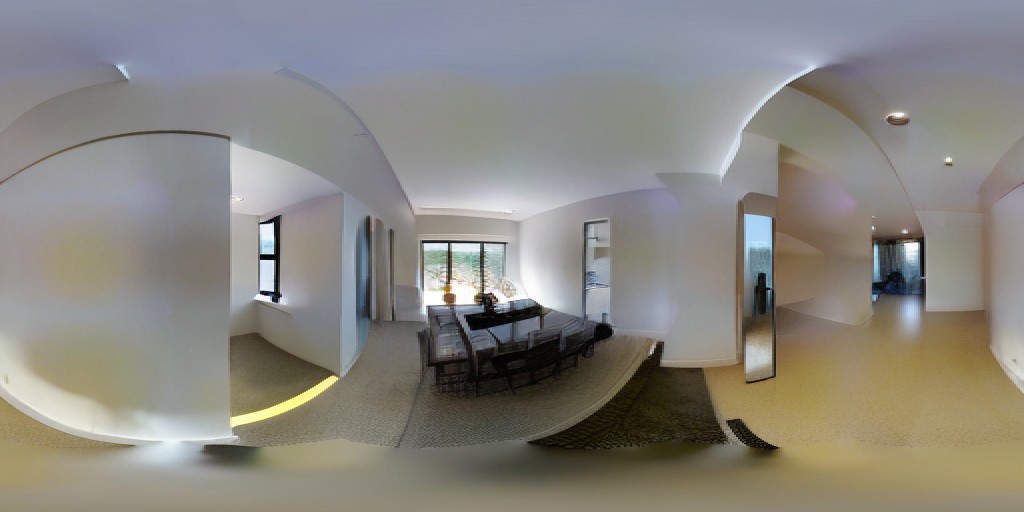}
        \includegraphics[width=\textwidth]{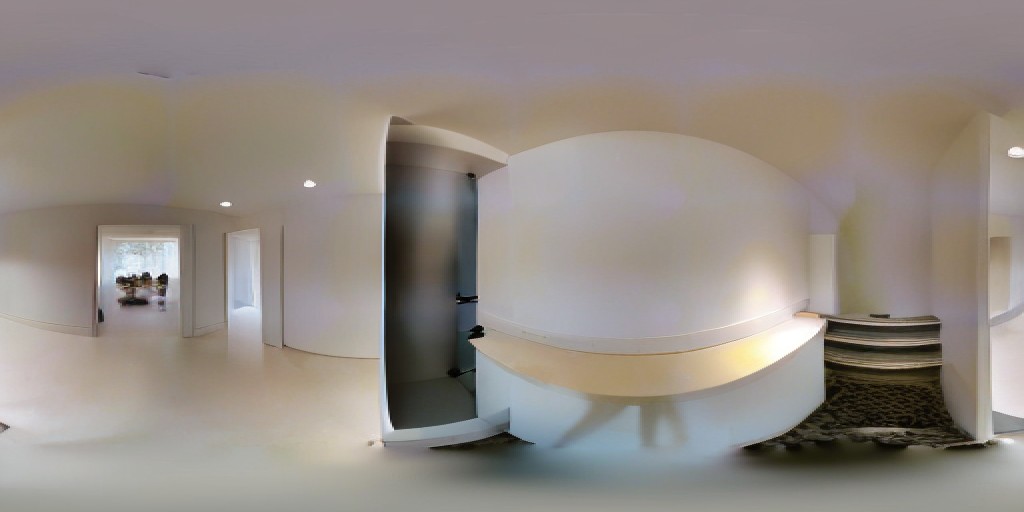}
        \includegraphics[width=\textwidth]{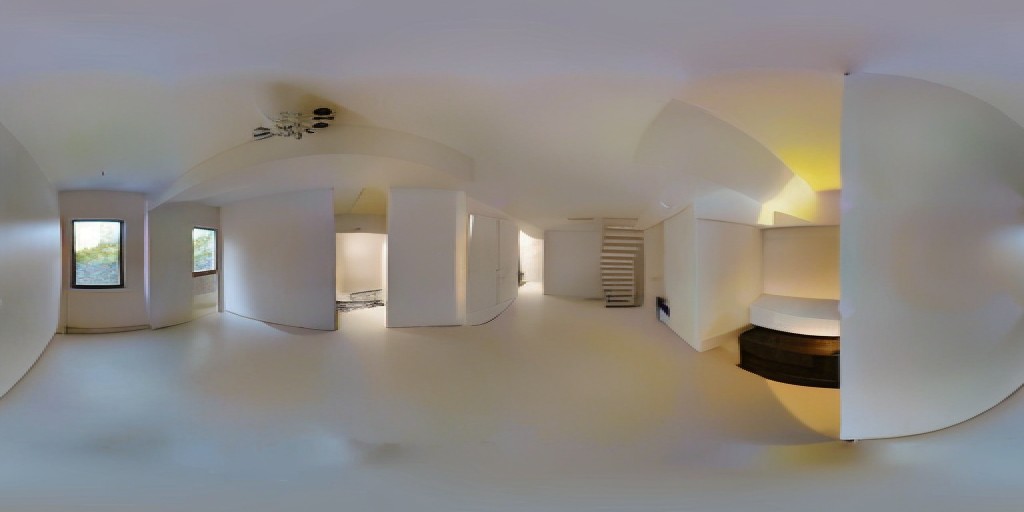}
        \includegraphics[width=\textwidth]{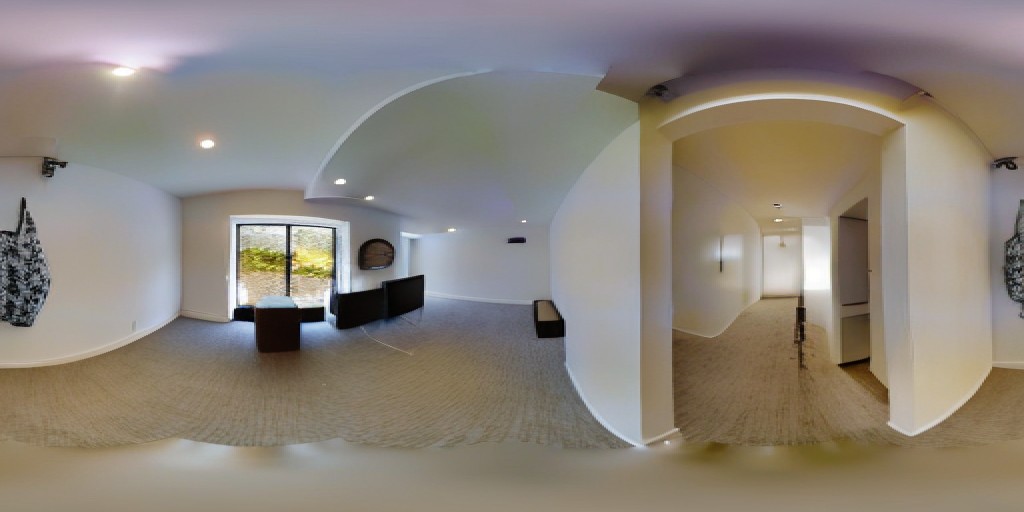}
        \includegraphics[width=\textwidth]{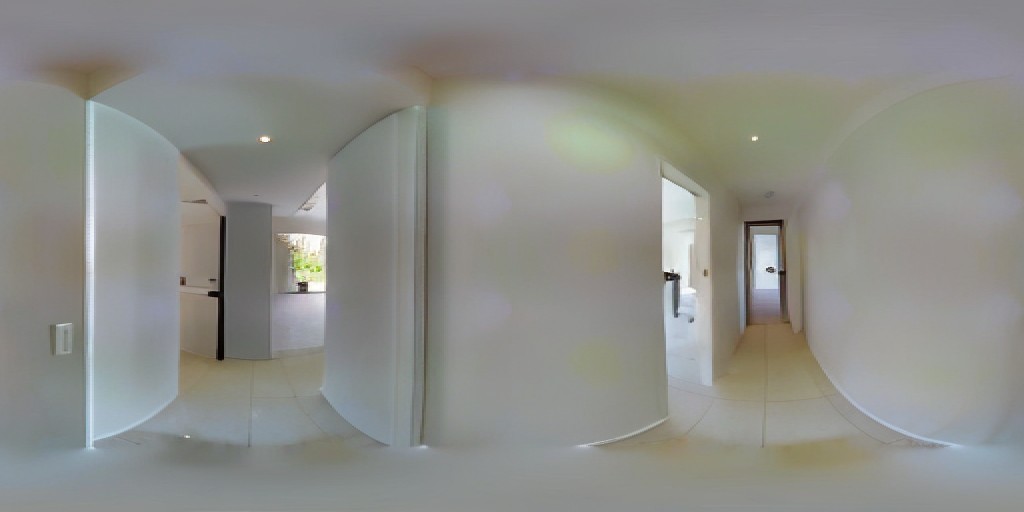}
        \includegraphics[width=\textwidth]{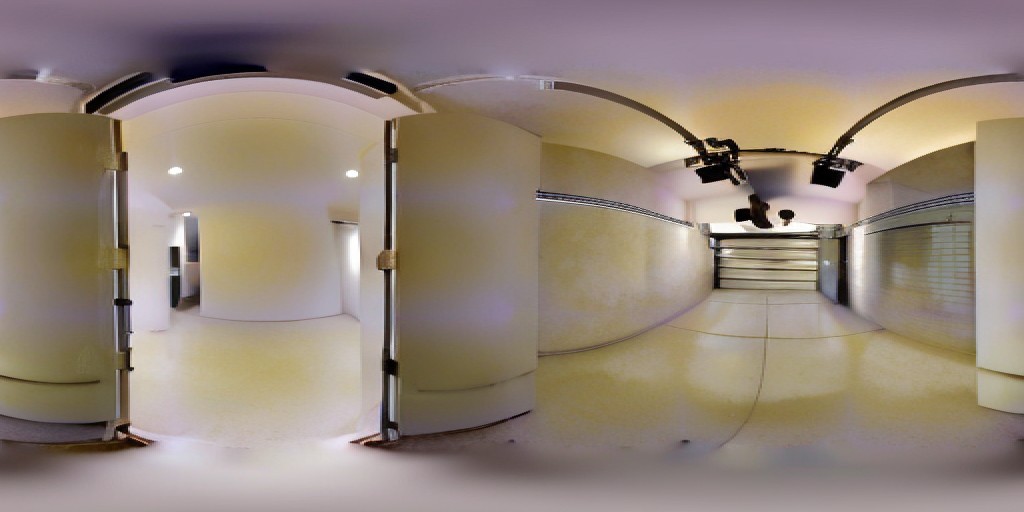}
        \includegraphics[width=\textwidth]{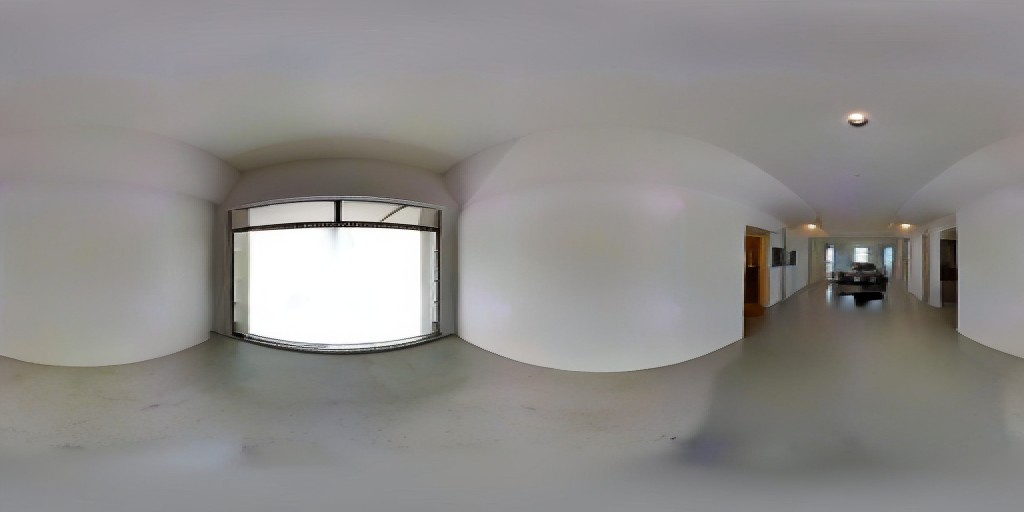}
        \caption{w/o color}
    \end{subfigure}
    \hfill
    \begin{subfigure}[t]{0.13\textwidth}
       \includegraphics[width=\textwidth]{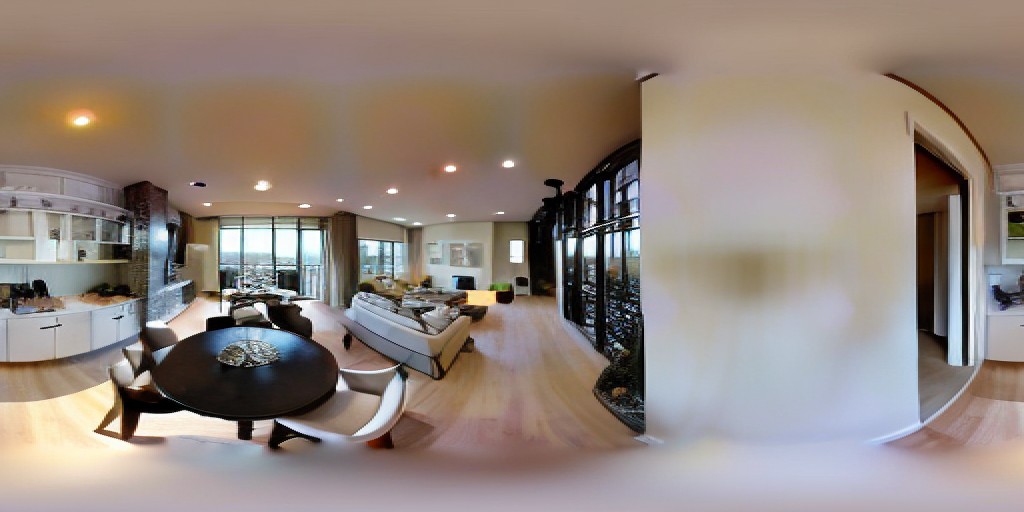}
        \includegraphics[width=\textwidth]{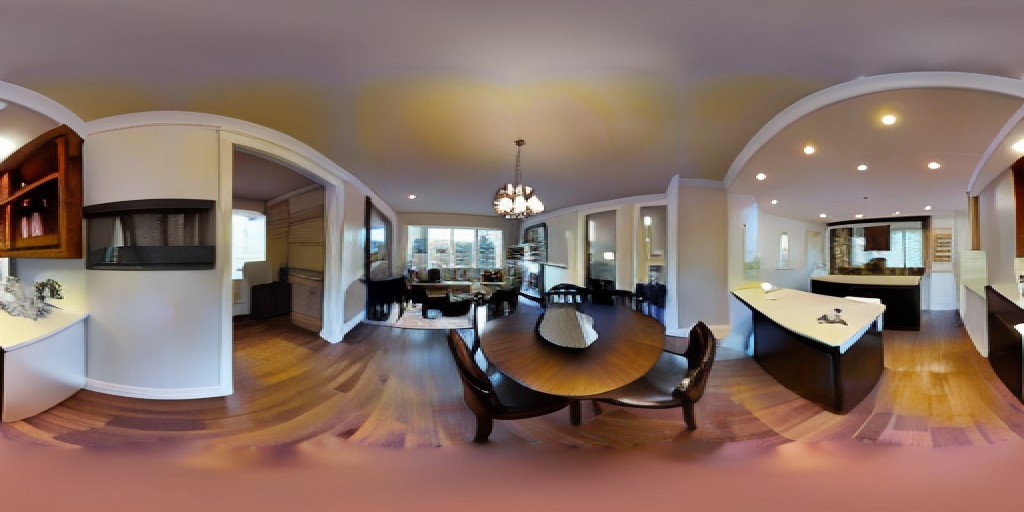}
        \includegraphics[width=\textwidth]{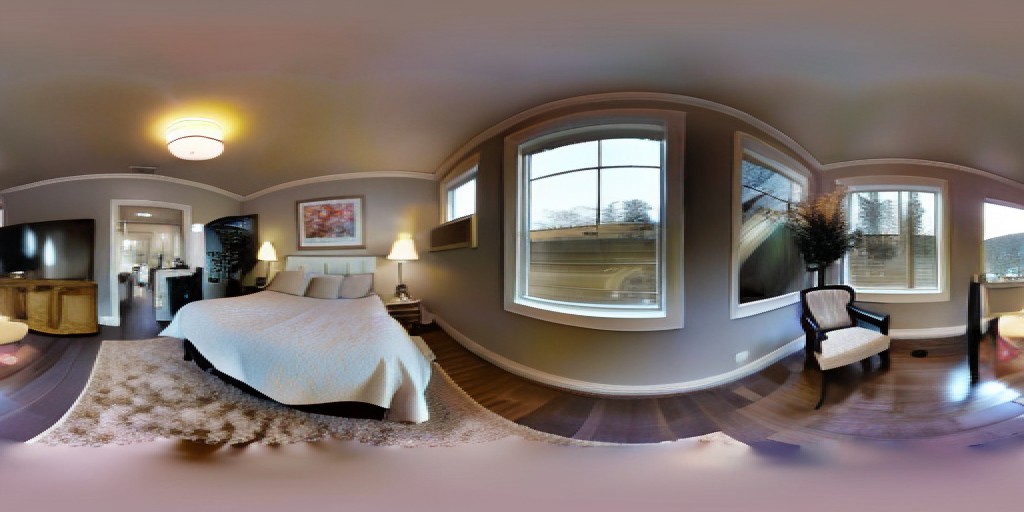}
        \includegraphics[width=\textwidth]{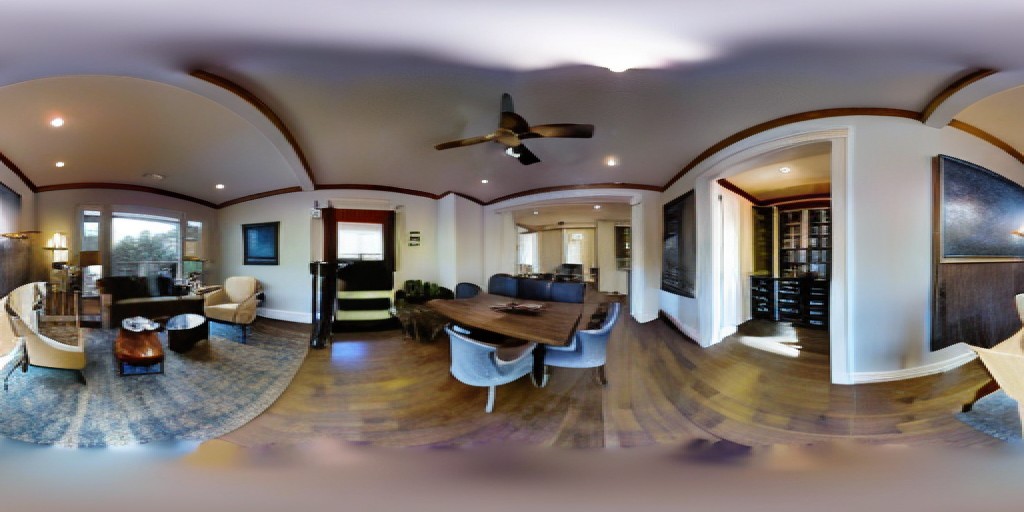}
        \includegraphics[width=\textwidth]{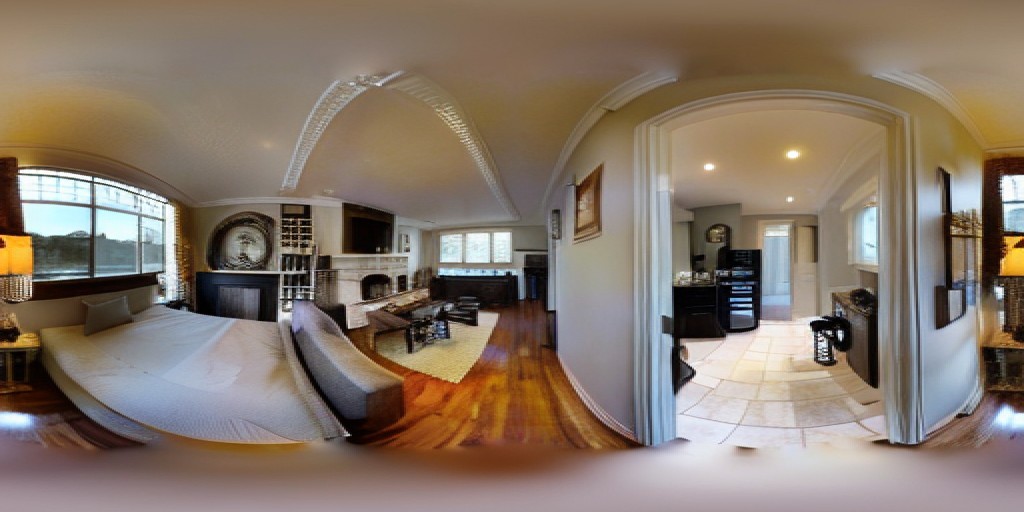}
        \includegraphics[width=\textwidth]{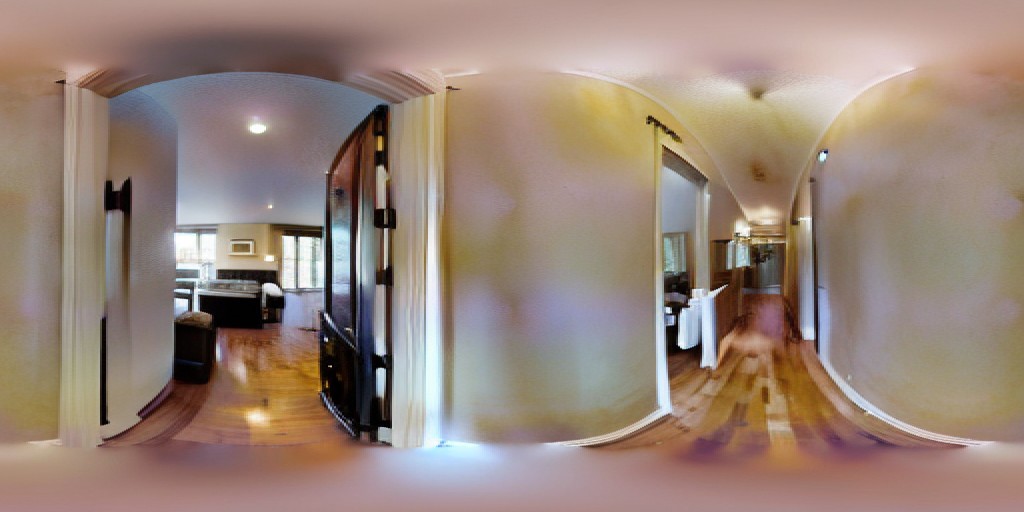}
        \includegraphics[width=\textwidth]{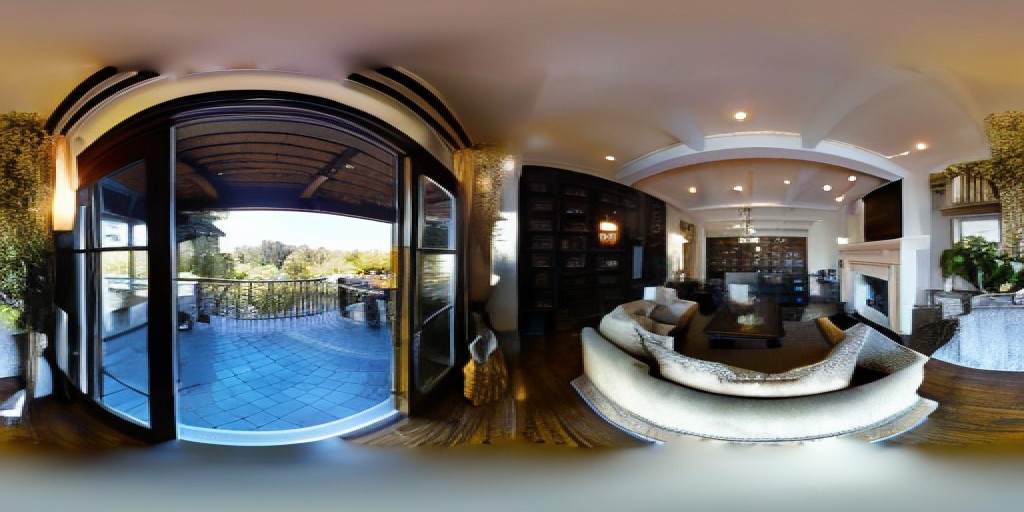}
        \includegraphics[width=\textwidth]{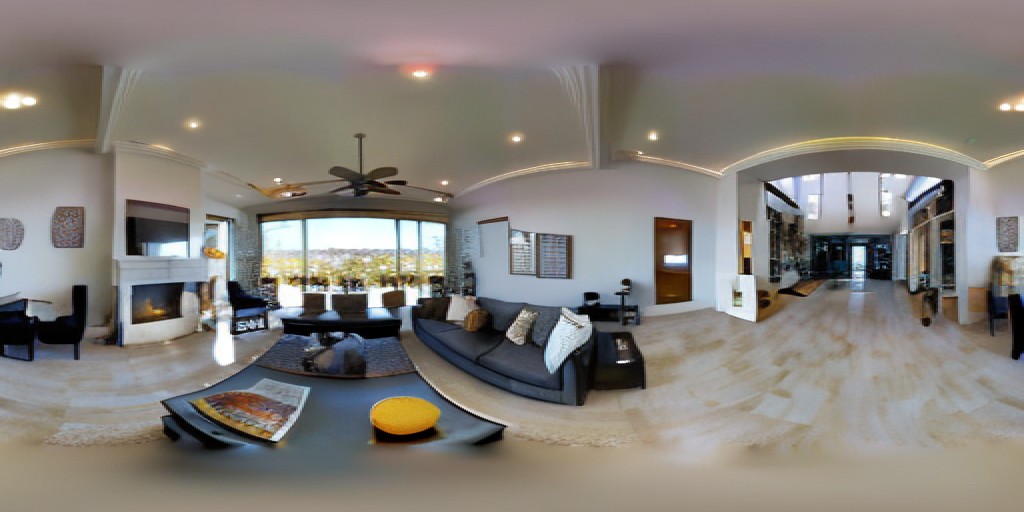}
        \caption{full model}
    \end{subfigure}
    \hfill
    \begin{subfigure}[t]{0.13\textwidth}
        \includegraphics[width=\textwidth]{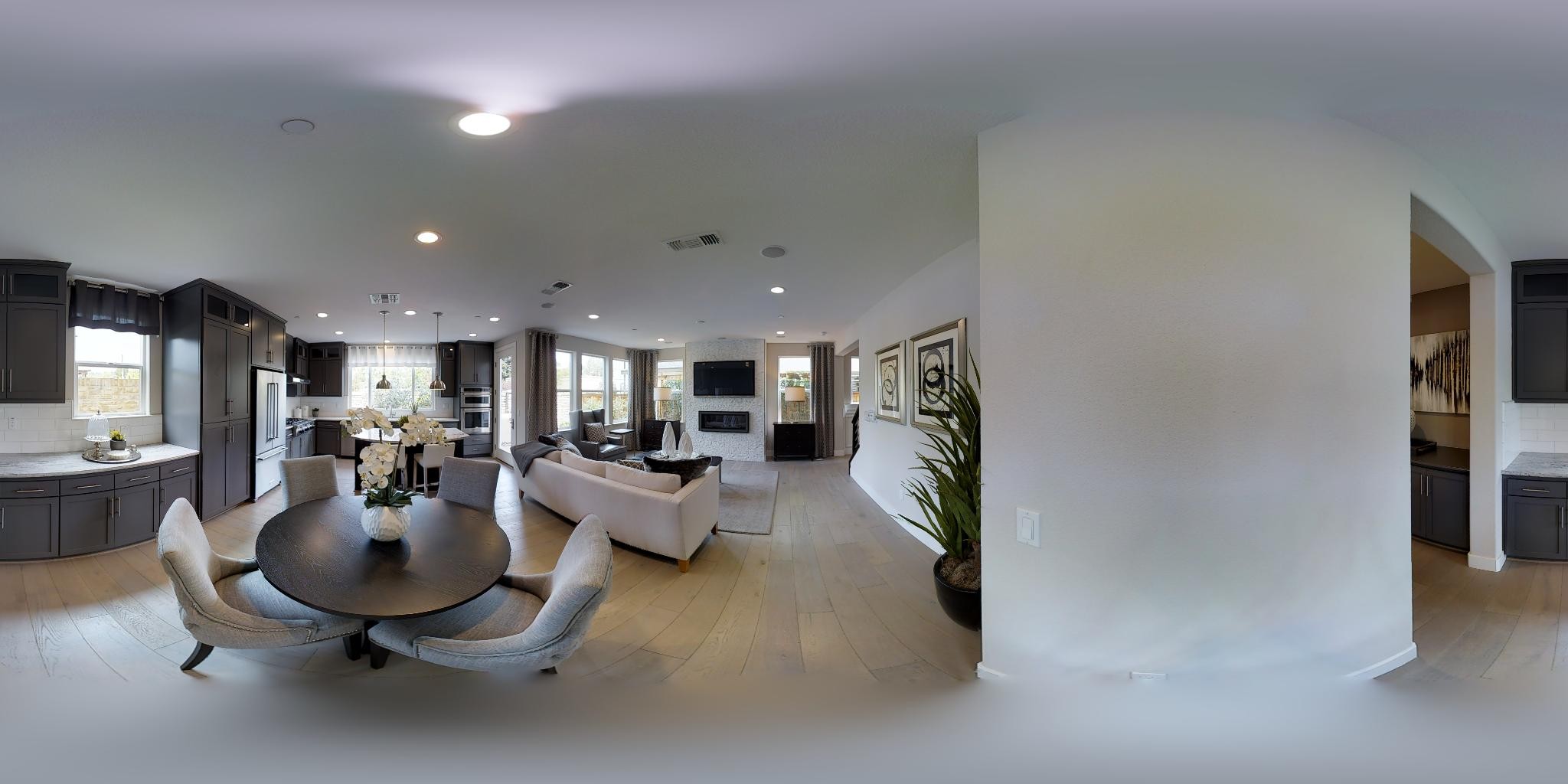}
        \includegraphics[width=\textwidth]{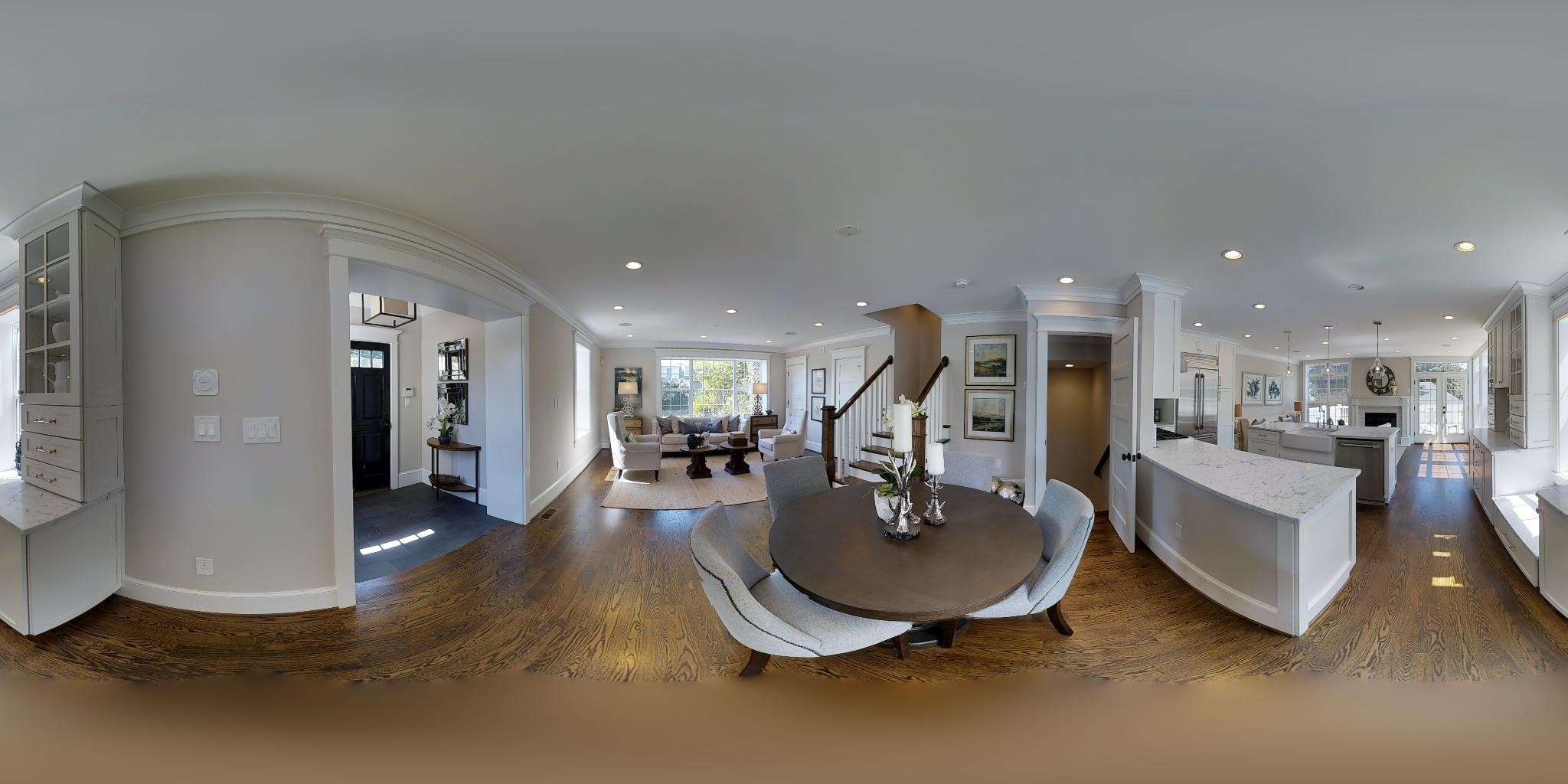}
        \includegraphics[width=\textwidth]{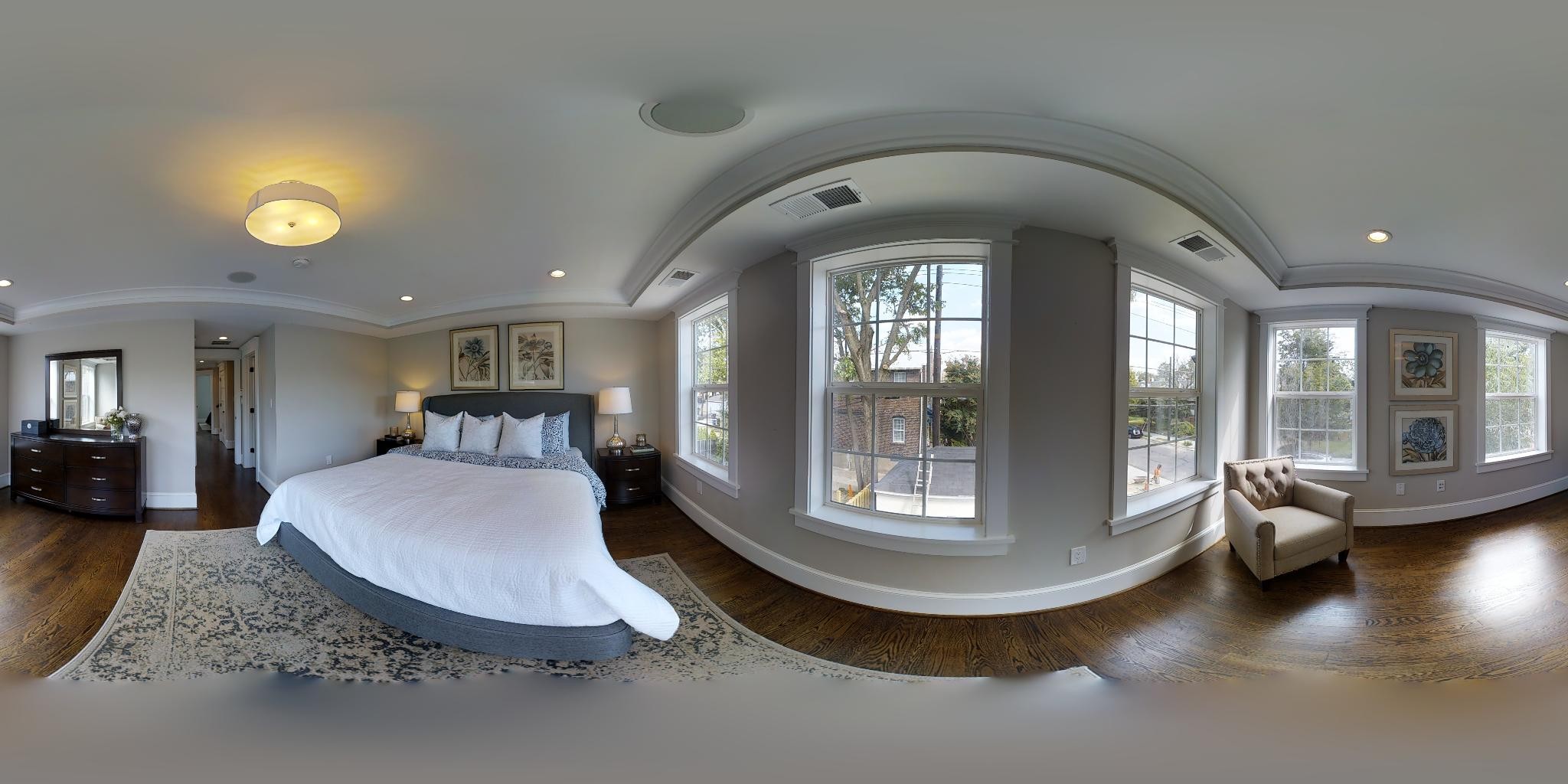}
        \includegraphics[width=\textwidth]{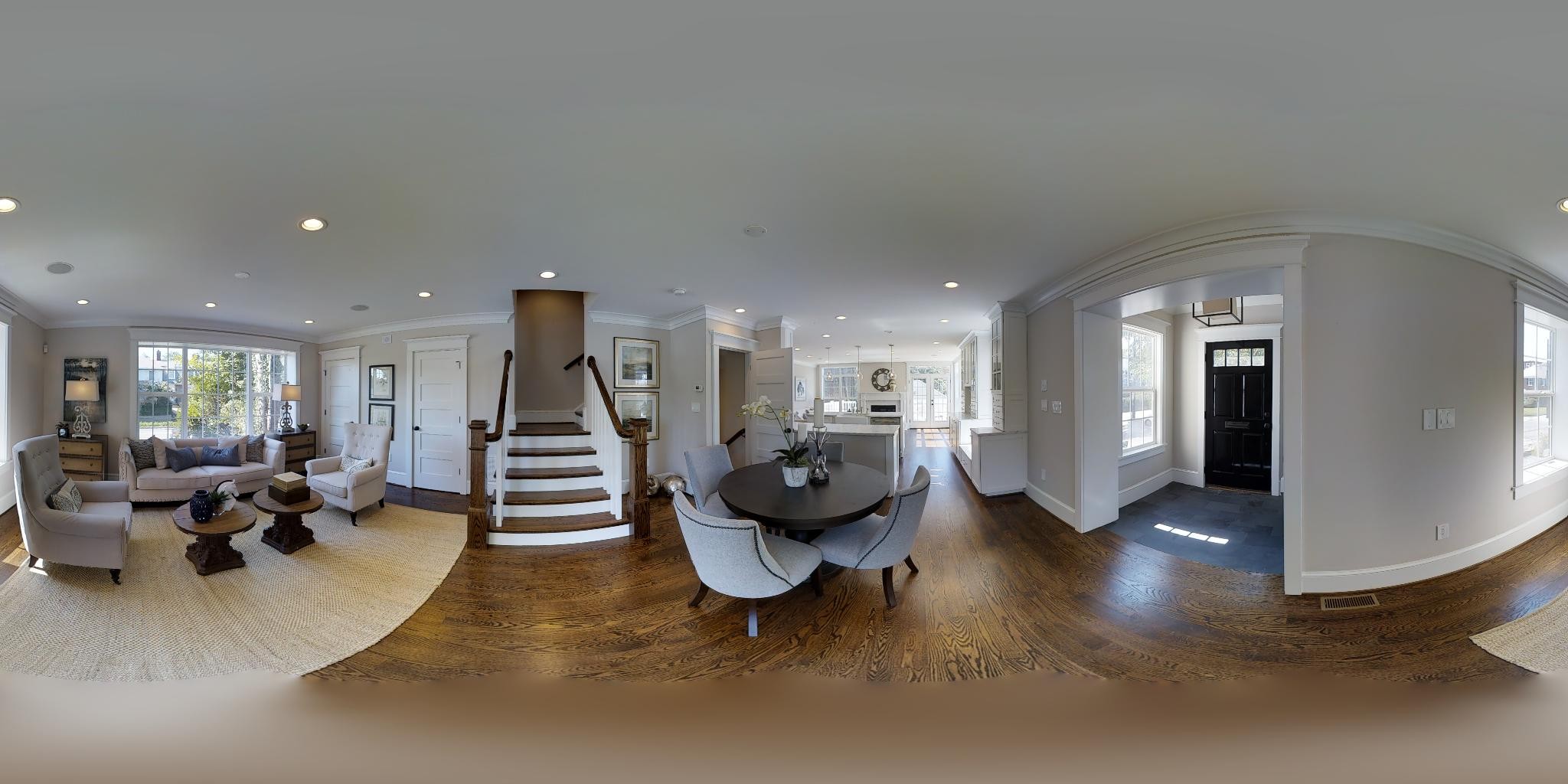}
        \includegraphics[width=\textwidth]{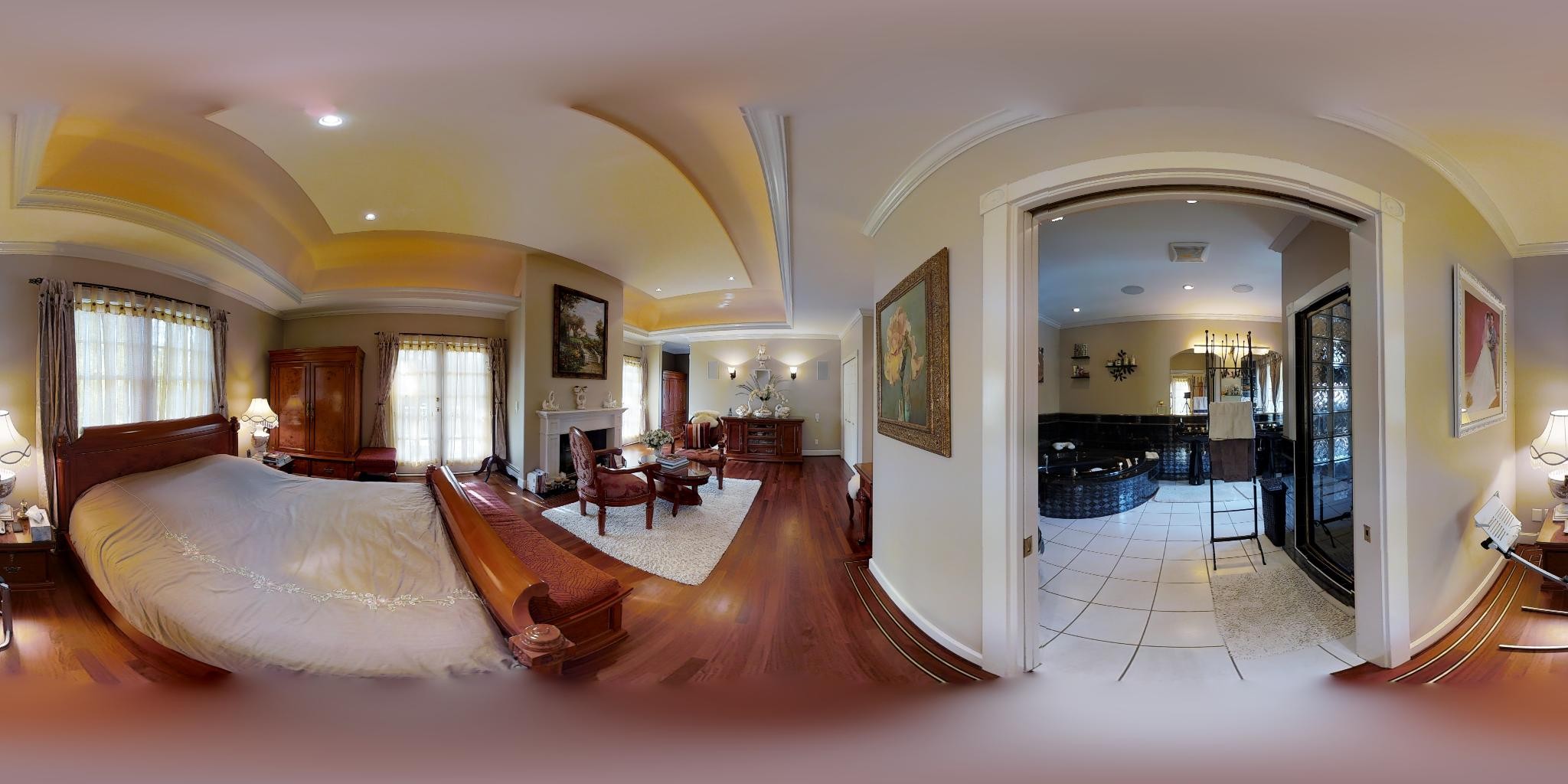}
        \includegraphics[width=\textwidth]{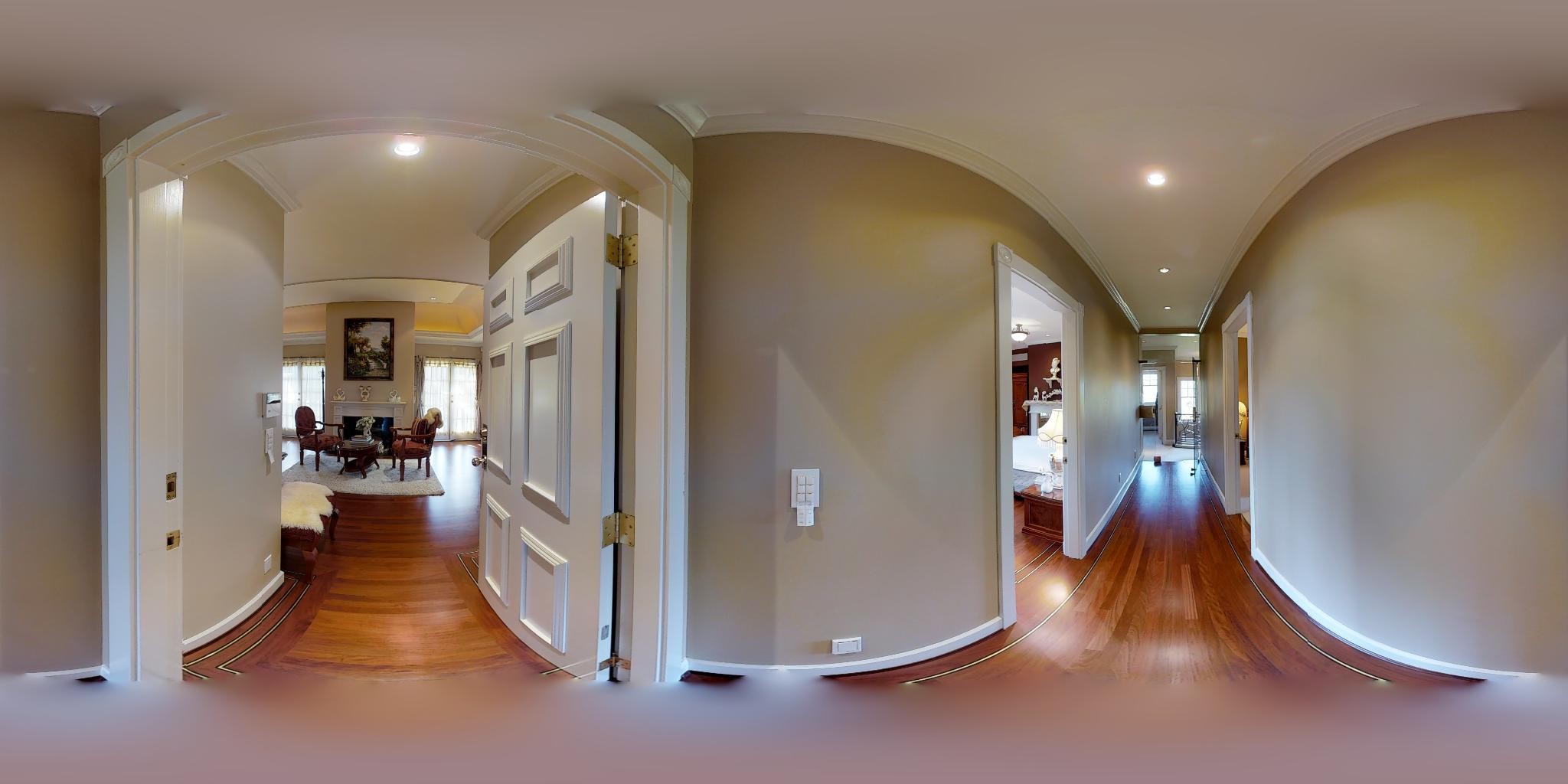}
        \includegraphics[width=\textwidth]{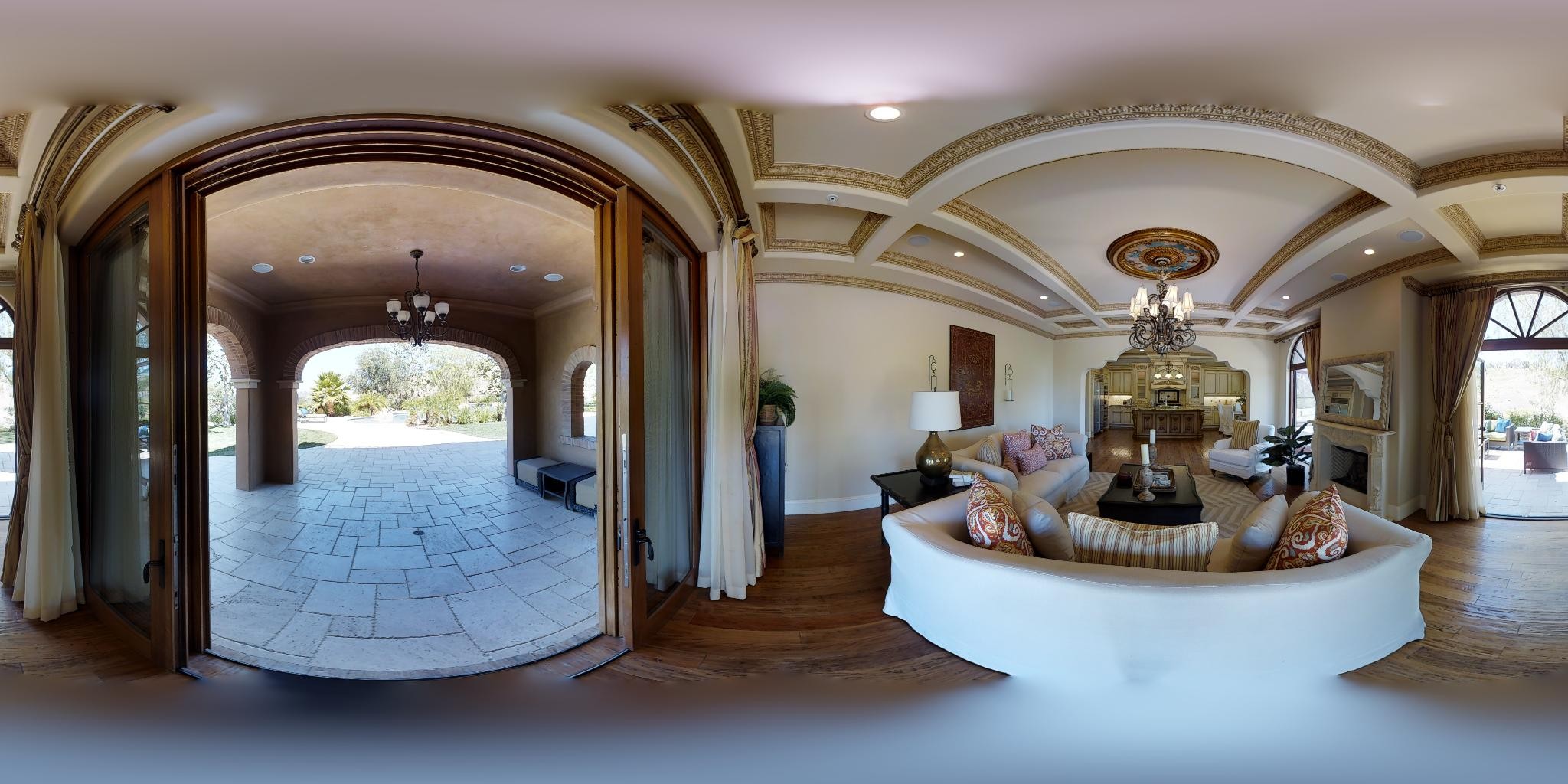}
        \includegraphics[width=\textwidth]{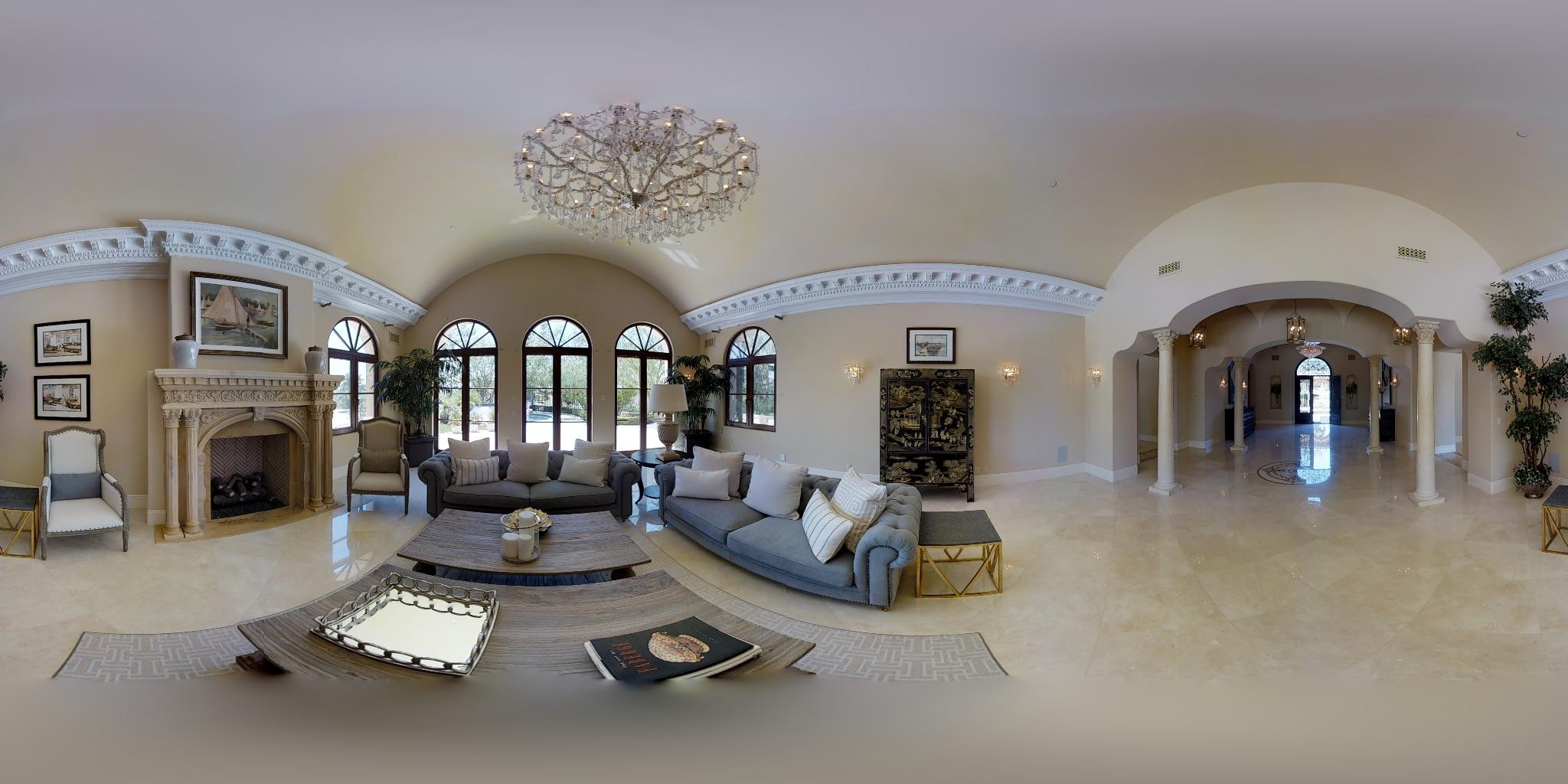}
        \caption{ground truth}
    \end{subfigure}
    \vspace{-2mm}
    \caption{\small Qualitative results of ablation experiments on the Matterport3D dataset.}
    \label{fig:Mablation}
\end{figure*}

\vspace{-2mm}

\begin{figure*}[h]
    \centering 
    \begin{subfigure}[t]{0.13\textwidth}
        \includegraphics[width=\textwidth]{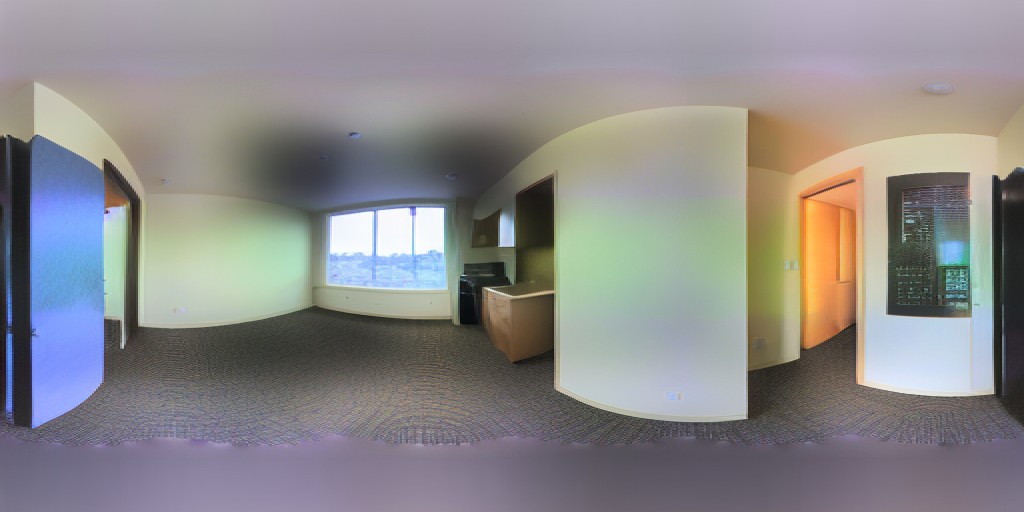}
        \includegraphics[width=\textwidth]{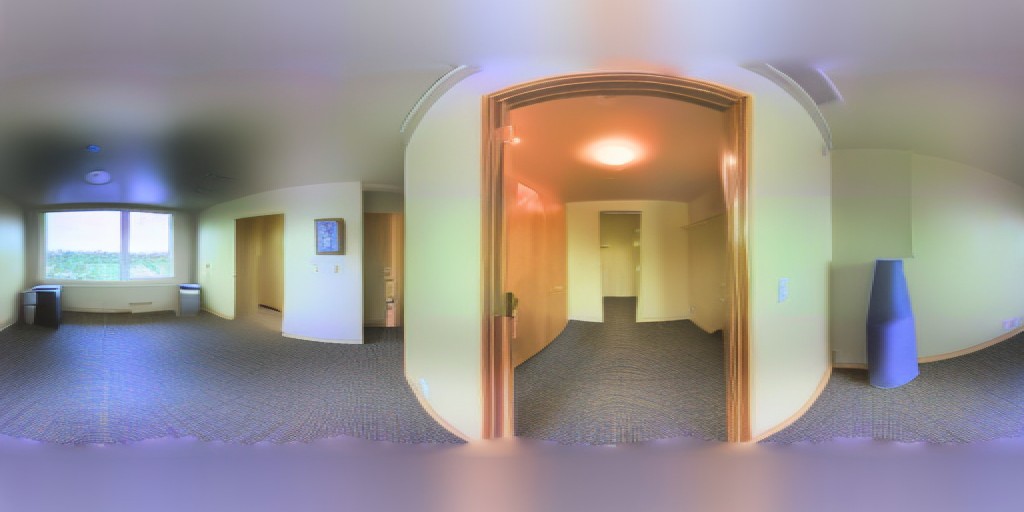}
        \includegraphics[width=\textwidth]{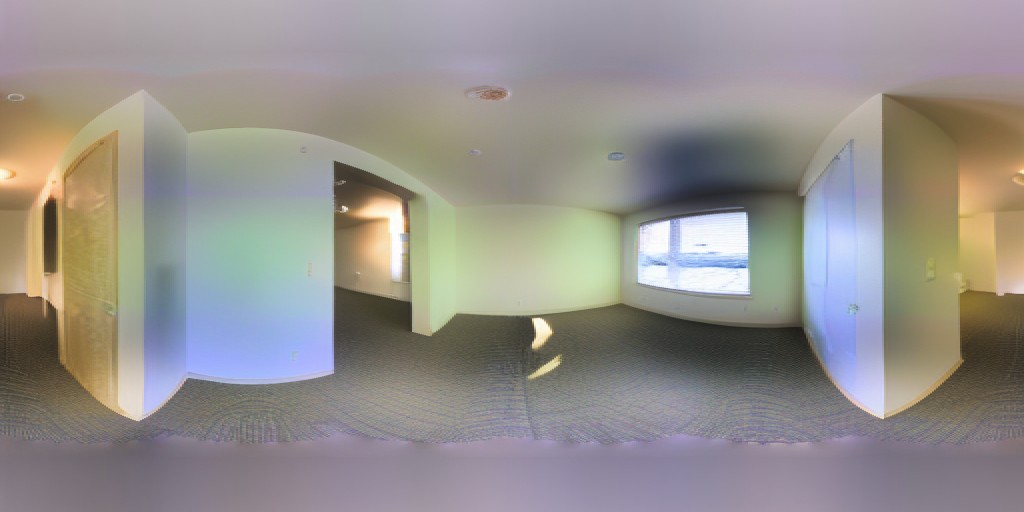}
        \includegraphics[width=\textwidth]{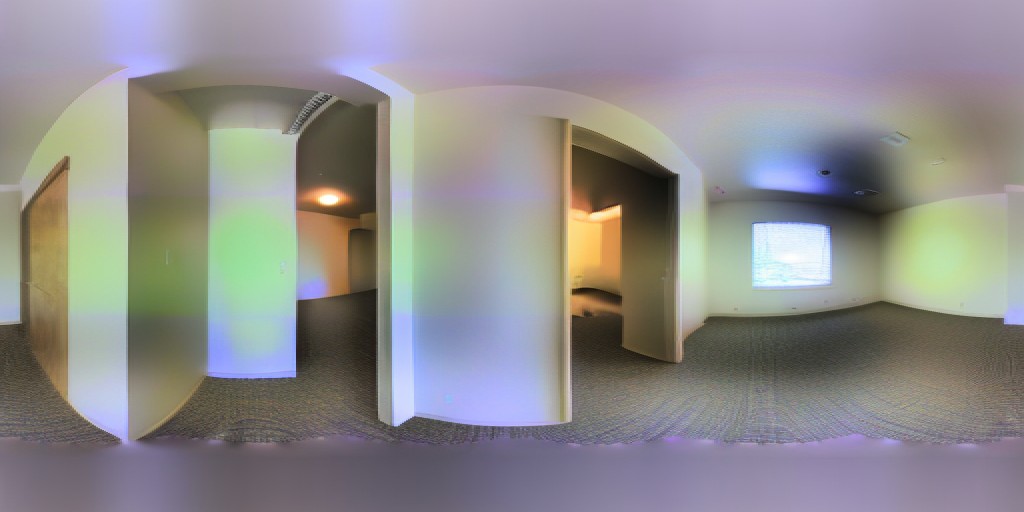}
        \includegraphics[width=\textwidth]{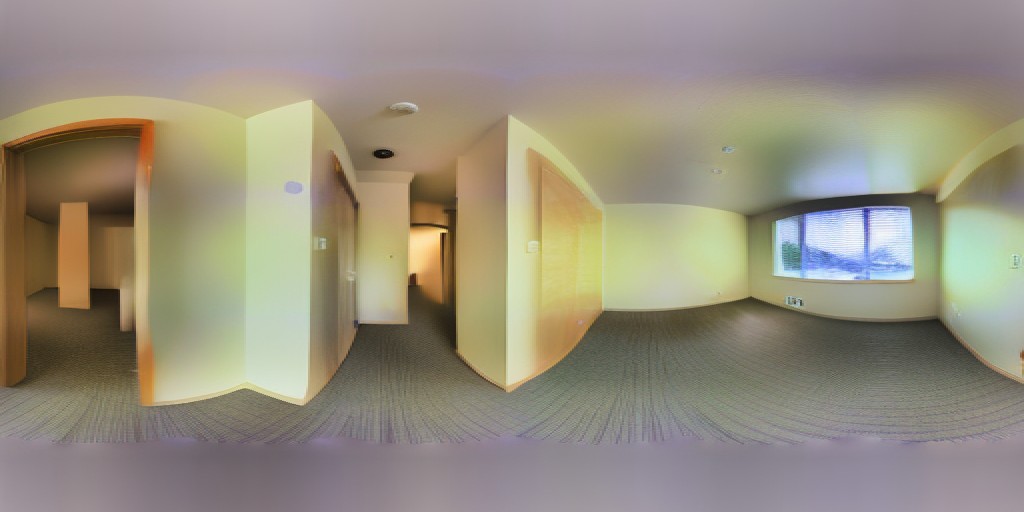}
        \includegraphics[width=\textwidth]{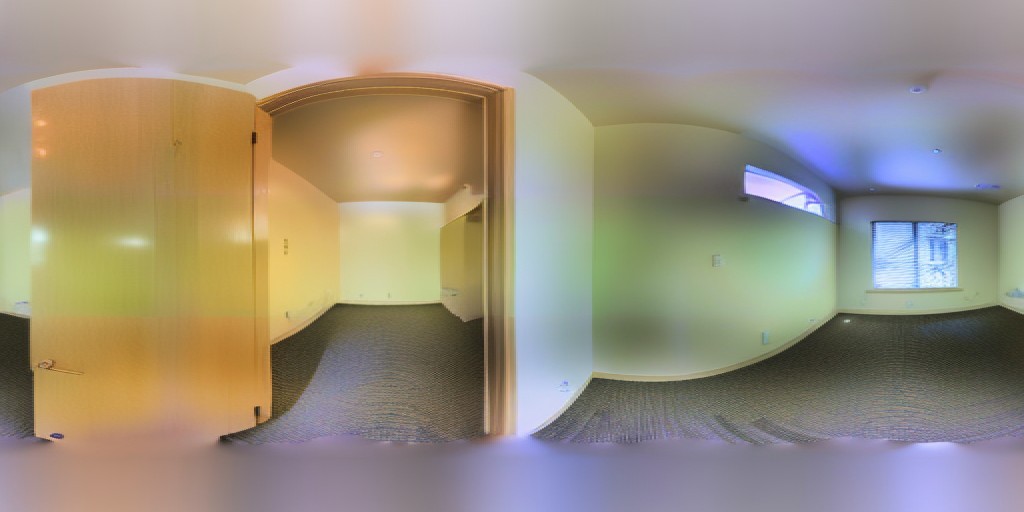}
        \includegraphics[width=\textwidth]{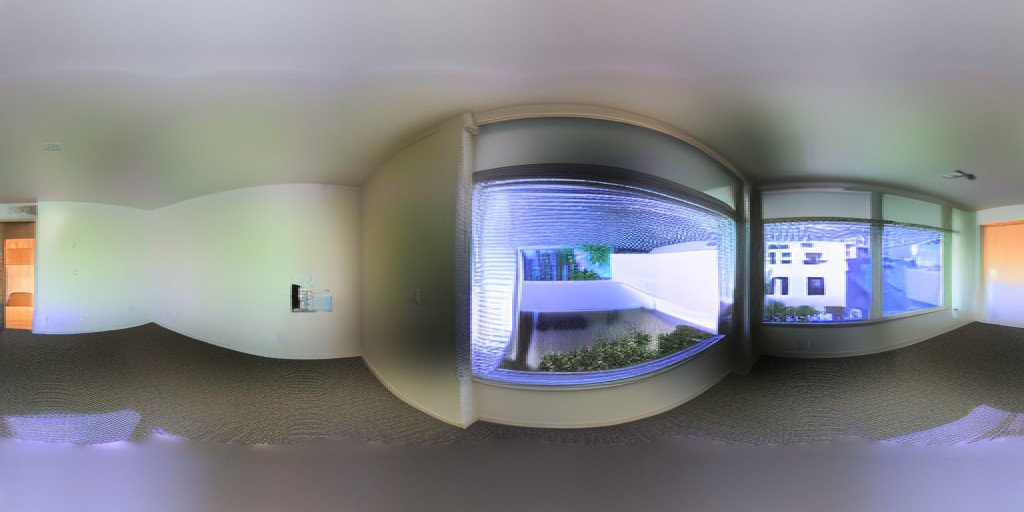}
        \includegraphics[width=\textwidth]{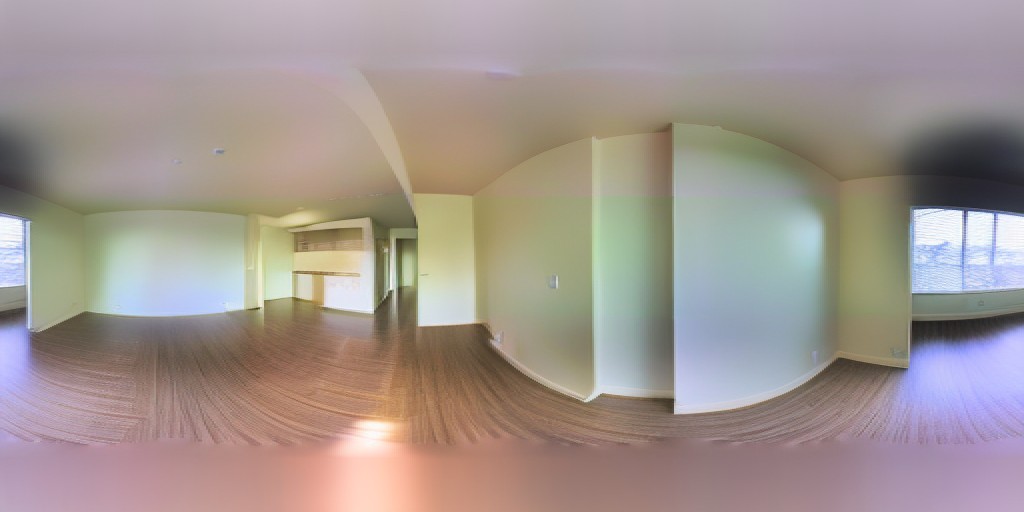}
        \caption{w/o floor}
    \end{subfigure}
    \hfill
    \begin{subfigure}[t]{0.13\textwidth}
        \includegraphics[width=\textwidth]{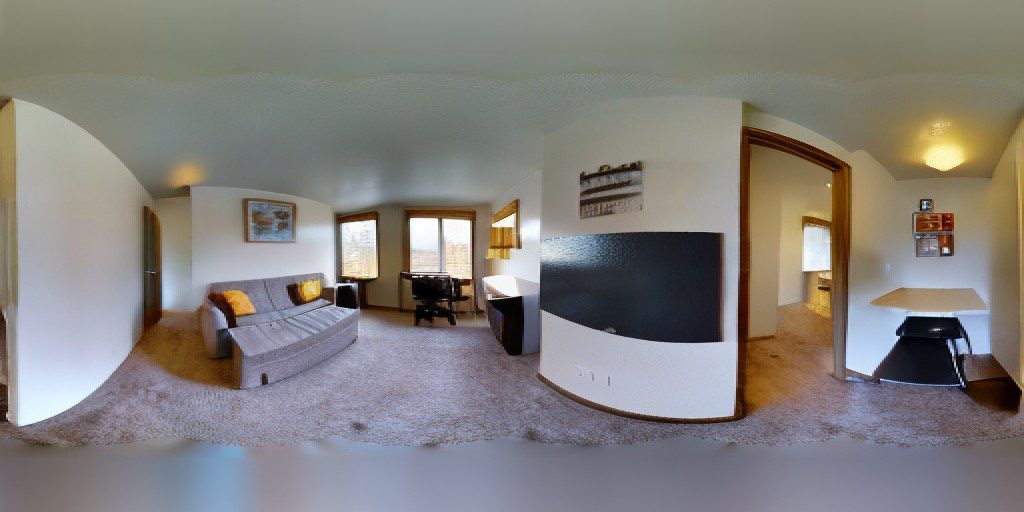}
        \includegraphics[width=\textwidth]{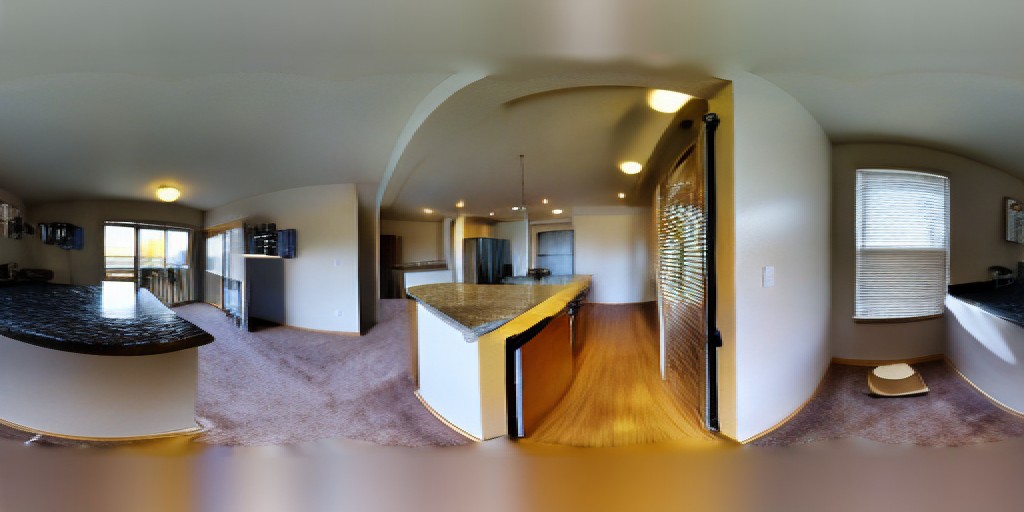}
        \includegraphics[width=\textwidth]{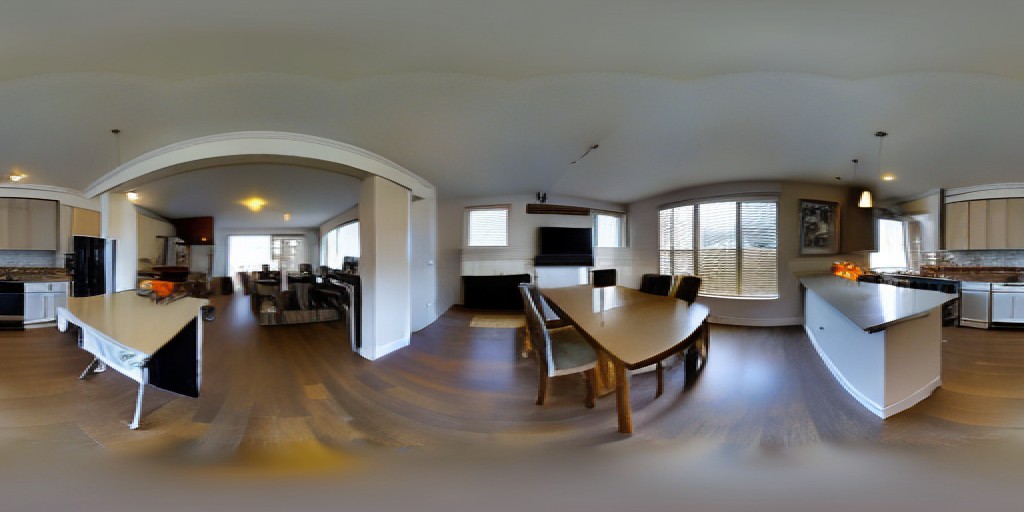}
        \includegraphics[width=\textwidth]{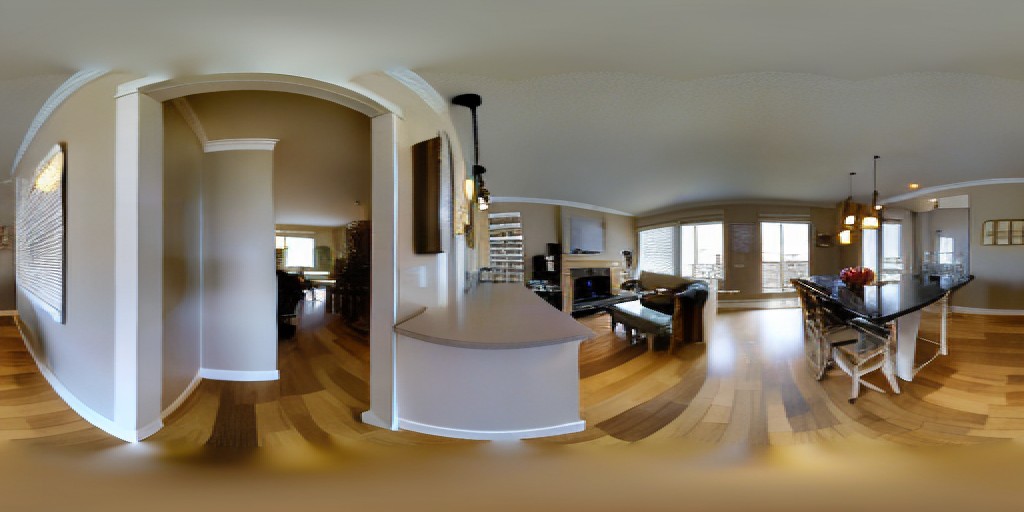}
        \includegraphics[width=\textwidth]{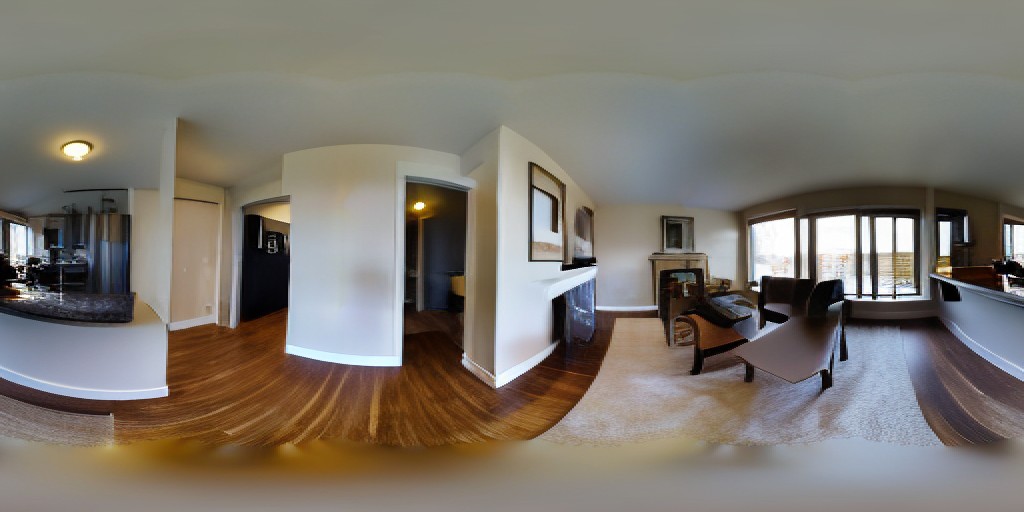}
        \includegraphics[width=\textwidth]{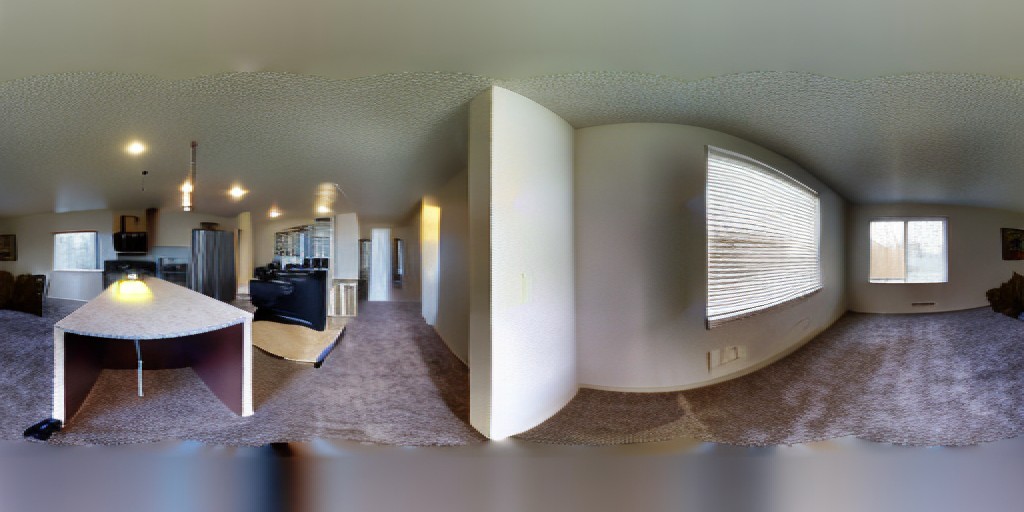}
        \includegraphics[width=\textwidth]{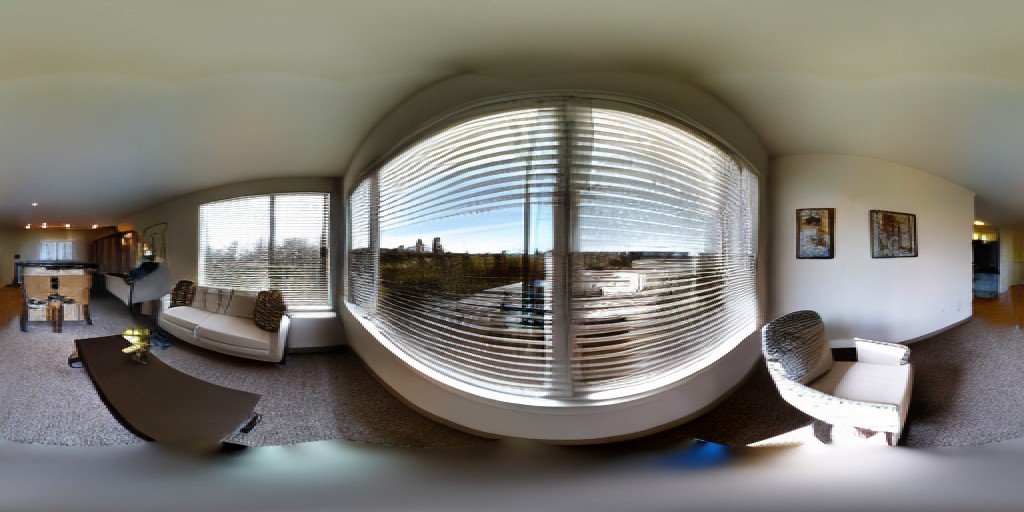}
        \includegraphics[width=\textwidth]{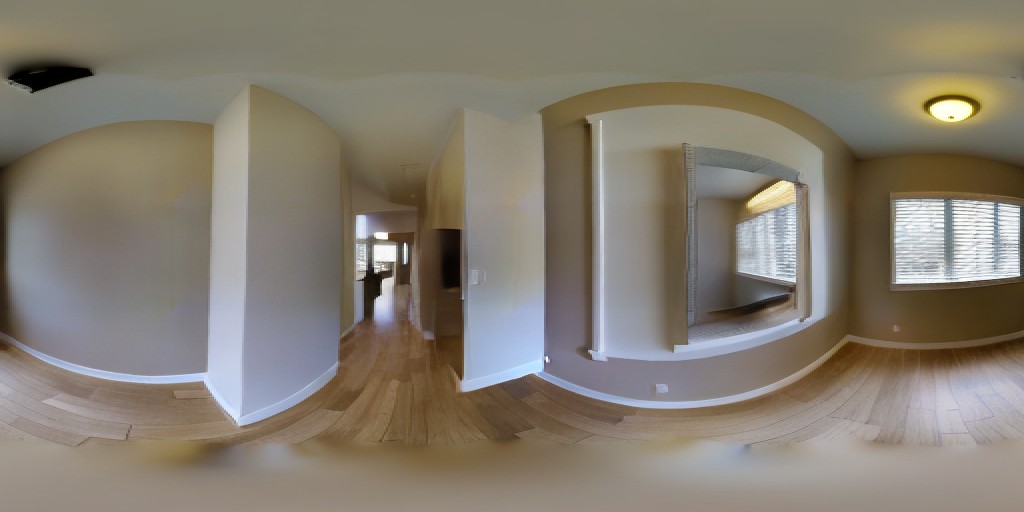}
        \caption{w/o wall}
    \end{subfigure}
    \hfill
    \begin{subfigure}[t]{0.13\textwidth}
        \includegraphics[width=\textwidth]{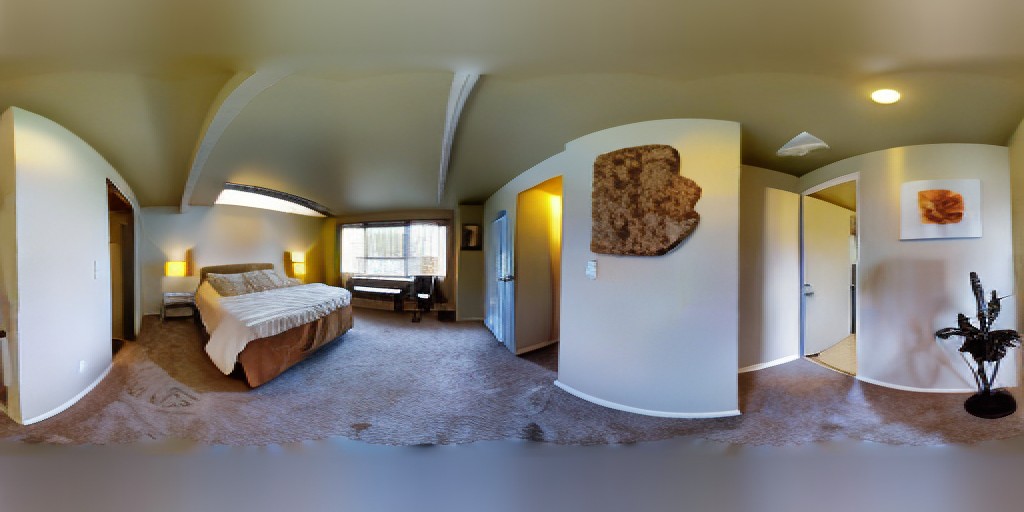}
        \includegraphics[width=\textwidth]{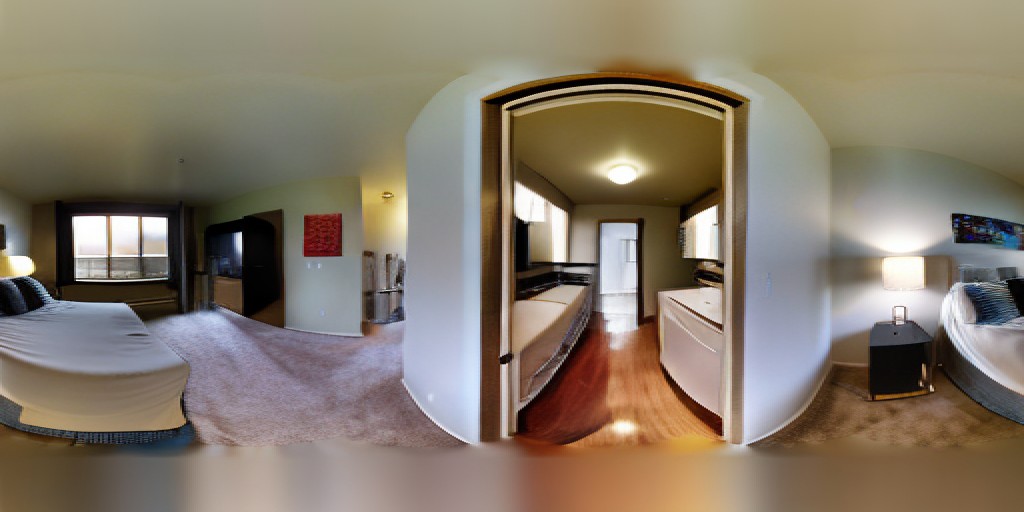}
        \includegraphics[width=\textwidth]{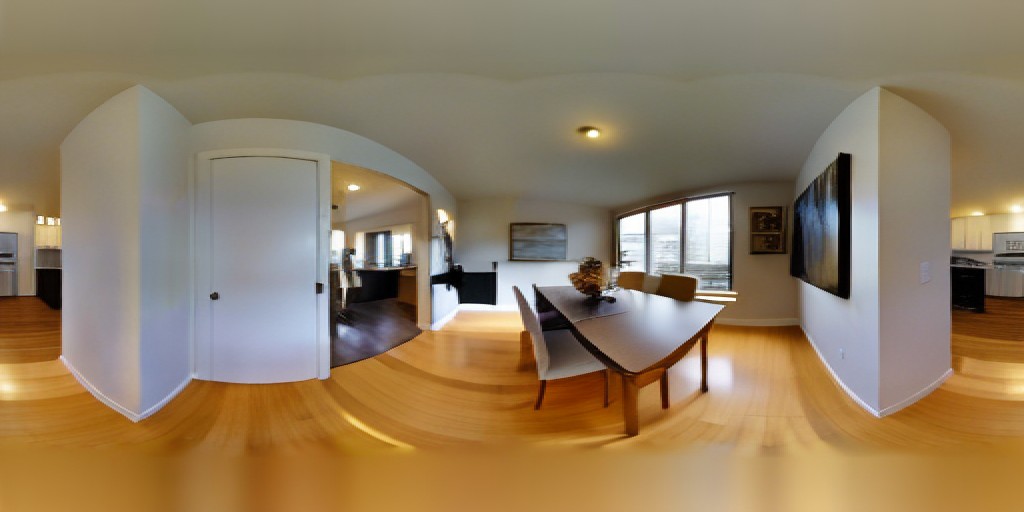}
        \includegraphics[width=\textwidth]{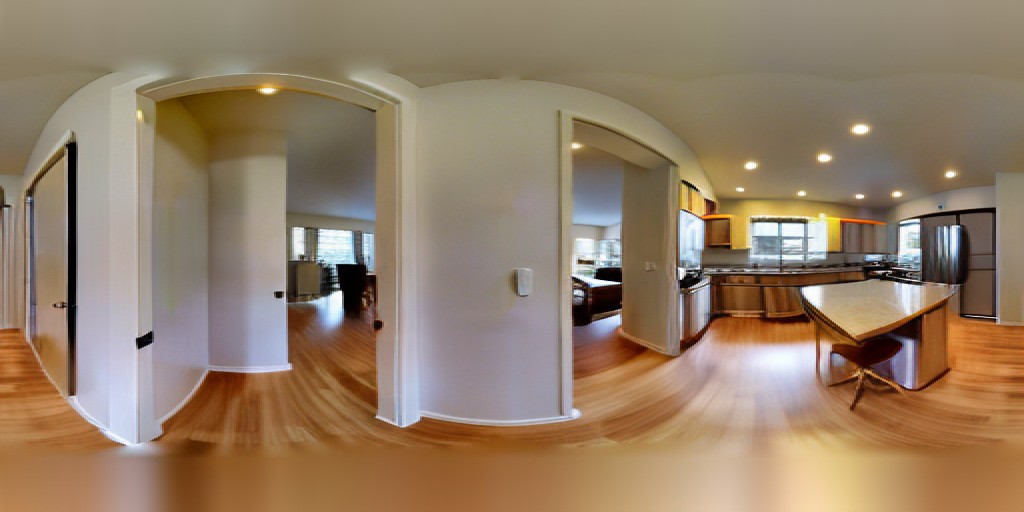}
        \includegraphics[width=\textwidth]{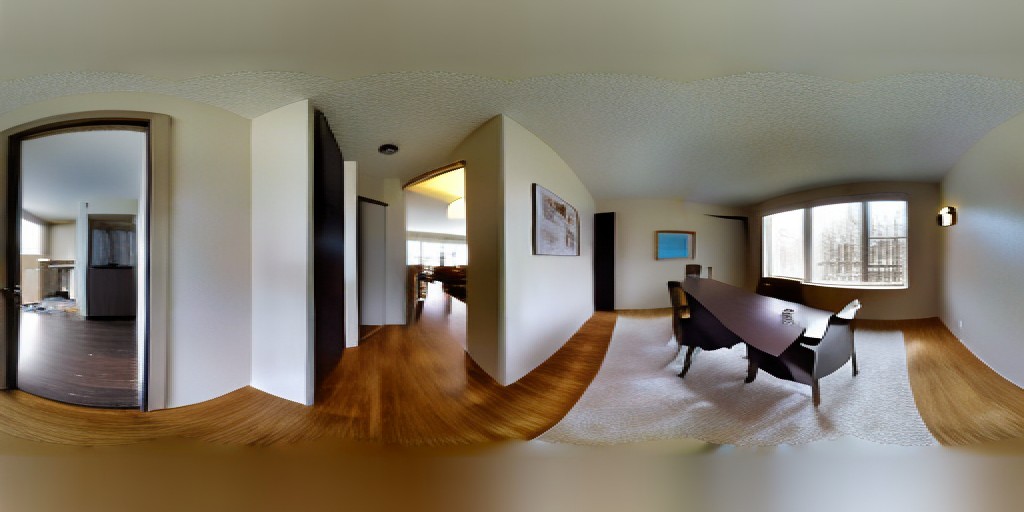}
        \includegraphics[width=\textwidth]{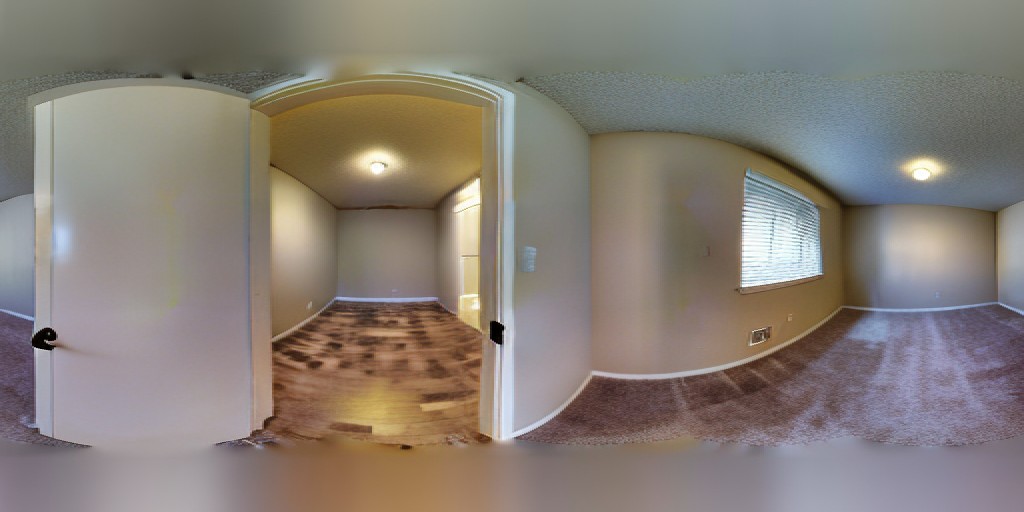}
        \includegraphics[width=\textwidth]{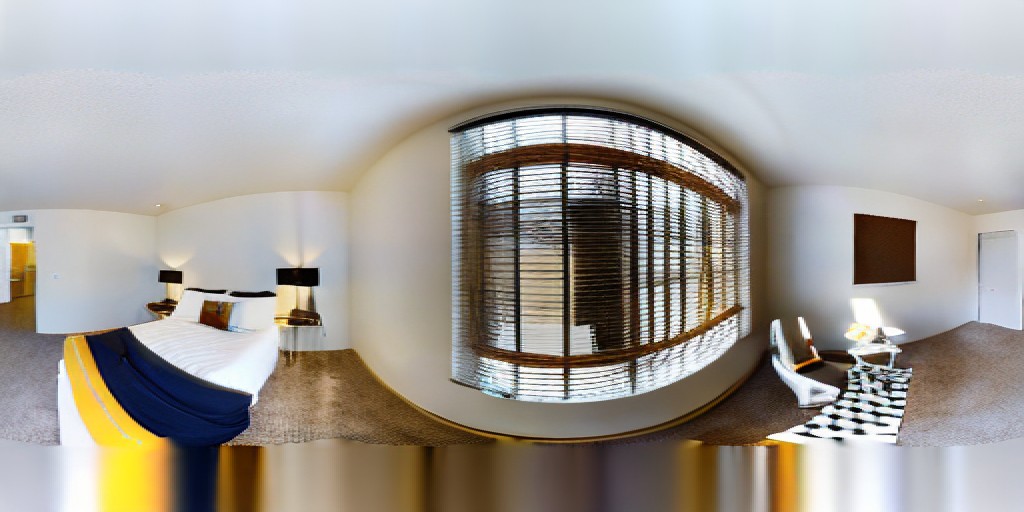}
        \includegraphics[width=\textwidth]{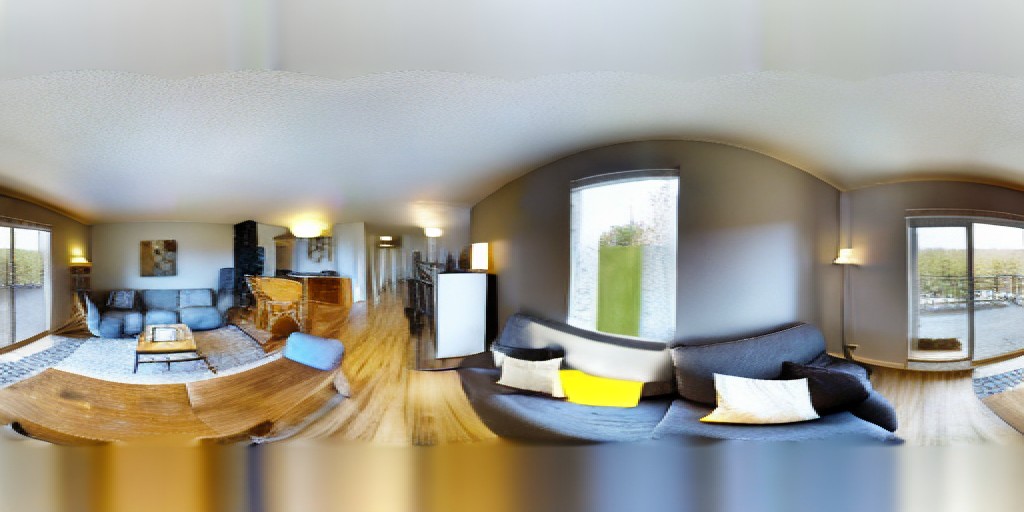}
        \caption{w/o segment}
    \end{subfigure}
    \hfill
    \begin{subfigure}[t]{0.13\textwidth}
        \includegraphics[width=\textwidth]{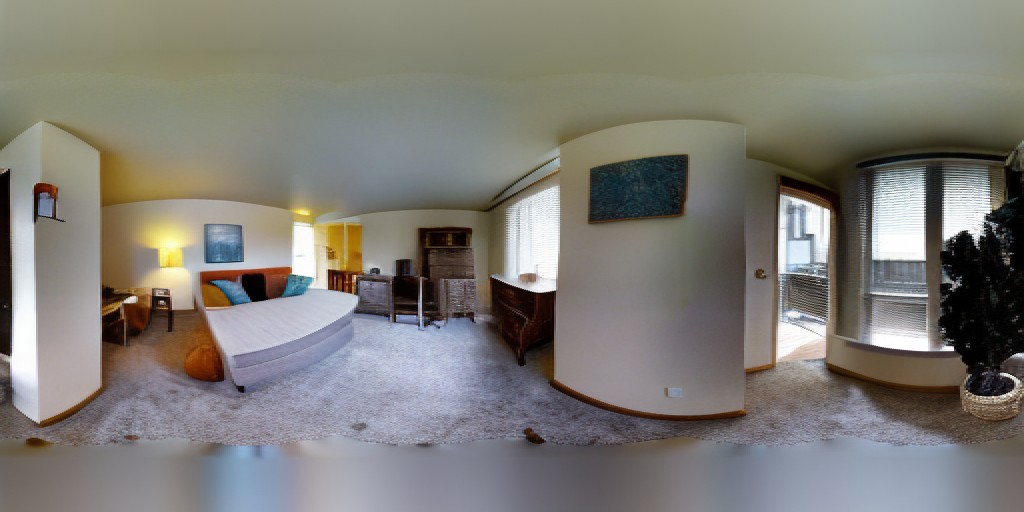}
        \includegraphics[width=\textwidth]{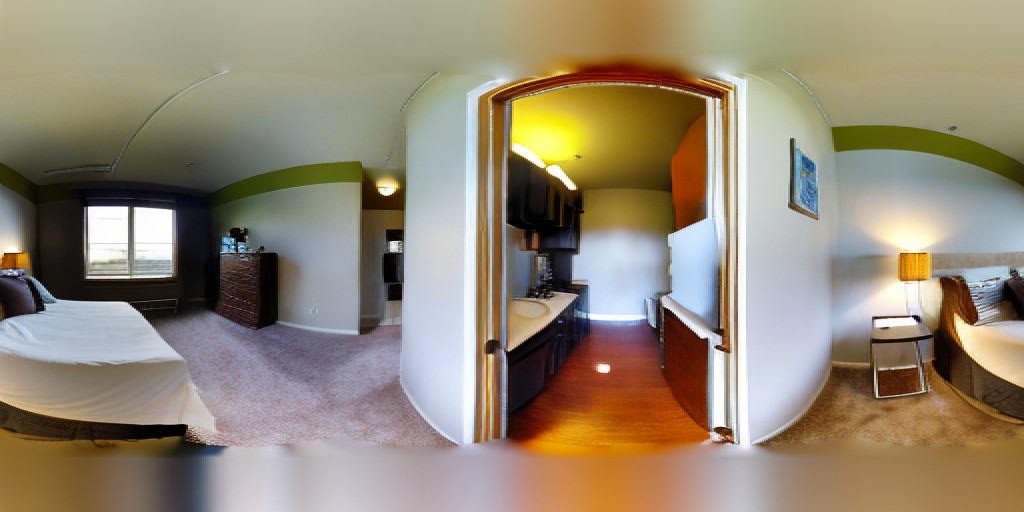}
        \includegraphics[width=\textwidth]{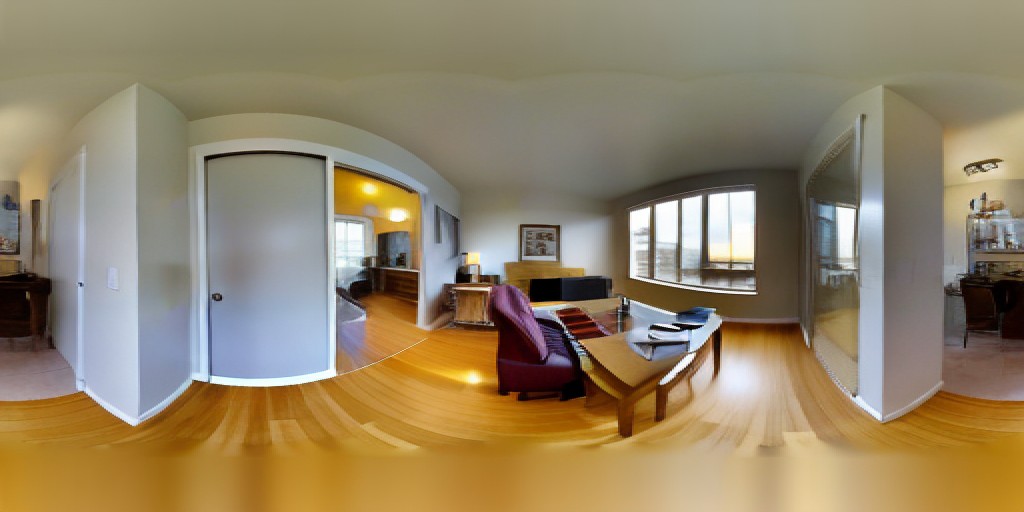}
        \includegraphics[width=\textwidth]{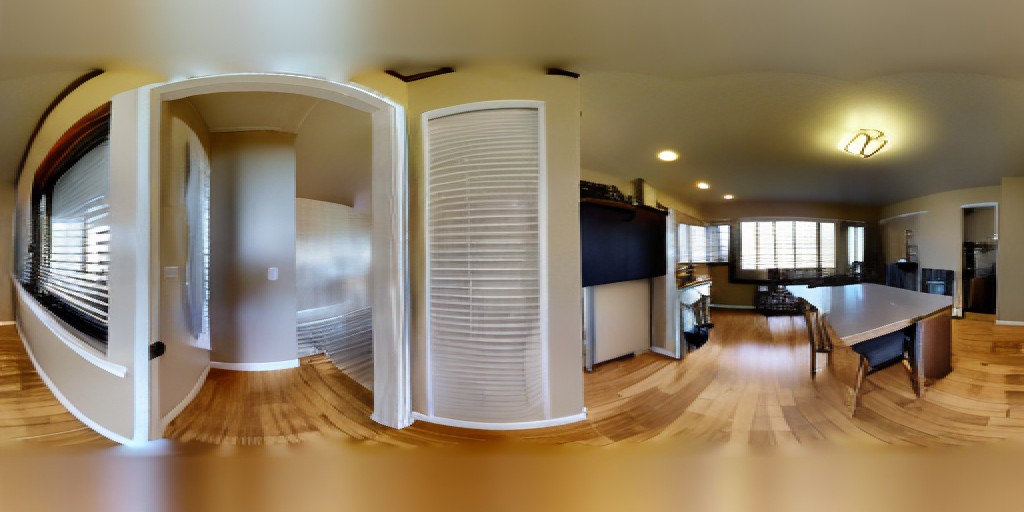}
        \includegraphics[width=\textwidth]{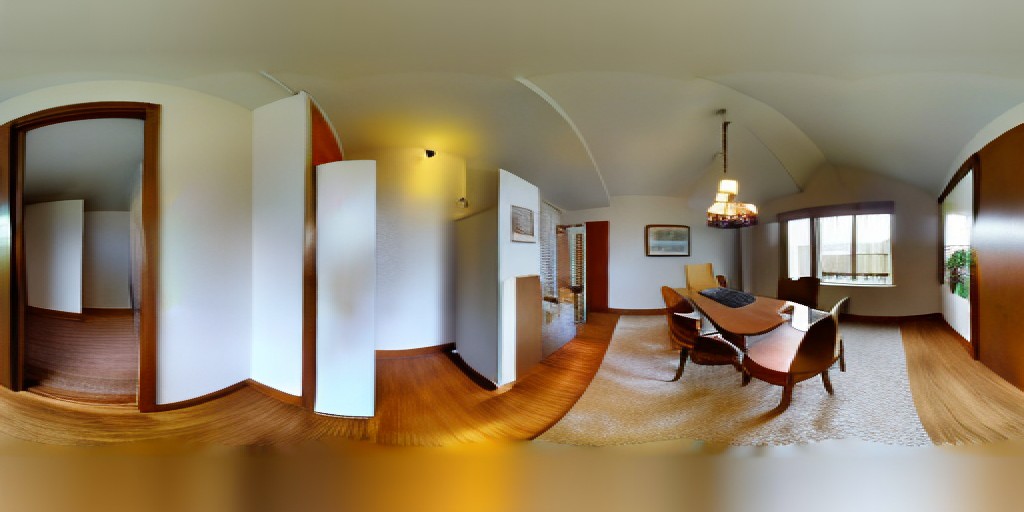}
        \includegraphics[width=\textwidth]{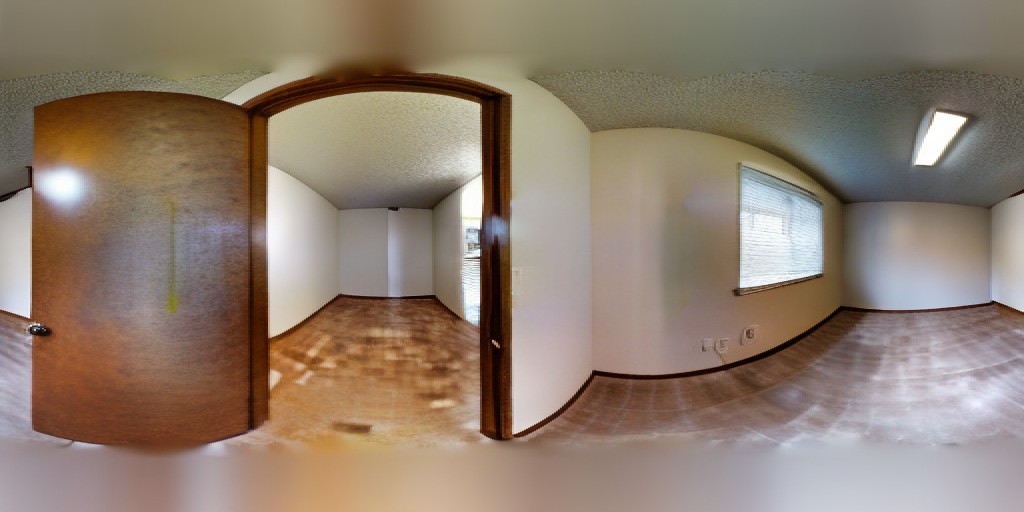}
        \includegraphics[width=\textwidth]{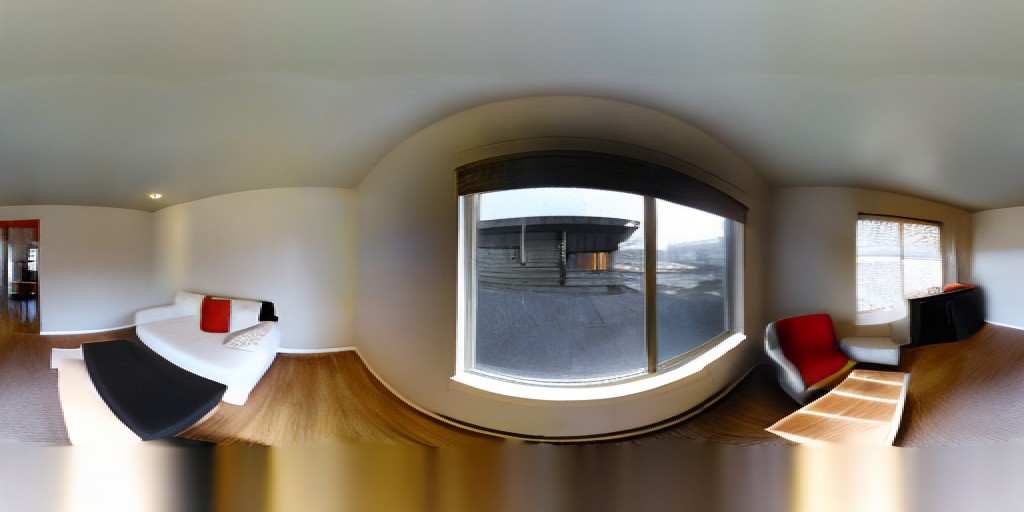}
        \includegraphics[width=\textwidth]{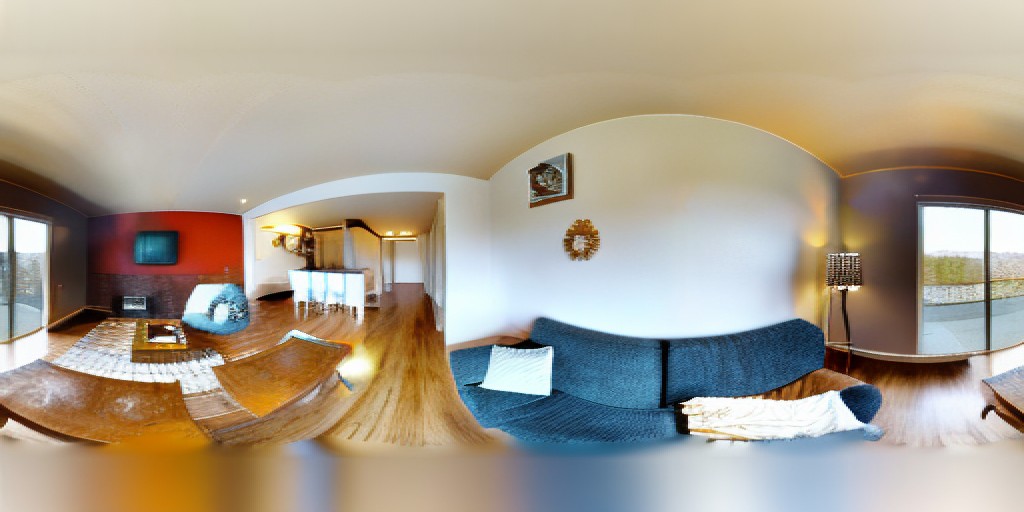}
        \caption{w/o depth}
    \end{subfigure}
    \hfill
    \begin{subfigure}[t]{0.13\textwidth}
        \includegraphics[width=\textwidth]{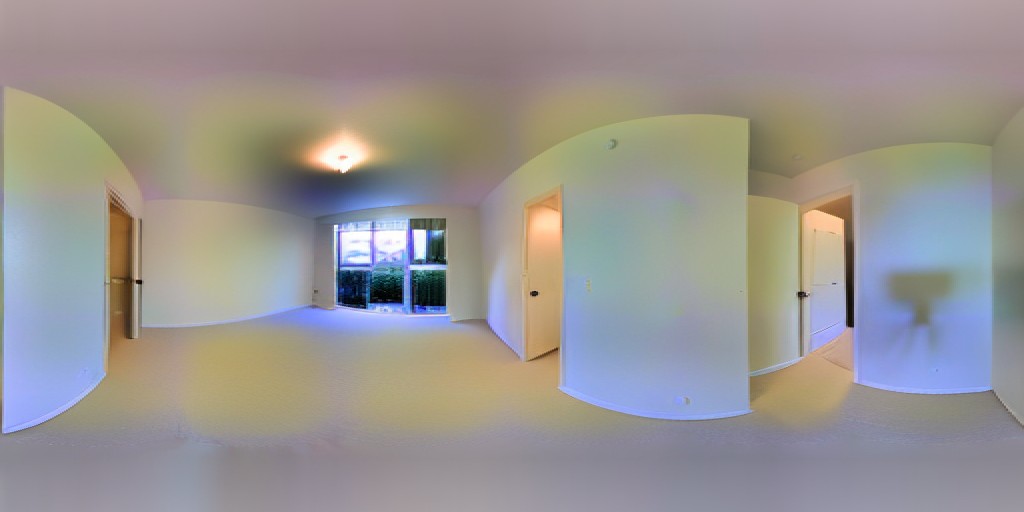}
        \includegraphics[width=\textwidth]{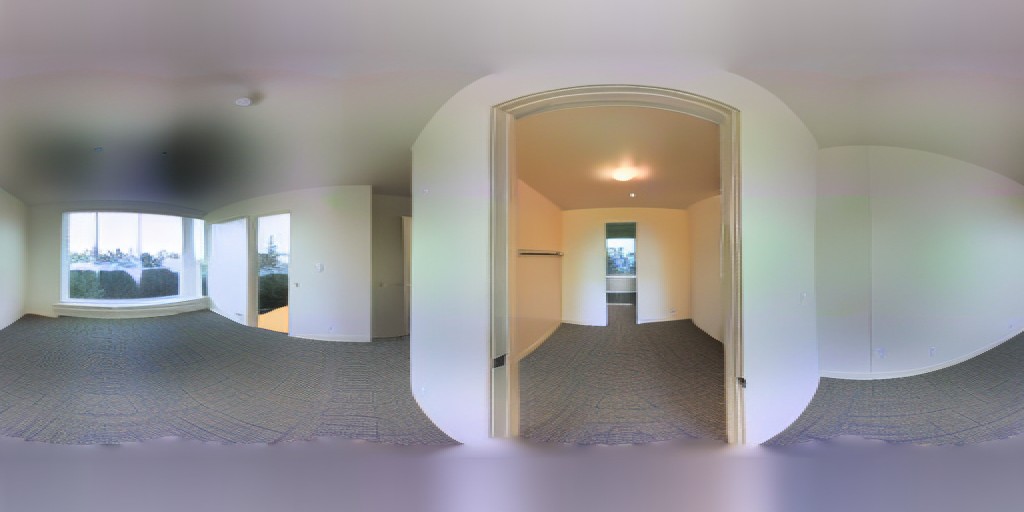}
        \includegraphics[width=\textwidth]{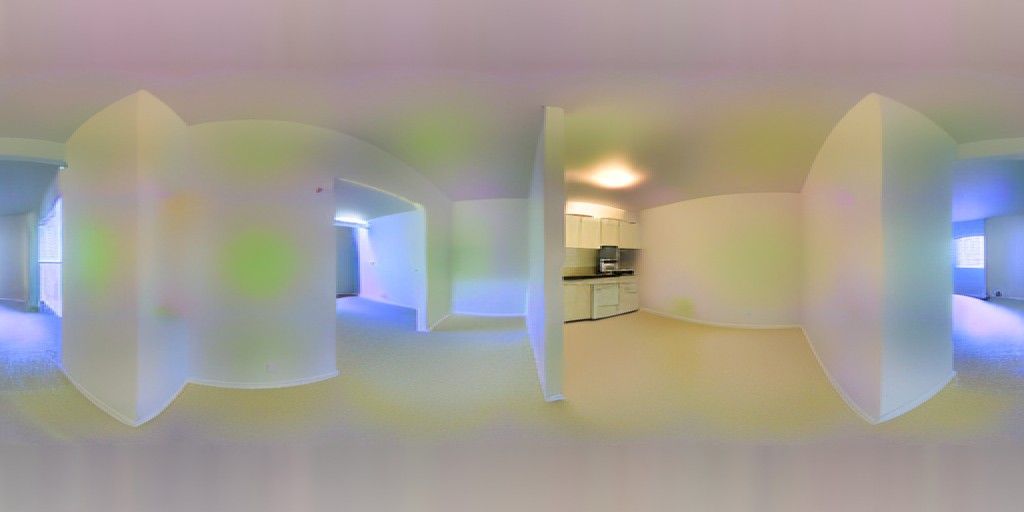}
        \includegraphics[width=\textwidth]{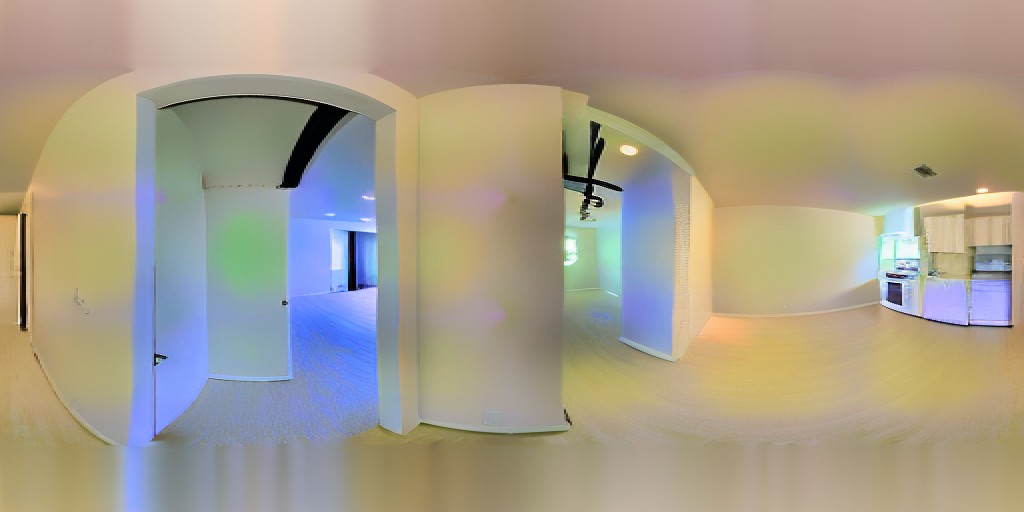}
        \includegraphics[width=\textwidth]{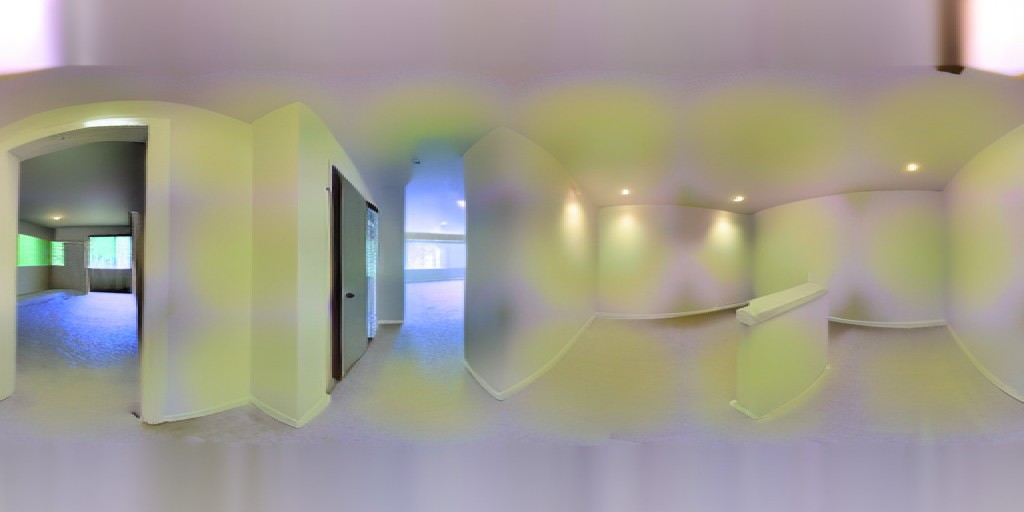}
        \includegraphics[width=\textwidth]{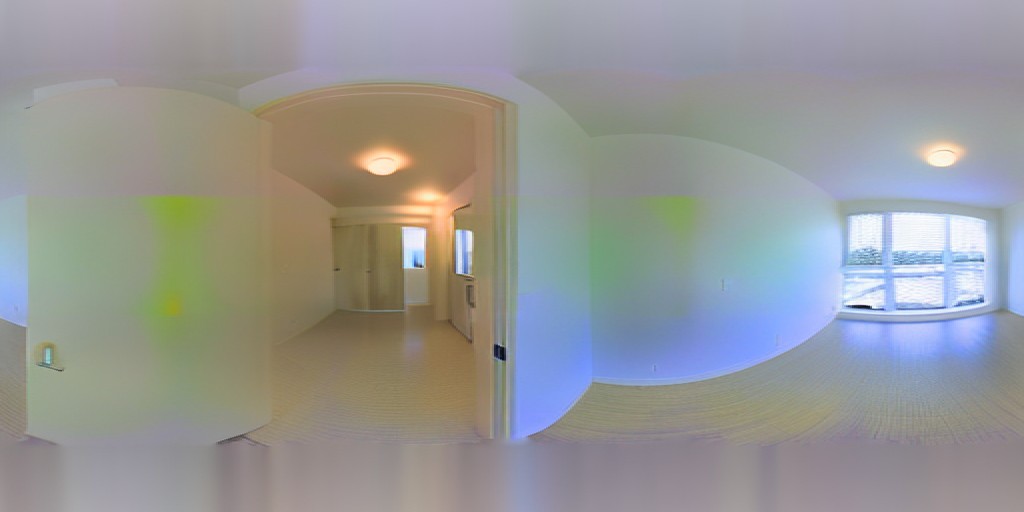}
        \includegraphics[width=\textwidth]{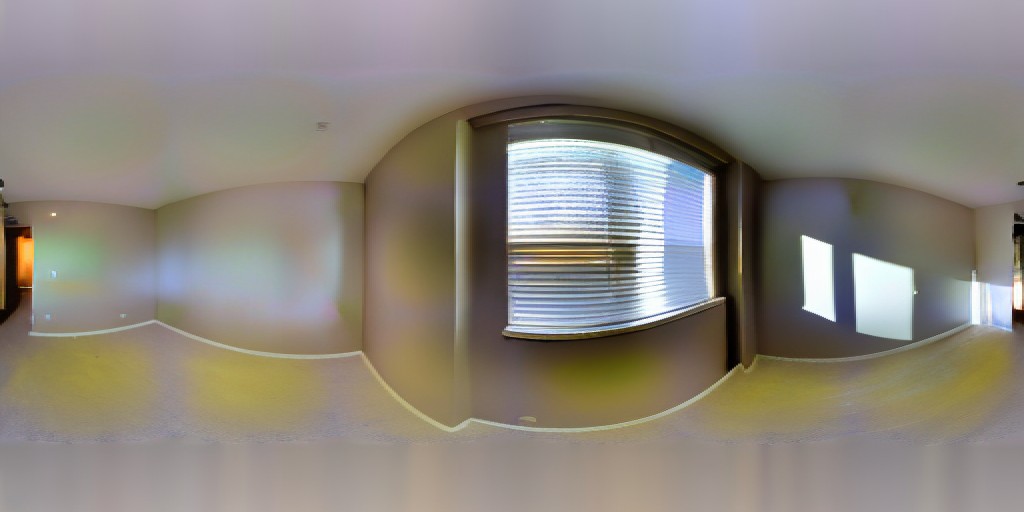}
        \includegraphics[width=\textwidth]{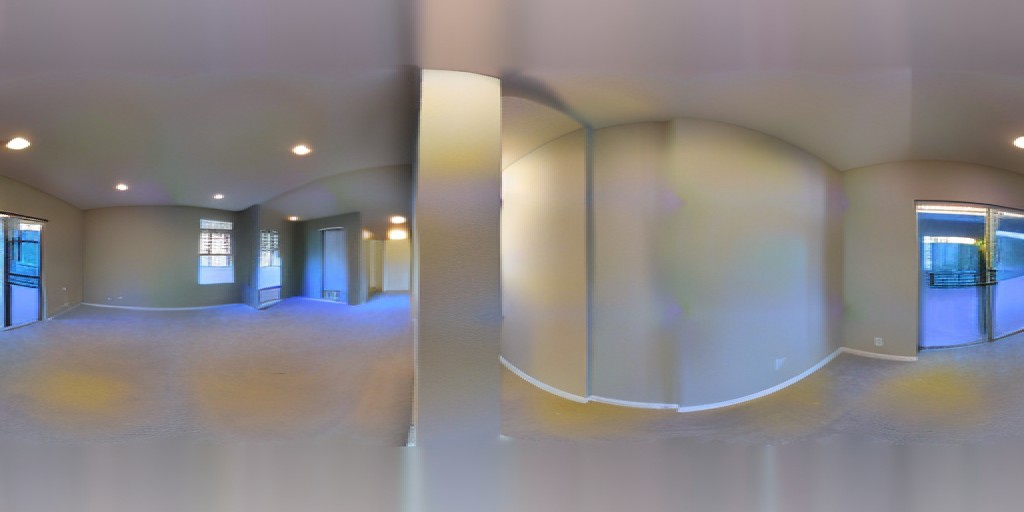}
        \caption{w/o color}
    \end{subfigure}
    \hfill
    \begin{subfigure}[t]{0.13\textwidth}
        \includegraphics[width=\textwidth]{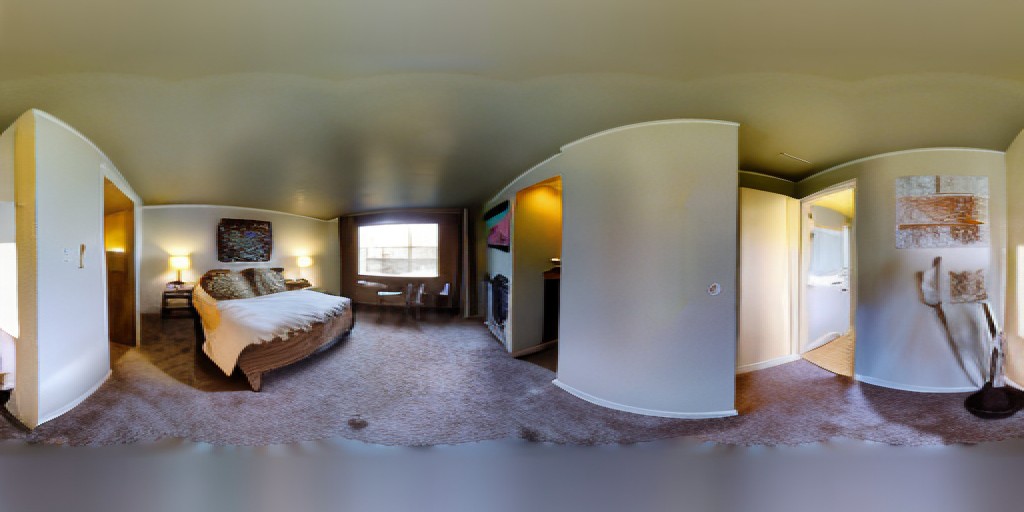}
        \includegraphics[width=\textwidth]{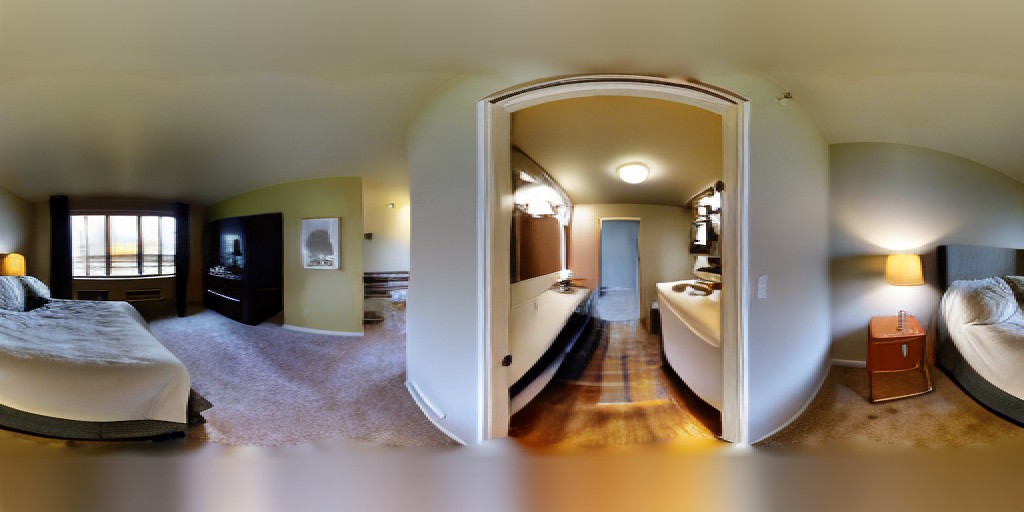}
        \includegraphics[width=\textwidth]{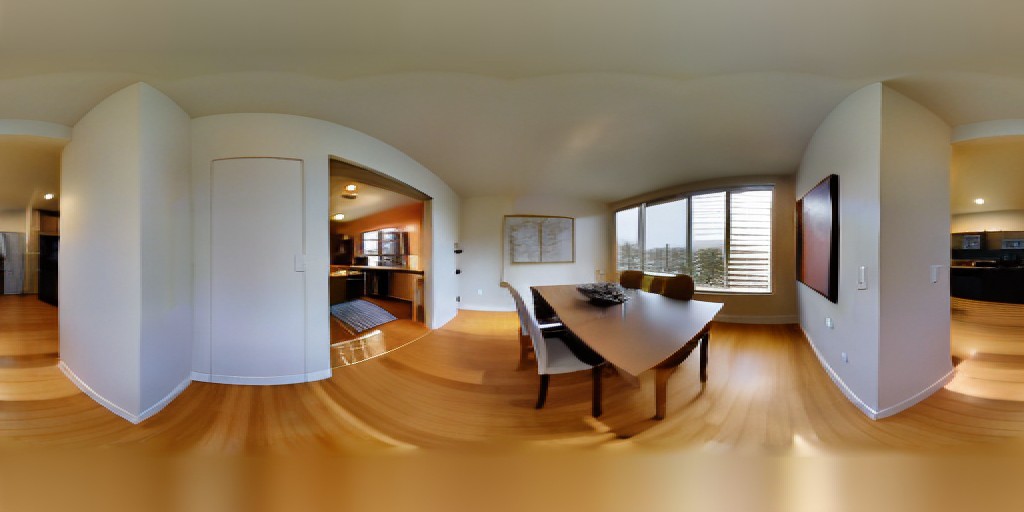}
        \includegraphics[width=\textwidth]{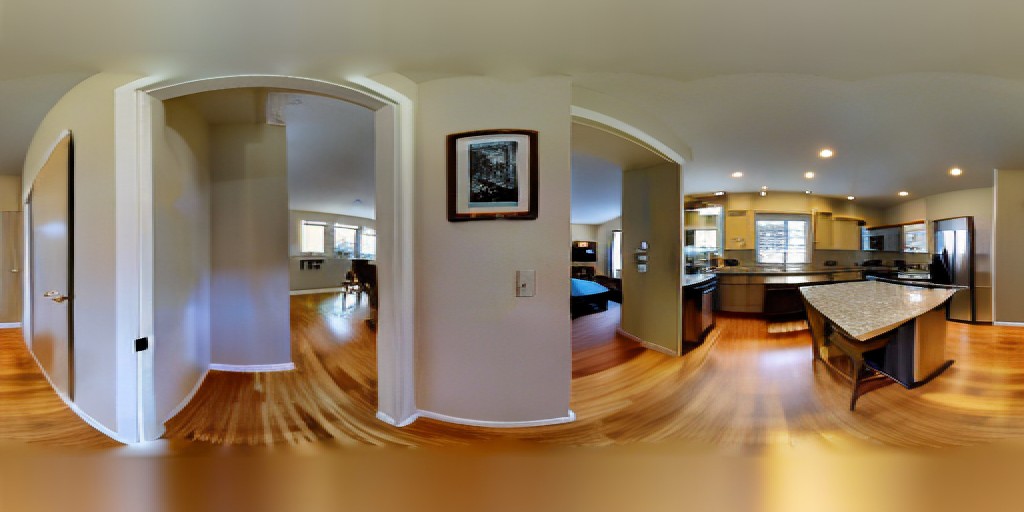}
        \includegraphics[width=\textwidth]{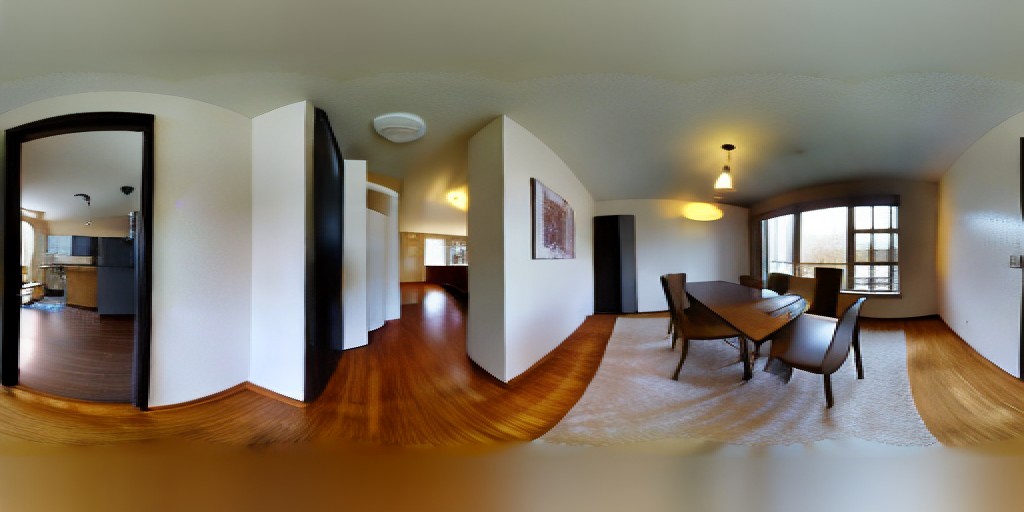}
        \includegraphics[width=\textwidth]{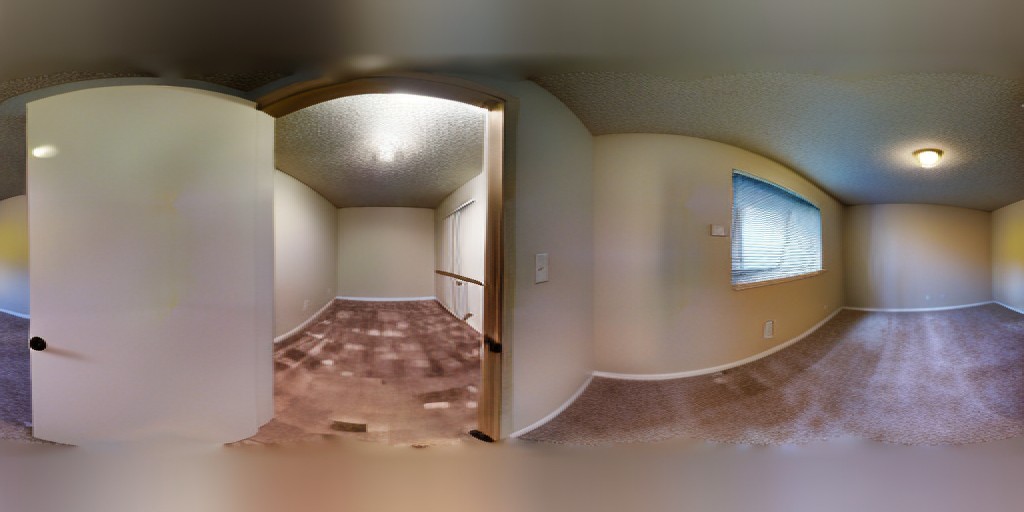}
        \includegraphics[width=\textwidth]{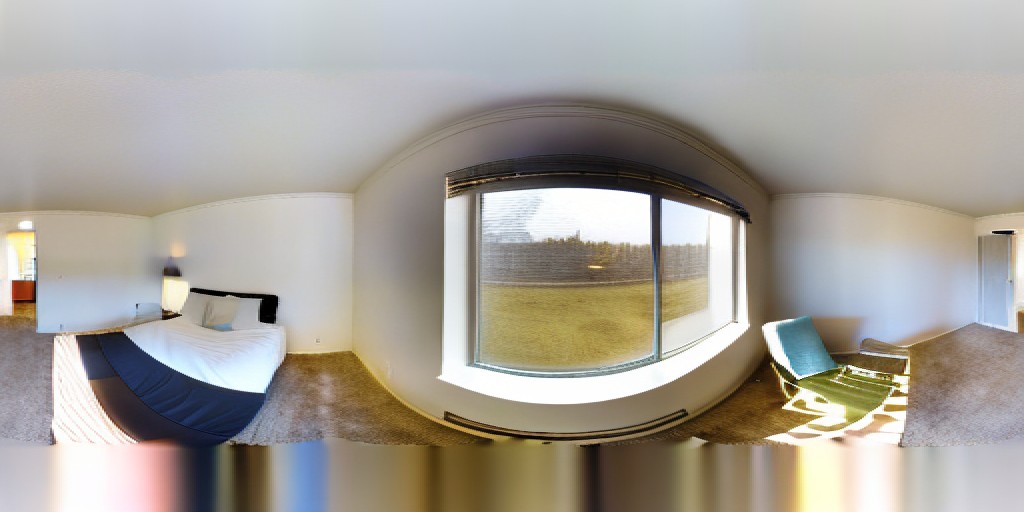}
        \includegraphics[width=\textwidth]{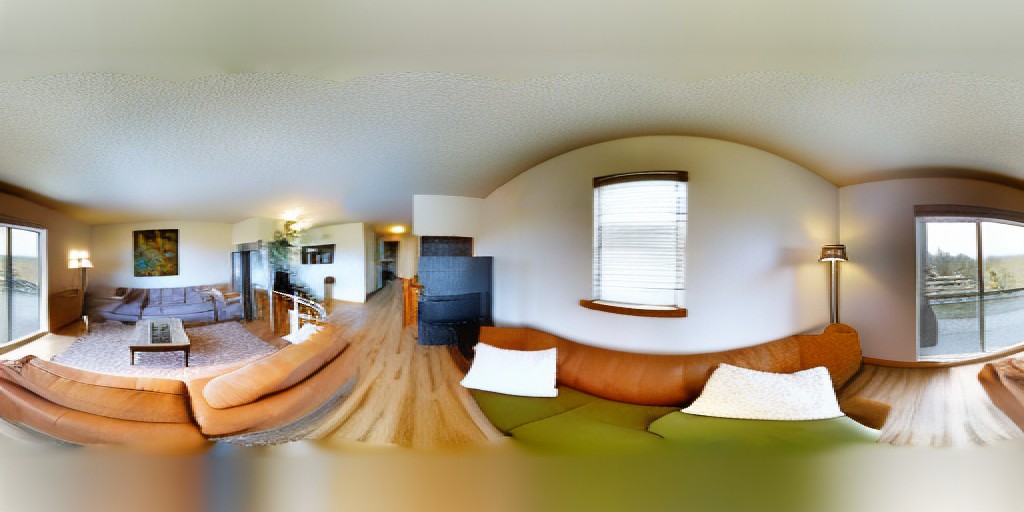}
        \caption{full model}
    \end{subfigure}
    \hfill
    \begin{subfigure}[t]{0.13\textwidth}
        \includegraphics[width=\textwidth]{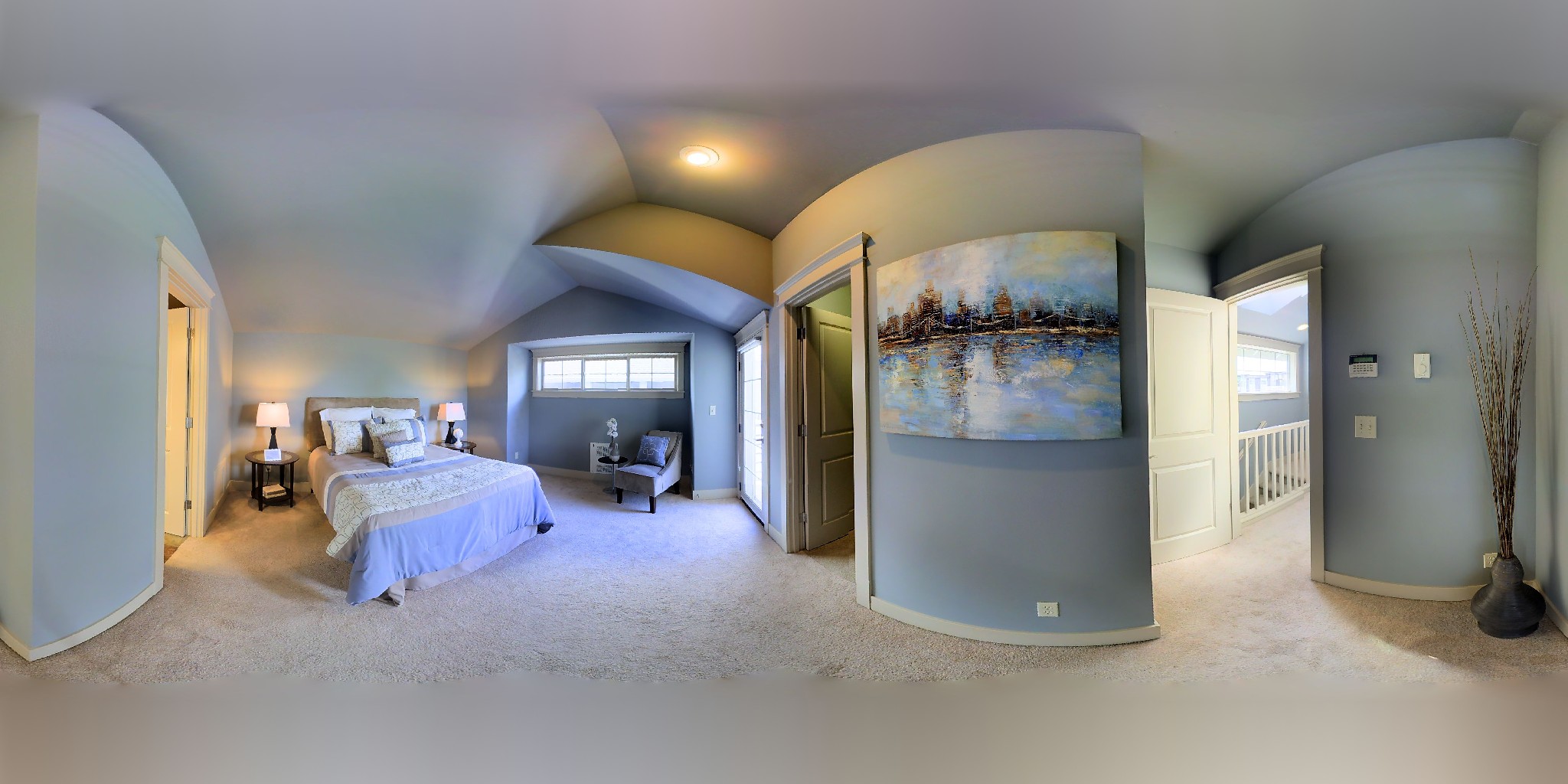}
        \includegraphics[width=\textwidth]{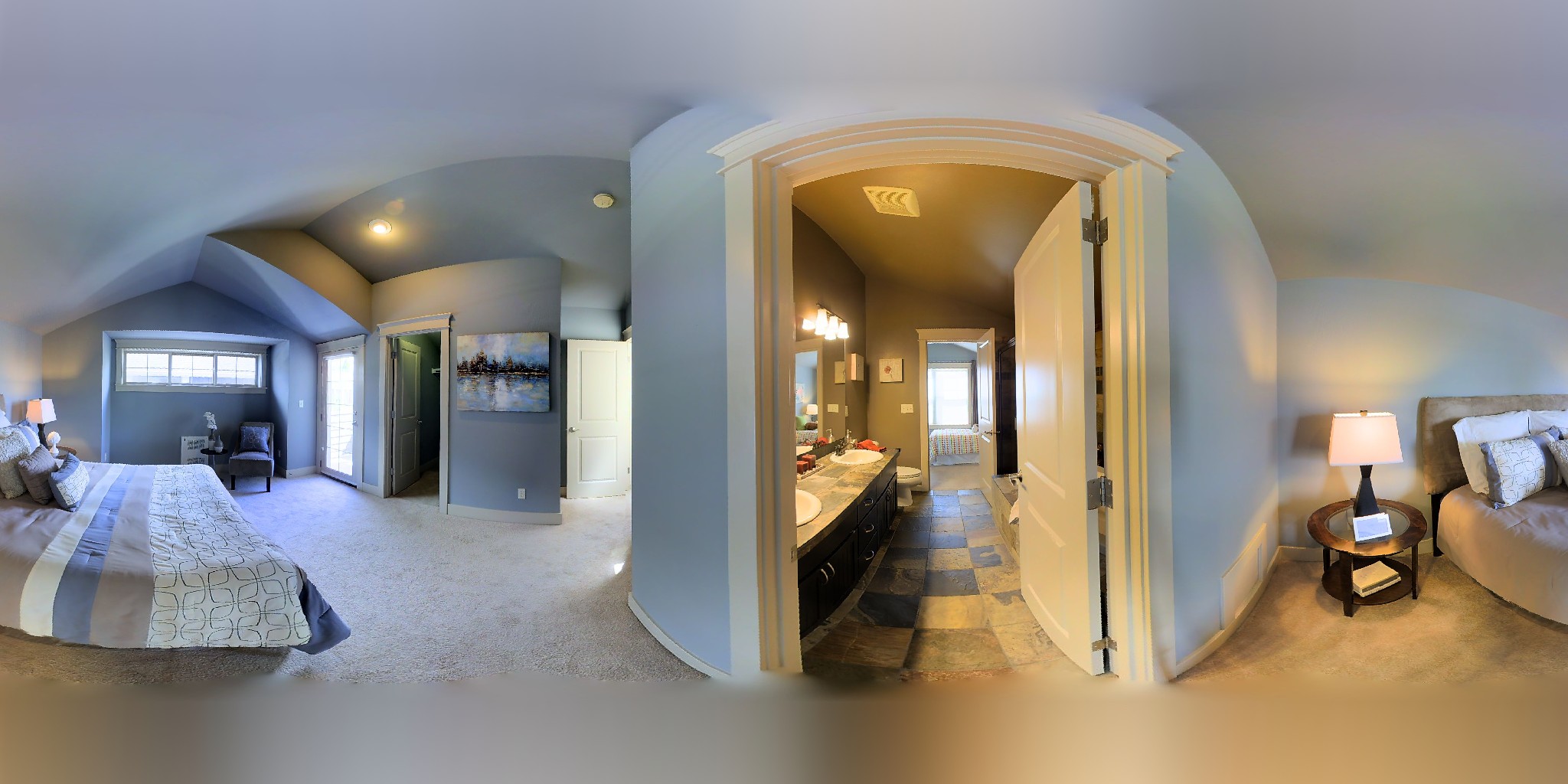}
        \includegraphics[width=\textwidth]{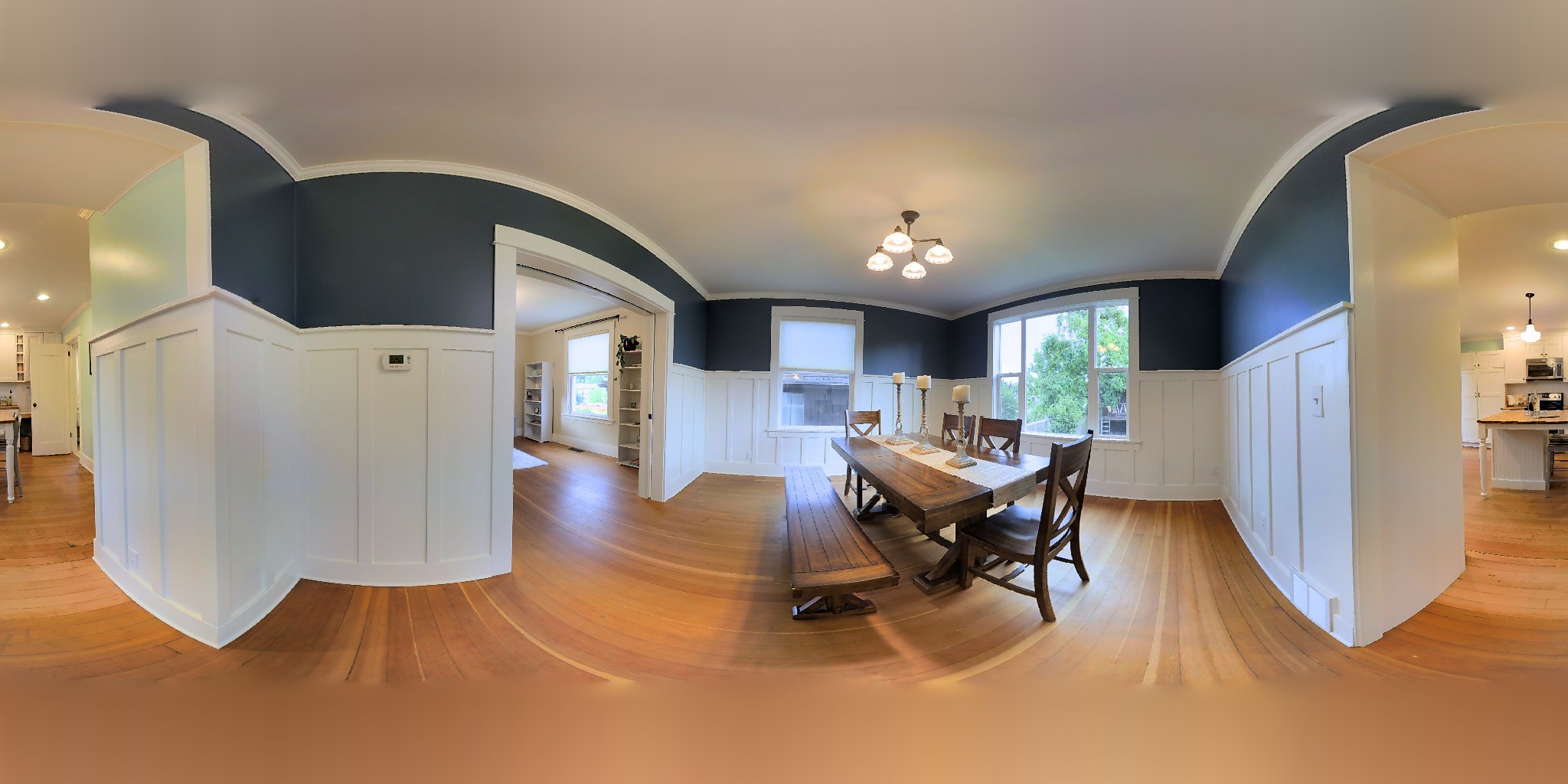}
        \includegraphics[width=\textwidth]{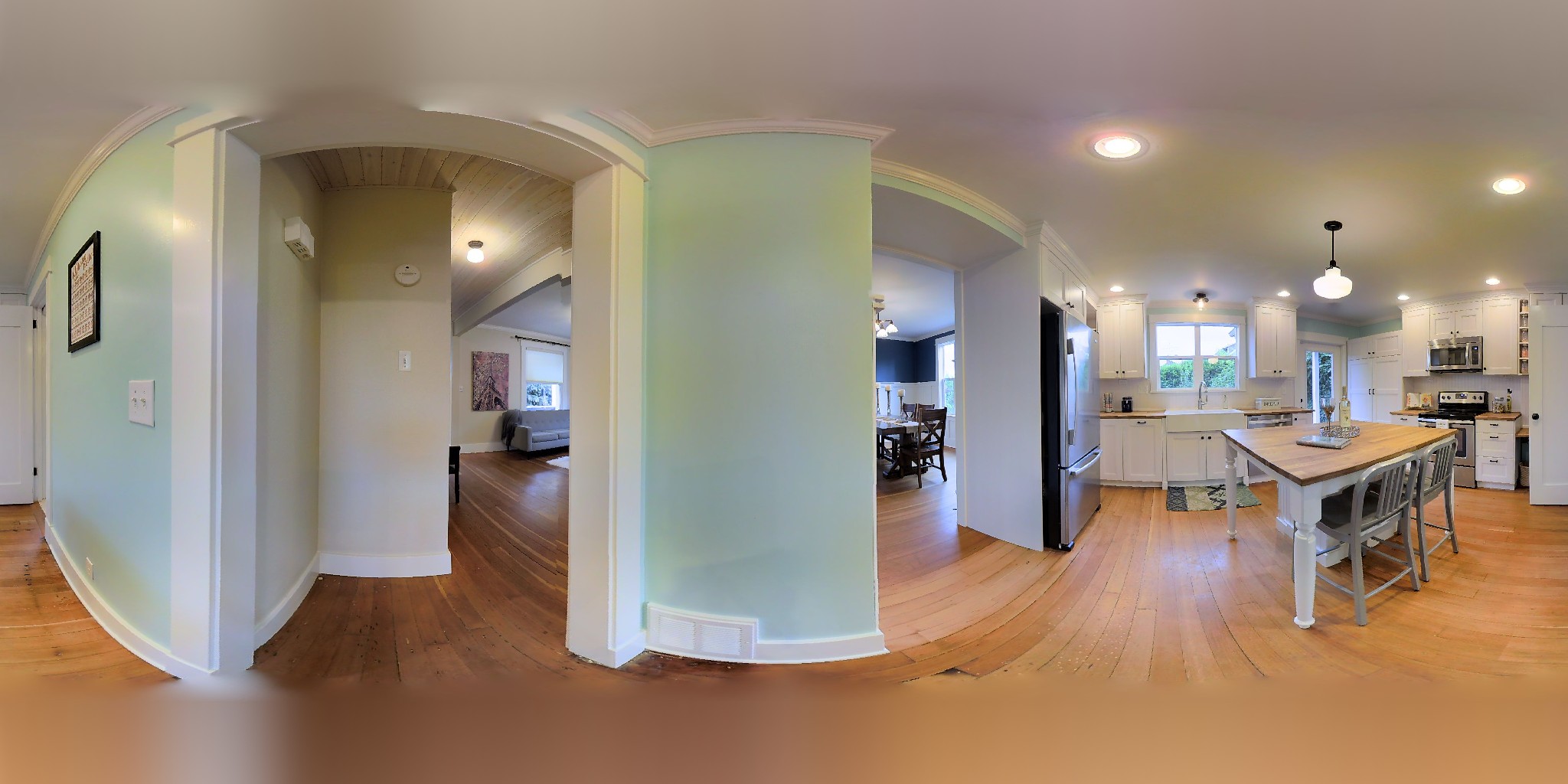}
        \includegraphics[width=\textwidth]{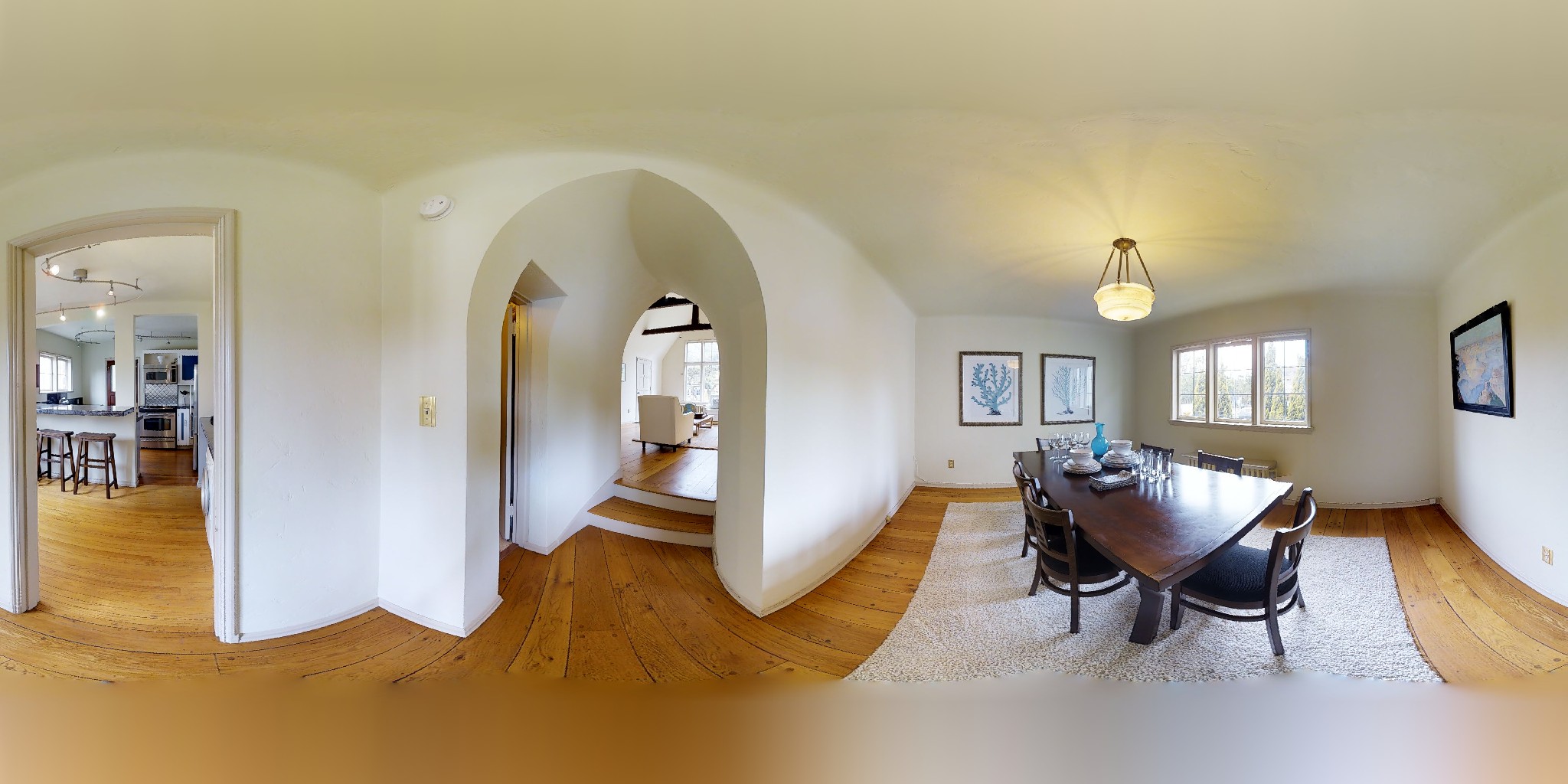}
        \includegraphics[width=\textwidth]{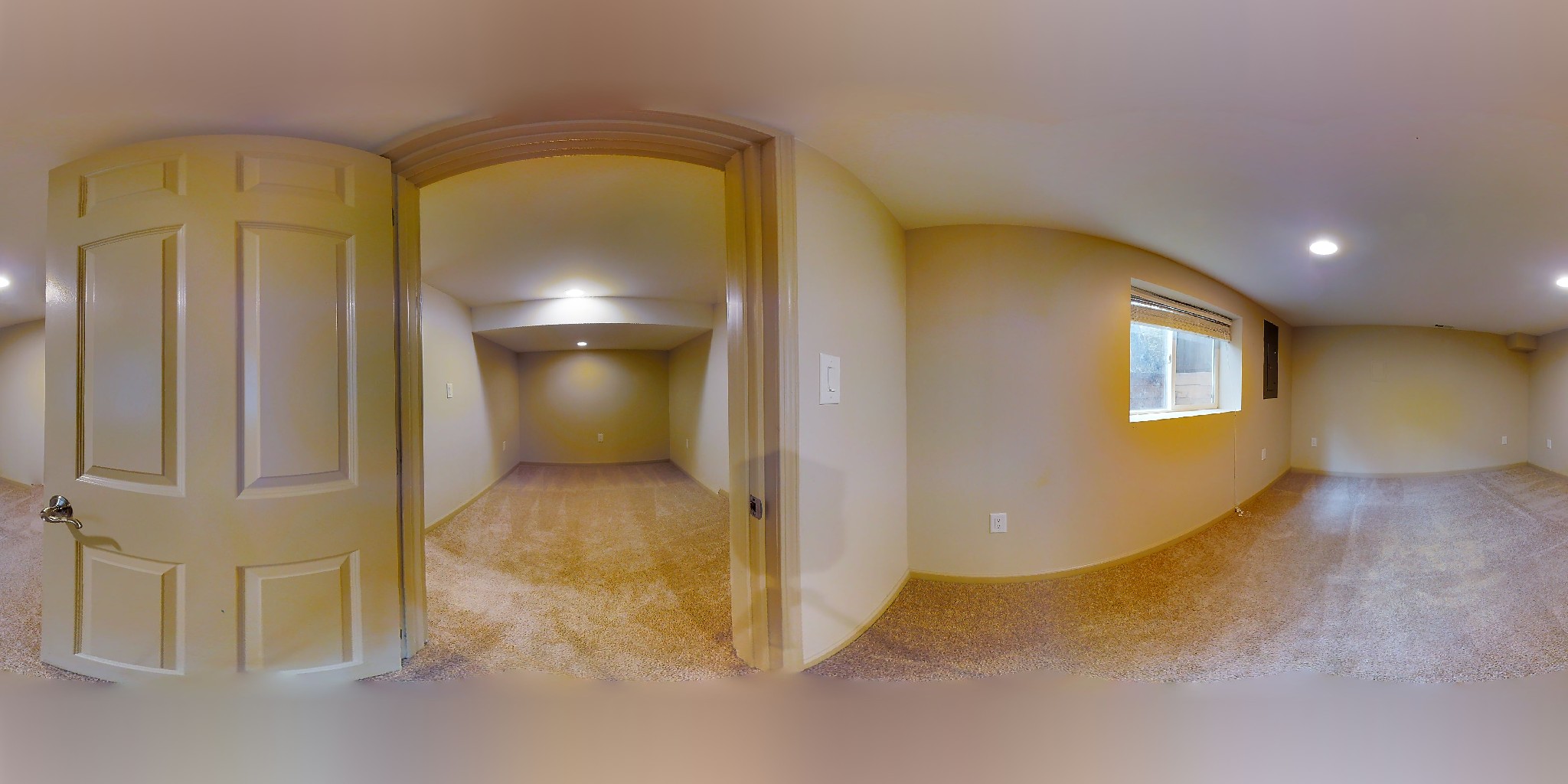}
        \includegraphics[width=\textwidth]{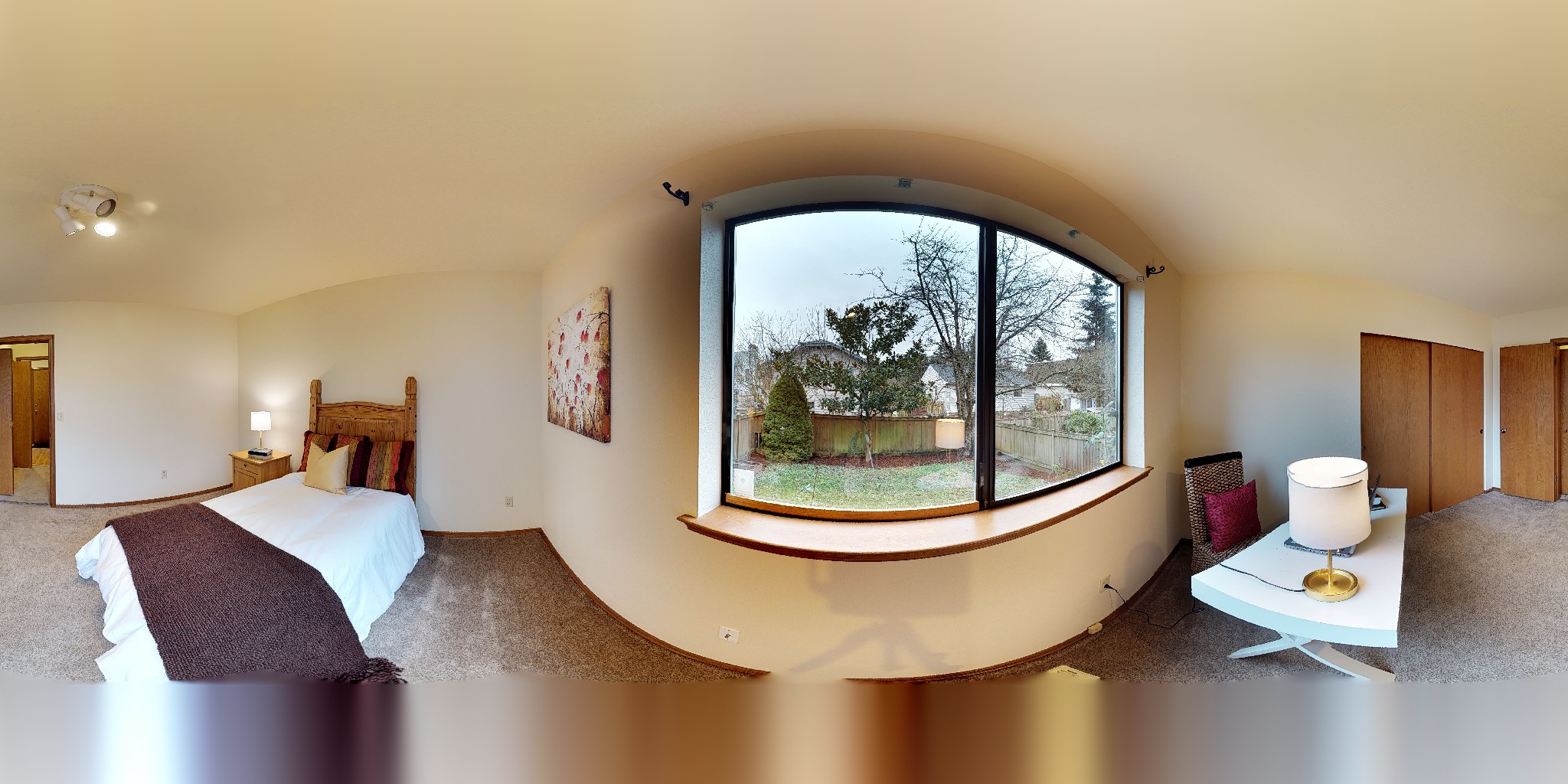}
        \includegraphics[width=\textwidth]{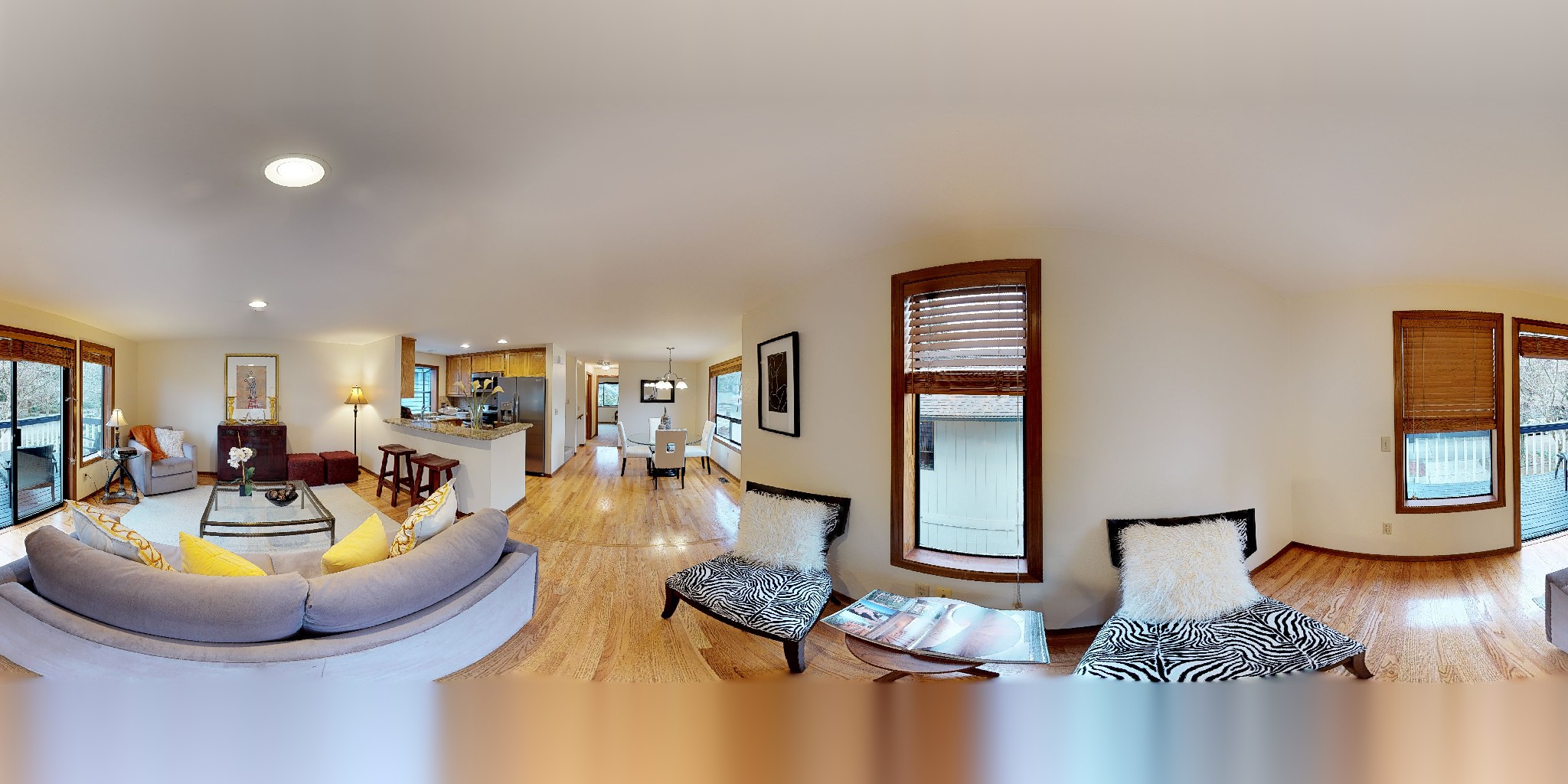}
        \caption{ground truth}
    \end{subfigure}
    \vspace{-2mm}
    \caption{\small Qualitative results of ablation experiments on the Gibson dataset.}
    \label{fig:Gablation}
\end{figure*}

\end{document}